# Large-scale empirical validation of Bayesian Network structure learning algorithms with noisy data

Anthony C. Constantinou[1,2], Yang Liu[1], Kiattikun Chobtham[1], Zhigao Guo[1], and Neville K. Kitson[1].

1. Bayesian Artificial Intelligence research lab, Risk and Information Management (RIM) Research Group, School of EECS, Queen Mary University of London (QMUL), London, UK, E1 4NS.
   E-mails: a.constantinou@qmul.ac.uk, yangliu@qmul.ac.uk, k.chobtham@qmul.ac.uk, zhigao.guo@qmul.ac.uk, and kenkitson@gmail.com.

2. The Alan Turing Institute, British Library, 96 Euston Road, London, UK, NW1 2DB.

**ABSTRACT:** Numerous Bayesian Network (BN) structure learning algorithms have been proposed in the literature over the past few decades. Each publication makes an empirical or theoretical case for the algorithm proposed in that publication and results across studies are often inconsistent in their claims about which algorithm is 'best'. This is partly because there is no agreed evaluation approach to determine their effectiveness. Moreover, each algorithm is based on a set of assumptions, such as complete data and causal sufficiency, and tend to be evaluated with data that conforms to these assumptions, however unrealistic these assumptions may be in the real world. As a result, it is widely accepted that synthetic performance overestimates real performance, although to what degree this may happen remains unknown. This paper investigates the performance of 15 state-of-the-art, well-established, or recent promising structure learning algorithms. We propose a methodology that applies the algorithms to data that incorporates synthetic noise, in an effort to better understand the performance of structure learning algorithms when applied to real data. Each algorithm is tested over multiple case studies, sample sizes, types of noise, and assessed with multiple evaluation criteria. This work involved learning approximately 10,000 graphs with a total structure learning runtime of seven months. It provides the first large-scale empirical validation of BN structure learning algorithms under different assumptions of data noise. The results suggest that traditional synthetic performance may overestimate real-world performance by anywhere between 10% and more than 50%. They also show that while score-based learning is generally superior to constraint-based learning, a higher fitting score does not necessarily imply a more accurate causal graph. To facilitate comparisons with future studies, we have made all data, raw results, graphs and BN models freely available online.

## 1. INTRODUCTION

A Bayesian Network (BN) graph has two different interpretations. If we assume that the edges between variables represent causation, the BN is viewed as a unique Directed Acyclic Graph (DAG), also referred to as a Causal Bayesian Network (CBN). If, however, we assume that the edges between variables can represent some dependency that is not necessarily causal, then the BN is viewed as a dependence graph that can be represented by a Complete Partial Directed Acyclic Graph (CPDAG), where undirected edges indicate relationships that produce identical posterior distributions irrespective of the direction of the edge. Specifically, a CPDAG represents a set of Markov equivalent DAGs.

One of the reasons CBNs have become popular in real-world applications is because they enable decision makers to reason with causal assumptions under uncertainty, which in turn enable them to simulate the effect of interventions and extend them to counterfactual reasoning [1] [2]. As a result, this paper focuses on assessing the various structure learning algorithms in terms of reconstructing the ground truth DAG, rather than in terms of reconstructing a Markov equivalence class that contains the true graph.

In the literature, there are two main classes of algorithms for BN structure learning, known as constraint-based and score-based learning. The constraint-based algorithms use conditional independence tests to construct a graph, whereas the score-based algorithms view the BN structure learning process as a classic machine learning problem where algorithms search the space of possible graphs and return the graph that maximises some score.

The constraint-based algorithms are often referred to as *causal discovery* algorithms [3], although with some attendant controversy [4] [5] [6] [7] [8] [9], and return a CPDAG where directed edges represent causation and undirected edges represent dependency in which the direction of causation cannot be determined by observational data. In addition to these two main classes of algorithms, there exists a class of hybrid learning which adopts strategies from both constraint-based and score-based learning, and such algorithms may occasionally return a CPDAG, although they will generally return a DAG.

The PC algorithm is one of the first and more popular constraint-based algorithms [10]. Several variants of this algorithm have been published that address different issues. Well-established variants include PC-Stable that resolves the issue of PC with regards to its dependency on the order of the variables as they appear in the data [11], and the FCI which extends PC to account for the possibility of latent variables [12]. Other variants include the H2PC that uses rules of PC to construct a skeleton with score-based learning determining the final output [13], and the recent PC-MI algorithm [14] which is said to improve the speed and accuracy of PC by modifying the order in which the edges are assessed by means of mutual information.

Score-based learning provides both exact and approximate solutions. Exact solutions guarantee to return the graph that maximises an objective function subject to limiting the maximum in-degree. Algorithms such as those based on integer linear programming approaches offer exact solutions assuming infinite runtime [15, 16]. However, because structure learning is NP-hard where the difficulty grows with the number of the variables in the data, exact solutions are generally restricted to smaller networks. In general, BN structure learning provides exact solutions to low dimensionality problems, and approximate solutions that scale up to hundreds or thousands of variables.

Various score-based approaches have been proposed for approximate learning. Some of the more widely adopted strategies are based on the Greedy Equivalence Search (GES) algorithm proposed by Meek [17], which explores the space of graphs across Markov equivalence classes rather than unique DAGs. Numerous variants of GES have been proposed [18], including the optimised GES (also called FGES) variant by Chickering [19]. Classic search heuristics are also popular such as Hill-Climbing and Tabu search that form part in various score-based [20] and hybrid learning algorithms [13, 21, 22, 23, 24]. Approaches that rely on bounded treewidth that represent a low or a minimum width over all tree decompositions of a graph, such as the k-MAX method [25], and other order-based approaches [26], reduce the complexity of the inferences and enable structure learning to scale up to thousands of variables. Other approaches achieve similar scale by restricting search to local structures that represent subnetworks of the global network. These approaches, however, may lead to conflicts between subnetworks which tend to be solved by ordering subnetworks by score, or by symmetry correction [27], and some of these approaches are extended to work with both discrete and continuous data [28]. Lastly, Ant Colony [29] and Bee Colony [30] are examples of swarm intelligence approaches to DAG structure learning that are in many ways similar to classic heuristic approaches, such as hill-climbing, although they include an element of randomness [31].

While structure learning algorithms demonstrate good performance in synthetic experiments, it is widely acknowledged that this level of performance tends not to extend to real applications. However, the level of difference in performance between synthetic and real experiments remains uncertain. One of the reasons it is difficult to measure the 'real' performance of algorithms is because in the real world we normally have no knowledge of the ground truth causal graph on which the real-world dataset is based, and which is required to validate these algorithms in terms of their ability to reconstruct the true graph. This paper is motivated by this problem and investigates the impact of data noise on structure learning performance.

A recent relevant study by Scutari et al [32] examined the difference in accuracy and speed of the various structure learning algorithms implemented in the *bnlearn* R package [33], and a previous study investigated the performance of PC, GOBNILP and K2 methods in terms of learning gene networks [34]. In this paper, we test a larger set of algorithms available in several structure learning packages, where the focus is to better approximate the real-world performance of these algorithms by incorporating different types and levels of noise in the data. We investigate the performance of algorithms in all three classes of learning, with and without data noise. The paper is structured as follows: Section 2 discusses the case studies, Section 3 describes the process of generating synthetic and noisy data, Section 4 discusses the selected algorithms, Section 5 describes the evaluation process, Section 6 presents the results, and we provide our concluding remarks in Section 7.

## 2. CASE STUDIES

Six discrete BN case studies are used to generate data. The first three of them represent well-established examples from the BN structure learning literature, whereas the other three represent new cases and are based on recent BN real-world applications. Specifically,

i. **Asia**: A small toy network for diagnosing patients at a clinic [35];

ii. **Alarm**: A medium-sized network based on an alarm message system for patient monitoring [36];

iii. **Pathfinder**: A very large network that was designed to assist surgical pathologists with the diagnosis of lymph-node diseases [37];

iv. **Sports**: A small BN that combines football team ratings with various team performance statistics to predict a series of match outcomes [38];

v. **ForMed**: A large BN that captures the risk of violent reoffending of mentally ill prisoners, along with multiple interventions for managing this risk [39];

vi. **Property**: A medium BN that assesses investment decisions in the UK property market [40].

The case studies are restricted to problems of up to 100s of variables due to time constraints[1]. While 100s of variables are more than sufficient to model causal relationships in most real-world areas, the results presented in this paper may not apply to bioinformatics where biology datasets often include 1,000s of variables. The properties of the six case studies are detailed in Table 1.

[1] The current tests have led to more than ten thousand graphs and required months of runtime to complete all categories of data noise.

**Table 1.** The properties of the six case studies.

| Case study | Nodes | Arcs | Average in-degree | Max in-degree | Free parameters |
|---|---|---|---|---|---|
| Asia | 8 | 8 | 2.00 | 2 | 18 |
| Alarm | 37 | 46 | 2.49 | 4 | 509 |
| Pathfinder | 109 | 195 | 3.58 | 5 | 71890 |
| Sports | 9 | 15 | 3.33 | 2 | 1049 |
| ForMed | 88 | 138 | 3.14 | 6 | 912 |
| Property | 27 | 31 | 2.30 | 3 | 3056 |

Still, the selected case studies offer a good range of old and new, as well as simple and complex, BNs that come from different application domains. Moreover, the three traditional networks of Asia, Alarm and Pathfinder are based on knowledge-based priors, whereas the parameters of the Sports and ForMed networks are determined by real data, and the Property network is based on parameters determined by clearly defined rules and regulating protocols[2]. It is also worth mentioning that the Sports network is the only case study in which all data variables are ordinal.

The case studies provide a good range of model dimensionality that results from varied max in-degree, number of nodes and arcs. In BNs, a good measure of dimensionality is the number of free parameters $p$ (also known as independent parameters) that each network incorporates. In this paper, $V$ represents the set of the variables $v_i$ in graph $G$, and $|V|$ is the size of set $V$. The number of free parameters $p$ is:

$$p = \sum_{i}^{|V|} (r_i - 1) \prod_{j}^{|\pi_{v_i}|} q_j$$

where $r_i$ is the number of states of $v_i$, $\pi_{v_i}$ is the parent set of $v_i$, $|\pi_{v_i}|$ is the size of set $\pi_{v_i}$, and $q_j$ is the number of states of $v_j$ in parent set $\pi_{v_i}$. For example, while both Asia and Sports are small BNs that incorporate eight and nine nodes respectively, they differ considerably in model dimensionality since there are 18 free parameters in Asia while there are 1,049 free parameters in Sports. A similar comparison applies to ForMed and Pathfinder which incorporate 88 and 109 nodes respectively, yet there are just 912 free parameters in ForMed and 71,890 free parameters in Pathfinder.

All case studies represent discrete BNs and incorporate different numbers of states. For example, while in Asia all variables are Boolean, the number of states per variable in Sports ranges from five to eight, which also explains the large difference in model dimensionality between the two. Moreover, Pathfinder represents an interesting challenge since, in addition to incorporating the highest number of nodes and arcs, it includes a variable that has 63 states, which also partly explains the high dimensionality of this network.

[2] Due to human-made policies. For example, mortgage payments are determined by the interest rate in conjunction with the mount borrowed, and tax payments are based on specific thresholds associated to different levels of income.

## 3. METHODOLOGY

Each of the BN case studies presented in Section 2 is used to generate synthetic data under various assumptions of data noise, so that multiple datasets are produced for each case study. The methodology involves producing multiple datasets that fall under different categories of synthetic data, with and without data noise. We briefly describe each of the synthetic data categories below, and further details on how each of these datasets are generated or modified with noise, are provided in Appendix A. The categories are:

i. **No noise (N)**: this represents the traditional approach used throughout the literature where data are generated as determined by the CPTs of each BN.

Each $N$ dataset is then manipulated to incorporate noise. The following types of data noise are used:

ii. **Missing values (M)**: where each data value has 5% or 10% risk[3] (both percentages tested) to be replaced with a new value called 'missing'. Most algorithms do not accept datasets that incorporate empty cells, which is why we introduced a new state called 'missing'. This assumption aims to approximate the performance of the algorithms when applied to real datasets that incorporate missing values, although only covers the situation where values are Missing Completely At Random (MCAR).

iii. **Incorrect values (I)**: where each data value has 5% or 10% risk (both percentages tested) to be replaced with one of the other possible values for the variable. For example, a value of $a$ for a variable with states $\{a, b, c\}$ could be, modified into either $b$ or $c$ at random. This assumption aims to approximate the performance of the algorithms when applied to real datasets that incorporate some inaccurate data values.

iv. **Merged states (S)**: where approximately 5% or 10% of the variables (both cases tested) have two of their states merged into one. For example, a variable with states $\{a, b, c\}$ would have two random states, such as $a$ and $b$, both modified into a new state $ab$. This assumption aims to approximate the performance of the algorithms when applied to real datasets where some of the data variables have had their number of states decreased in an effort to reduce the dimensionality of the model.

v. **Latent variables (L)**: where approximately 5% or 10% of the variables (both cases tested) are randomly removed from the data. This assumption aims to approximate the performance of the algorithms when applied to datasets that incorporate latent variables.

vi. **Combo (C)**: this category represents all dual combinations of the noisy categories above (i.e., M, I, S, and L), plus the combination of all four noisy categories. Because these experiments incorporate multiple types of noise, we chose the rate of 5% as the default rate of noise for each type of noise incorporated into a dataset; although for some datasets we had to go with the rate of 10% for reasons we explain below.

[3] We give each data value a fixed chance to become noisy, rather than uniformly sampling a percentage from the dataset.

The above categories have led to the 16 different experiments, per case study per sample size, shown in Table 2, where the experiment code consists of a) letters that correspond to the type of noise and b) integers that correspond to the rate of noise. Moreover, experiment codes that start with letter 'c' represent category 'Combo', followed by the letters that indicate the types of noise used in that particular experiment.

**Table 2.** The 16 experiments E (i.e., synthetic datasets) where X indicates the type of noise used. Experiment code N is no noise, M is missing values, I is incorrect values, S is merged states, L is latent variables, and *c* is Combo, as defined in Section 2. The integers in the experiment code correspond to the rate of noise.

| E | E code | No noise | Missing values | | Incorrect values | | Merged states | | Latent variables | |
|---|---|---|---|---|---|---|---|---|---|---|
| | | | [5%] | [10%] | [5%] | [10%] | [5%] | [10%] | [5%] | [10%] |
| 1 | N | X | | | | | | | | |
| 2 | M5 | | X | | | | | | | |
| 3 | M10 | | | X | | | | | | |
| 4 | I5 | | | | X | | | | | |
| 5 | I10 | | | | | X | | | | |
| 6 | S5 | | | | | | X | | | |
| 7 | S10 | | | | | | | X | | |
| 8 | L5 | | | | | | | | X | |
| 9 | L10 | | | | | | | | | X |
| 10 | cMI | | X | | X | | | | | |
| 11 | cMS | | X | | | | X | | | |
| 12 | cML | | X | | | | | | X | |
| 13 | cIS | | | | X | | X | | | |
| 14 | cIL | | | | X | | | | X | |
| 15 | cSL | | | | | | X | | X | |
| 16 | cMISL | | X | | X | | X | | X | |

Table A1 shows which types of noise were not performed for reasons we clarify in Appendix A. For example, experiment S could not have been applied to Boolean BNs, such as the Asia BN, because the states of binary variables cannot be reduced further. Moreover, the Asia and Sports BNs incorporate eight and nine nodes respectively, which means experiments S and L, which involve manipulation of variables rather than data values, could be performed only at 10% level of randomisation; i.e., manipulating just one variable corresponds to a rate of manipulation just above 10% in both cases. For further details, see Appendix A.

Each experiment enumerated in Table 2 is tested with five different sample sizes. These are 0.1k, 1k, 10k, 100k, and 1000k samples. The different sample sizes are subsets of the first rows of the largest dataset. This makes the potential[4] number of possible experiments per case study per algorithm 80 (i.e., 16 experiments over five sample sizes), and the potential number of possible experiments per algorithm 480 (i.e., 80 experiments over six case studies). This information is displayed in Table 3. Appendix A provides additional supplementary information.

[4] This number is a potential maximum because, as previously discussed, not all types of noise could have been applied to all case studies (refer to Appendix A).

**Table 3.** The number of experiments carried out per algorithm per case study, given five different sample sizes for each experiment.

| Experiment | Experiments per case study per algorithm | Experiments per algorithm |
|---|---|---|
| No noise | 5 | 30 |
| Missing values | 10 | 60 |
| Incorrect values | 10 | 60 |
| Merging values | 10 | 60 |
| Latent variables | 10 | 60 |
| Combo | 35 | 210 |
| TOTAL | 80 | 480 |

## 4. ALGORITHMS

The selected algorithms represent state-of-the-art and/or well-established implementations, including some recent algorithms. We aimed for a varied selection of algorithms across all three types of learning; constraint-based, score-based, and hybrid learning algorithms. A total of 15 algorithms have been evaluated. These are:

i. **PC-Stable**: a modified version of the classic constraint-based algorithm called PC [10] [11]. PC-Stable solves the issue on the order dependency of the variables in the data by changing the order of edge deletion and combining the process with the CPC algorithm [41]. The PC-Stable algorithm is commonly used when benchmarking constraint-based learning algorithms.

ii. **FGES**: an efficient version of the score-based algorithm proposed by Chickering [19] that was based on GES algorithm proposed by Meek [17]. Instead of searching in the DAG space which grows super-exponentially with the number of variables, FGES searches in the space of equivalence classes of DAGs [18]. While FGES adds and removes a polynomial number of edges, the space of equivalence classes is likely to also be super-exponential according to Gillispie and Perlman [42]. FGES is commonly used as a state-of-the-art algorithm when benchmarking score-based learning algorithms.

iii. **FCI**: Fast Causal Inference (FCI) is a constraint-based algorithm that can be viewed as an extended version of the PC algorithm that accounts for the possibility of latent variables in the data [12]. It performs a two-phase learning process that produces the skeleton of the graph (i.e., an undirected graph) followed by conditional independence tests that orientate some of those edges. FCI is commonly used when benchmarking algorithms under assumptions of causal insufficiency.

iv. **GFCI**: The Greedy FCI (GFCI) is a hybrid algorithm that combines the score-based FGES and the constraint-based FCI algorithms and hence, also accounts for the possibility of latent variables in the data [22]. Specifically, it obtains the skeleton of the graph using the first phase of FGES, although it does not preserve all of the edges of FGES, and then uses FCI-based orientation rules to identify the direction of some of the edges in the skeleton graph.

v. **RFCI-BSC**: is another FCI-based hybrid learning version of the constraint-based algorithm called RFCI [43], which in turn is a faster version of FCI. RFCI-BSC produces multiple probable models and selects the graph with the highest probability as the preferred graph [44]. Its process involves bootstrap sampling (re-sample with replacement) that generates multiple datasets and performs RFCI-based conditional independence tests on each of those datasets. This approach makes the algorithm non-deterministic. Because of this, we had to run RFCI-BSC 10 times on each experiment and record the average of the results. Therefore, this algorithm was executed 10 times more than the other algorithms used in this paper.

vi. **Inter-IAMB**: uses the concept of Markov Blanket to reduce the number of conditional independence tests. The Markov Blanket of a particular variable is defined as the conditioning set of variables which ensure that the particular node is conditionally independent of all other variables in the graph. Specifically, in BNs the Markov Blanket of a node consists of its parents, its children, and the parents of its children. Inter-IAMB is an improved version of IAMB that both minimises and improves the accuracy of the Markov Blanket candidate set for each node [45]. A smaller Markov Blanket set would yield a more precise CI test result.

vii. **MMHC**: The Max-Min Hill-Climbing (MMHC) algorithm first constructs the skeleton of the BN using a constraint-based algorithm know as Max-Min Parents and Children (MMPC) and then performs a greedy hill-climbing score-based learning search to orientate the edges [23]. It is one of the most popular BN structure learning algorithms applicable to high-dimensional problems, and is often used as a state-of-the-art algorithm when benchmarking structure learning algorithms.

viii. **GS**: Grow-Shrink (GS) was the first constraint-based algorithm to use the concept of Markov Blanket to reduce the number of conditional independence tests [46]. Specifically, GS first finds the Markov-Blanket of each variable and then identifies direct neighbours in the Markov Blanket, thus determining the graph's undirected skeleton, before orientating as many edges as possible to produce a CPDAG.

ix. **HC**: Hill Climbing (HC) is a score-based greedy algorithm which searches the space of DAGs [47]. The algorithm starts with an empty graph and iteratively performs local moves – arc additions, deletions or reversals - which most improves the graph's score. It terminates when the score cannot be improved, and this generally means at a local maximum.

x. **TABU**: is an adaptation of the HC algorithm described above which attempts to escape from local maxima by allowing some lower-scoring local moves. TABU also avoids revisiting DAGs encountered before, encouraging the algorithm to explore new regions of the DAG space [47].

xi. **H2PC**: a hybrid algorithm that combines HPC and HC [13]. HPC is an ensemble constraint-based algorithm which comprises three weak parents and children set learners (PC learner) in an attempt to produce a stronger PC learner.

xii. **SaiyanH**: a hybrid learning algorithm that incorporates restrictions in the search space of DAGs that force the algorithm to return a graph that enables full propagation of evidence, under the controversial assumption that the data variables are dependent [24]; i.e., every variable is dependent on at least one other variable in the data. It involves three learning phases: a) determining the skeleton of the initial graph using local learning, b) orientating the edges of the skeleton graph using constraint-based learning, and c) further modifying the graph using score-based learning.

xiii. **ILP**: an integer linear programming score-based algorithm that separates structure learning into two phases. The first phase computes the scores of candidate parent sets, whereas the second phase involves the optimal assignment of parents to each node [47]. ILP guarantees to return the graph that maximises a scoring function (i.e., it is an exact learning approach). However, this guarantee assumes unlimited runtime and no restriction on the maximum in-degree. In practise, exact learning can only be achieved for graphs that consist of up to approximately 30 nodes; although computational speed is also dependent on the sample size of the data. In this study, ILP is applied with runtime restrictions. Therefore, an exact result can only be guaranteed when the algorithm completes learning within a specified runtime limit, which we later discuss in Section 6.

xiv. **WINASOBS**: is similar to ILP in the sense that it relies on the same two learning phases; i.e., identification of a list of candidate parent sets for each node followed by an optimal assignment of the parent set of each node [26]. The difference from ILP is that it uses a simplified Bayesian Information Criterion (BIC) score that is easier to compute, referred to as BIC* (although in principle it can work with any scoring criterion like all other score-based algorithms), during the pre-processing phase to decide which local scores to compute and uses more aggressive pruning on the search space of possible DAGs that does not guarantee to return the graph that maximises BIC*. As a result, WINASOBS is an approximate, rather than exact, learning algorithm, but one which is scalable to thousands of nodes.

xv. **NOTEARS**: a score-based algorithm that converts the traditional score-based combinatorial optimisation problem (i.e., searching over possible graphs) into an equality-constrained problem that uses an equation composed by the weight matrix of the graph for optimisation [49]. This algorithm was originally designed for continuous data but has now been applied to ordinal discrete data.

The R package *r-causal* v1.1.1, which makes use of the Tetrad freeware implementation [50], was used to test algorithms $i$ to $v$. The *bnlearn* R package version 4.5 [33] was used to test algorithms $vi$ to $xi$. Lastly, SaiyanH was tested using the Bayesys v1.4 software [51], ILP was tested using the GOBNILP software [48], WINASOBS was tested using BLIP [52], and NOTEARS was tested using its original Python source code [49].

Structure learning algorithms enable users to modify some of their hyperparameters, such as the maximum node in-degree, the $p$-value threshold used in conditional independence tests, and the scoring metric used in search – typically either the BIC [53] or Bayesian Dirichlet equivalent uniform (BDeu) [54]. Modifying these hyperparameters may decrease or increase the learning performance of an algorithm in a particular experiment. However, it is impractical to test each of these algorithms over different learning hyperparameters in this paper. As a result, we have tested all algorithms with their default hyperparameter settings as implemented in each software, on the basis that is how most users would use them, and to keep the comparison between algorithms as fair as possible. These hyperparameter defaults are listed in Table 4.

**Table 4.** The default hyperparameters for each algorithm, where $a$ is the dependency threshold [with dependency score], MID is the max in-degree, $iss$ is the imaginary sample size (also known as equivalent sample size) for BDeu, and $|V|$ is the size of variable set $V$. Moreover, learning classes C, S and H represent constraint-based, score-based, and hybrid-based learning.

| Algorithm | Learning class | Software | Default hyperparameters |
|---|---|---|---|
| PC-Stable | C | Tetrad | $a = 0.01[G^2]$ |
| FGES | S | Tetrad | MID = 4, Greedy search, score BDeu with $iss = 1$ |
| FCI | C | Tetrad | $a = 0.01[Chi^2]$ |
| GFCI | H | Tetrad | $a = 0.01[Chi^2]$, Greedy search, score BDeu with $iss = 1$ |
| RFCI-BSC | H | Tetrad | $a = 0.01[Chi^2]$, sepset $a = 0.5$, bootstrap 50, lower/upper bounds 0.3/0.7, models search 10 |
| Inter-IAMB | C | bnlean | $a = 0.05[\text{MI}]$ |
| MMHC | H | bnlean | $a = 0.05[\text{MI}]$, search HC, score BIC |
| GS | C | bnlean | $a = 0.05[\text{MI}]$, MID = $inf$ |
| HC | S | bnlean | random restarts 0 and iterations inf, MID = inf, score BIC |
| TABU | S | bnlean | Tabu list 10, escapes 10 and iterations inf, MID = inf, score BIC |
| H2PC | H | bnlean | $a = 0.05[\text{MI}]$, search HC, score BDeu with $iss = 1$ |
| SaiyanH | H | Bayesys | $a = 0.05[MMD]$, indep. threshold 50%, Tabu escapes $\|V\|(\|V\| - 1)$, score BIC |
| ILP | S | GOBNILP | MID = 3, score BDeu with $iss = 1$. |
| NOTEARS | S | Source code | loss function = $l2$, regularisation = $l1$ |
| WINASOBS | S | BLIP | MID = 6, score BIC*, runtime search limit = 10s. |

## 5. EVALUATION

The algorithms are evaluated in terms of how well their learned graphs predict the true graphs, rather than how well the fitted distributions agree with the empirical distributions. This means that the evaluation is fully orientated towards graphical discovery rather than inference, which is also the preferred method in the structure learning literature when the aim is to assess the capability of an algorithm in terms of discovering causal structure. The scoring metrics used for this purpose are covered in subsection 5.1, whereas subsection 5.2 describes the approach we followed to assess the accuracy of learned graphs when based on data that incorporate latent variables; also known as structure learning under causal insufficiency. Lastly, subsection 5.3 describes the approach we followed to assess the time complexity of the algorithms.

### *5.1. The scoring metrics*

Three different scoring metrics are considered to assess the accuracy of a learned graph with respect to the true graph. These are:

i. the F1 score which represents the harmonic mean of Recall and Precision, which are themselves the most used metrics in this field of research. F1 is defined as:

$$\text{F1} = 2\frac{rp}{r+p}$$

where $r$ is Recall and $p$ is Precision. The F1 score ranges from 0 to 1, where 1 represents the highest score (with perfect Precision and Recall) and 0 the lowest.

ii. the Structural Hamming Distance (SHD) metric that penalises each change required to transform the learned graph into the ground truth graph [23]. The SHD score has become well-established in this field of research, partly due to its simplicity. However, it is important to highlight that the SHD score represents classic *classification accuracy*, which is often considered misleading. For example, consider a true graph with 1% dependencies (e.g., 100 edges) and 99% independencies (e.g., 9,900 independencies). An algorithm that returns an empty graph would be judged as being 99% accurate by SHD (i.e., an error of 100 out of possible 10,000 errors), despite the empty graph being useless in practice. Tsamardinos et al. [23], who proposed the SHD, acknowledged that it is biased towards the sensitivity of identifying edges versus specificity.

iii. the Balanced Scoring Function (BSF), which is a recent metric that addresses the bias in the SHD score. It eliminates this bias by taking into consideration all of the confusion matrix parameters (i.e., TP, TN, FP, and FN) to balance the score between dependencies and independencies [55]. Specifically,

$$\text{BSF} = 0.5\left(\frac{\text{TP}}{a} + \frac{\text{TN}}{i} - \frac{\text{FP}}{i} - \frac{\text{FN}}{a}\right)$$

where $a$ is the numbers of edges and $i$ is the number of independencies in the true graph, where $i$ can be calculated as $= \frac{|V|(|V|-1)}{2} - a$ where $|V|$ is the size of variable set $V$. The BSF score ranges from -1 to 1, where -1 corresponds to the worst possible graph, 1 to the graph that matches the true graph, whereas a score of 0 corresponds to an empty or a fully connected graph. This means that, in the example used above to highlight the bias in SHD, the empty graph would have returned a BSF score of 0, which BSF assumes as the baseline.

The above three metrics are interesting in their own way, and their differences are reflected in the results presented in Section 6; although F1 and BSF scores are largely in agreement.

Table 5 presents the penalty weights used to compute all the three scoring metrics. Note that rule #2 implies a minor change in the original definition of SHD. For example, in [23] a score of 10 indicates that the learned graph requires 10 changes – deletions, additions, or reversals - before it matches the true graph. In this paper, however, reversing an arc carries half the penalty of removing or adding an arc, under the assumption that the dependency has been correctly discovered, although with the wrong orientation, and this assumption is consistent with other SHD variants [56].

**Table 5.** The penalty weights used by the three scoring metrics, where o-o and o→ represent edges produced by structure learning algorithms designed for causally insufficient systems (see Section 5.2) and which indicate that the orientation is uncertain.

| Rule | True graph | Learned graph | Penalty | Reasoning |
|---|---|---|---|---|
| 1 | A → B | A → B, A o→ B | 0 | Complete match |
| 2 | A → B | A ↔ B, A − B , Ao−oB, A ← B, A←o B | 0.5 | Partial match |
| 3 | A → B | A ⊥ B | 1 | No match |
| 4 | A ↔ B | Any edge/arc | 0 | Latent confounder |
| 5 | A ⊥ B | A ⊥ B | 0 | Complete match |
| 6 | A ⊥ B | Any edge/arc | 1 | Incorrect dependency discovered |

### *5.2. Predicting the true causal graph, under causal sufficiency and causal insufficiency*

What the structure learning algorithms can and cannot discover is heavily debated in the literature [4] [5] [6] [7]. Further to what has been discussed in Introduction, most of the score-based and hybrid algorithms are implemented with score-equivalent functions such as BIC and BDeu. Score-equivalent functions cannot differentiate between two Markov equivalent DAGs. This means that while some of these algorithms will often return a DAG, in practice what they return is a DAG of the highest scoring CPDAG, and the preferred Markov equivalent DAG is sometimes sensitive to the order of the variables as they appear in the data (e.g., for the HC and TABU algorithm). Because of this, it can be argued that it would be more appropriate to compare these algorithms in terms of their CPDAG, rather than their DAG, score. However, the CPDAG score creates what we consider to be a more serious issue in that it leads to an overestimation of the ability of the algorithms to recover the true causal DAG which entails more information than its corresponding CPDAG. For example, if we consider the CPDAG scores, then algorithms that return a DAG that is part of the Markov equivalence class of the true DAG will receive a perfect score. However, in the real world, the Markov equivalence DAGs of the true DAG are incorrect and will inevitably lead to erroneous decisions about intervention. Since this paper investigates the usefulness of these algorithms in real-world settings where we tend to require CBNs for intervention, the assessment is driven by how well the algorithms achieve this objective (i.e., their ability to recover the true causal DAG), rather than driven by what some of the algorithms, or implementations of the algorithms, can and cannot do.

When it comes to causally insufficient experiments (i.e., those which incorporate latent variables), the learned graphs are assessed with respect to the ground truth Maximal Ancestral Graph (MAG). A MAG is an extension of a DAG that represents a set of observed variables in a DAG, where variables which are not part of that set are assumed to be latent. A MAG includes both directed and bi-directed (i.e., $\leftrightarrow$) edges, where a bi-directed edge indicates a latent confounder between connected nodes, whereas a directed edge represents direct or ancestral relationships. While a bi-directed arc entails more information than a directed arc, we assume that if $A \leftrightarrow B$ is in the true MAG then both $A \rightarrow B$ and $A \leftarrow B$ are acceptable as arcs in the CBN (refer to rule #4 in Table 5).

Moreover, a Partial Ancestral Graph (PAG) represents a set of Markov equivalent MAGs under causal insufficiency, in the same way a CPDAG represents a set of Markov equivalent DAGs under causal sufficiency. Some latent variable algorithms produce a PAG in the same way that some algorithms that do not account for latent variables produce a CPDAG. The scoring system specified in Table 5 allows us to compare the learned graphs (DAG, CPDAG, MAGs, or PAG) with either the true DAG or the true MAG. Moreover, Table 6 indicates which experiments are based on the true DAGs and which experiments are based on specific MAGs. Appendix B provides an example of a DAG being converted into a MAG, where Fig B1 shows the true DAG of Alarm, with its corresponding MAG5 depicted in Fig B2, and where approximately 5% of the variables in Fig B1 are missing in Fig B2 (i.e., they are latent).

**Table 6.** The ground truth graphs used in each experiment, where DAG is the true graph, MAG5 is the true graph with 5% latent variables, and MAG10 is the true graph with 10% latent variables. Experiments not performed are indicated as ‘n/a’ (refer to Table A1).

| | | Case study | | | | | |
|---|---|---|---|---|---|---|---|
| Experiment | Experiment code | Asia | Alarm | Pathfinder | Formed | Sports | Property |
| 1 | N | DAG | DAG | DAG | DAG | DAG | DAG |
| 2 | M5 | DAG | DAG | DAG | DAG | DAG | DAG |
| 3 | M10 | DAG | DAG | DAG | DAG | DAG | DAG |
| 4 | I5 | DAG | DAG | DAG | DAG | DAG | DAG |
| 5 | I10 | DAG | DAG | DAG | DAG | DAG | DAG |
| 6 | S5 | n/a | DAG | DAG | DAG | n/a | DAG |
| 7 | S10 | n/a | DAG | DAG | DAG | DAG | DAG |
| 8 | L5 | n/a | MAG5 | MAG5 | MAG5 | n/a | MAG5 |
| 9 | L10 | MAG10 | MAG10 | MAG10 | MAG10 | MAG10 | MAG10 |
| 10 | cMI | DAG | DAG | DAG | DAG | DAG | DAG |
| 11 | cMS | n/a | DAG | DAG | DAG | DAG | DAG |
| 12 | cML | MAG10 | MAG5 | MAG5 | MAG5 | MAG10 | MAG5 |
| 13 | cIS | n/a | DAG | DAG | DAG | DAG | DAG |
| 14 | cIL | MAG10 | MAG5 | MAG5 | MAG5 | MAG10 | MAG5 |
| 15 | cSL | n/a | MAG5 | MAG5 | MAG5 | DAG | MAG5 |
| 16 | cMISL | MAG10 | MAG5 | MAG5 | MAG5 | MAG10 | MAG5 |

### *5.3. Measuring time complexity*

Time complexity is measured by means of elapsed time (runtime). Due to the scale of the experiments, this study involved different members of the research team running the structure learning algorithms on different machines over multiple months. Since some machines are faster than others, we had to adjust structure learning runtimes for CPU speed.

We have selected the CPU with the highest market share as the baseline CPU to which all runtimes are adjusted. As of 31st of January 2020, the CPU[5] with the highest market share of 3.4% is the AMD Ryzen 5 3600 [57]. Table 7 lists all the machines used by the research team along with the adjusted runtime against the baseline CPU, where the adjustment is determined by the difference in single-core performance scores published by UserBenchmark [58]. Our tests showed that the algorithms were using only one of the available CPU cores, which is why runtime is adjusted relative to the single-core performance difference between CPUs. Therefore, our runtime results approximate elapsed time for structure learning on the AMD Ryzen 5 3600 Desktop CPU.

[5] The baseline CPU was not used in our experiments.

**Table 7.** Structure learning runtime adjusted for the baseline Desktop CPU AMD Ryzen 5 3600. The runtime adjustments are based on single-core CPU benchmarks retrieved from UserBenchmark [58]. RAM and the algorithms tested on each computing platform are also shown.

| CPU | CPU class | RAM | Single-core benchmark score | Runtime adjustment | Algorithm/s tested |
|---|---|---|---|---|---|
| Intel Core i9 9900K | Desktop | 32GB | 144 | 110.77% | SaiyanH |
| AMD Ryzen 5 3600 | Desktop | n/a | 130 | 100% | n/a |
| Intel Core i7 8750H | Mobile | 16GB | 110 | 84.62% | ILP, WINASOBS |
| Intel Core i7 6700 | Desktop | 32GB | 108 | 83.08% | SaiyanH |
| Intel i7 4770S | Desktop | 16GB | 103 | 79.23% | H2PC, Inter-IAMB |
| Intel Core i5 7360U | Mobile | 8GB | 99.1 | 76.23% | FCI, GFCI, MMHC, RFCI-BSC |
| Intel Core i7 8550U | Mobile | 16GB | 98.5 | 75.77% | SaiyanH |
| Intel Core i5 8250U | Mobile | 8GB | 92.4 | 71.08% | FCI, GFCI, MMHC, RFCI-BSC |
| Intel Core i7 6500U | Mobile | 8GB | 80.2 | 61.69% | HC, TABU, GS |
| Intel Core i5 5350U | Mobile | 8GB | 75.8 | 58.31% | PC-Stable |

## 6. RESULTS AND DISCUSSION

Because the results are based on almost 10,000 individual runs, we had to restrict the learning time per run to six hours. Algorithms that fail to return a result within the 6-hour limit are assigned the lowest rank for that graph. For example, if five out of the 15 algorithms fail to return a graph in a test, then all those five algorithms will receive rank 11 for that particular test. The ILP algorithm represents a 'relaxed' exception to this rule, and this is because it provides the option for the user to stop the learning process at any point in time and retrieve the 'best' graph discovered up to that point. We have taken advantage of this feature in ILP, which is not offered by other algorithms, and this can be argued as an unfair advantage for ILP. Furthermore, algorithms that fail to generate a graph due to a known or unknown error are also assigned the lowest rank for that graph. Detailed information on structure learning failures is provided in Appendix C for all algorithms and over all the experiments. Lastly, the results for NOTEARS are restricted to the Sports case study. This is because NOTEARS works only with binomial data (ordinal scale distributions), something which only the Sports case study conforms to in this paper.

The results are separated into those with no synthetic noise (i.e., experiment N) which represent the typical approach to synthetic data used in this field of research, and those with different types of synthetic noise (i.e., all other experiments) which represent the new approach used in this study in an effort to better approximate real-world performance. The subsections 6.1 and 6.2 cover these two approaches in turn. The overall results, both with and without data noise, are summarised in subsection 6.3.

### *6.1. Results without data noise*

We start by presenting the overall performance of the algorithms across all case studies and sample sizes associated with experiment N (i.e., no noise). Table 8 reports the average rank achieved by each algorithm, as determined by each of the three scoring metrics, averaged over all case studies and sample sizes in N. The reason we report the overall ranks rather than the overall scores is because not all algorithms produce a result in every single experiment. Averaging across scores that include missing scores would have biased the difference between scores, whereas assigning ranks (as indicated at the start of Section 6) avoids this bias.

Still, actual scores are needed to understand the difference in performance between two ranks and hence, we provide all scores generated by each algorithm for each experiment in Tables D1, D2, and D3, according to metrics F1, SHD and BSF respectively. Additionally, we provide the Precision and Recall scores, which are used to compute the F1 score, in Tables D4 and D5 respectively. These tables illustrate that the F1 and BSF metrics show similar patterns of results across all experiments, whereas SHD shows a much more homogeneous pattern within each case study. Because SHD captures the error between two graphs, it unsurprisingly shows relatively good performance for all algorithms across all sample sizes for small graphs (e.g., Asia and Sports) and relatively poor performance for the largest graphs (e.g., Formed and Pathfinder). F1 and BSF show more dependence on sample size within each case study – it being generally the case that performance improves with sample size. Across all three metrics, all algorithms and case studies it is usually the case that the best performance is found with either 100K or 1000K samples. It is also clear that all metrics and all algorithms show considerably worse performance on the Pathfinder case study than the other case studies. This can be partly explained by a variable in Pathfinder which consists of 63 states and serves as the parent for multiple other nodes, and this disproportionately increases the dimensionality of the model.

**Table 8.** Average and overall ranked performance of the algorithms over all case studies and sample sizes in experiment N (i.e., no noise), as determined by each of the three metrics, where Rank STD is the population standard deviation over all ranks achieved per algorithm.

| | F1 | | | SHD | | | BSF | | |
|---|---|---|---|---|---|---|---|---|---|
| Algorithm | Average rank | Rank STD | Overall rank | Average rank | Rank STD | Overall rank | Average rank | Rank STD | Overall rank |
| FCI | 7.7 | 3.92 | 9th | 6.57 | 3.96 | 7th | 7.67 | 3.45 | 9th |
| FGES | 7.5 | 2.38 | 8th | 7.83 | 2.84 | 10th | 7.1 | 2.62 | 8th |
| GFCI | 6.87 | 2.28 | 7th | 6.87 | 2.50 | 9th | 6.97 | 2.30 | 6th |
| GS | 11.87 | 2.03 | 14th | 10.43 | 3.36 | 13th | 11.9 | 2.04 | 14th |
| H2PC | 6.13 | 3.79 | 5th | 5.1 | 4.00 | 3rd | 6.97 | 3.80 | 6th |
| HC | 3.63 | 3.72 | 2nd | 4.77 | 4.06 | 2nd | 3.17 | 2.83 | 2nd |
| ILP | 4.8 | 3.40 | 3rd | 6.43 | 4.55 | 5th | 4.13 | 3.48 | 3rd |
| Inter-IAMB | 10 | 2.94 | 12th | 8.6 | 3.31 | 12th | 10.43 | 2.70 | 12th |
| MMHC | 7.77 | 2.74 | 10th | 6.47 | 2.86 | 6th | 8.6 | 2.51 | 11th |
| NOTEARS | 12 | 4.00 | 15th | 13 | 2.00 | 15th | 12 | 4.00 | 15th |
| PC-Stable | 8.1 | 3.83 | 11th | 6.83 | 4.06 | 8th | 8 | 3.35 | 10th |
| RFCI-BSC | 11.5 | 2.73 | 13th | 10.9 | 3.56 | 14th | 11.47 | 2.80 | 13th |
| SaiyanH | 5.33 | 2.26 | 4th | 8 | 3.50 | 11th | 4.77 | 2.80 | 4th |
| TABU | 3.27 | 3.07 | 1st | 4.43 | 3.55 | 1st | 3.1 | 2.77 | 1st |
| WINASOBS | 6.3 | 3.85 | 6th | 5.87 | 3.58 | 4th | 6.17 | 3.54 | 5th |

The results in Table 8 show that metrics F1 and BSF are generally in agreement in ranking the algorithms in terms of overall performance, since the discrepancy in ranking between these two metrics is never greater than one. In contrast, the ranking produced by SHD is often different. For example, SHD ranks MMHC 6th while F1 and BSF rank it 10th and 11th respectively. Another major discrepancy between metrics involves the SaiyanH algorithm, which SHD ranks 11th whereas both F1 and BSF rank 4th.

Fig 1 demonstrates how rankings fluctuate when the experiments are ordered on the $n/p$ scale, where $n$ is the number of samples over $p$ free parameters, and provides an indication of the risk of model overfitting. For example, the experiment with the highest risk of overfitting is Pathfinder with sample size 0.1k (the leftmost point in each chart in Fig 1), whereas the experiment with the lowest risk of overfitting is Asia with sample size 1,000k (the rightmost point in each chart in Fig 1). In the former, the $n/p$ score is 0.0 since there are 100 samples divided by 71,890 free parameters, whereas in the latter the $n/p$ score is 55,556 since there are 1,000,000 samples divided by 18 free parameters.

The oscillatory behaviour observed in the graphs of Fig 1 suggests that the risk of overfitting of an experiment is not a strong predictor of the relative performance between algorithms. Figure D1 reorders the results of Fig 1 by the number of nodes, followed by sample size. Some weak patterns emerge under this ordering, such as a) the relative performance of MMHC increases with the number of nodes, b) the relative performance of H2PC improves with sample size within each case study, and c) the relative performance of ILP decreases with the number of nodes since higher nodes make the task of exact learning progressively harder, which needs to end at the 6-hour limit. However, Fig D1 continues to suggest that it is difficult to deduce a pattern that explains the changes in the relative performance between algorithms across the different experiments with confidence. Note that even TABU, which is ranked 1st overall by all three metrics, often shows significant drops in relative performance for some experiments. Furthermore, the graphs in Fig 1 continue to support the notion that F1 and BSF are largely in agreement, whereas SHD often deviates from F1 and BSF rankings. Noticeable examples include GS (and to some extent MMHC) whose results are ranked highly by SHD (left side of the graph), something which strongly contradicts the F1 and BSF rankings, and vice versa for SaiyanH and ILP.

Fig 2, on the other hand, plots the number of edges learned[6] with respect to the true number of edges, per algorithm per case study per sample size. It is noticeable how each of the charts in Fig 2 shows a broadly upwards trend in the number of edges as $n/p$ increases, which is a reasonable outcome given that a higher $n/p$ score decreases the risk of overfitting and enables the algorithms to draw edges with higher confidence. In contrast, a very low $n/p$ score increases the risk of underfitting for some algorithms. The contradictions between metrics previously observed in Fig 1 can be partly explained by the results in Fig 2. For example, in the case of GS, the number of edges generated is extremely low with respect to the true number of edges, especially on the low $n/p$ scores, something which SHD is known to favour since it tends to be biased in favour of empty/underfitted graphs, but which F1 and BSF rank poorly. In contrast, SaiyanH and ILP are the only algorithms that produce a relatively high number of edges for the low $n/p$ scores and, although they do much better than other algorithms in approximating the true number of edges, SHD penalises them heavily which, once more, highly contradicts the F1 and BSF rankings.

[6] Fig 2 excludes failed attempts by the algorithms to generate a graph (refer to Appendix C).

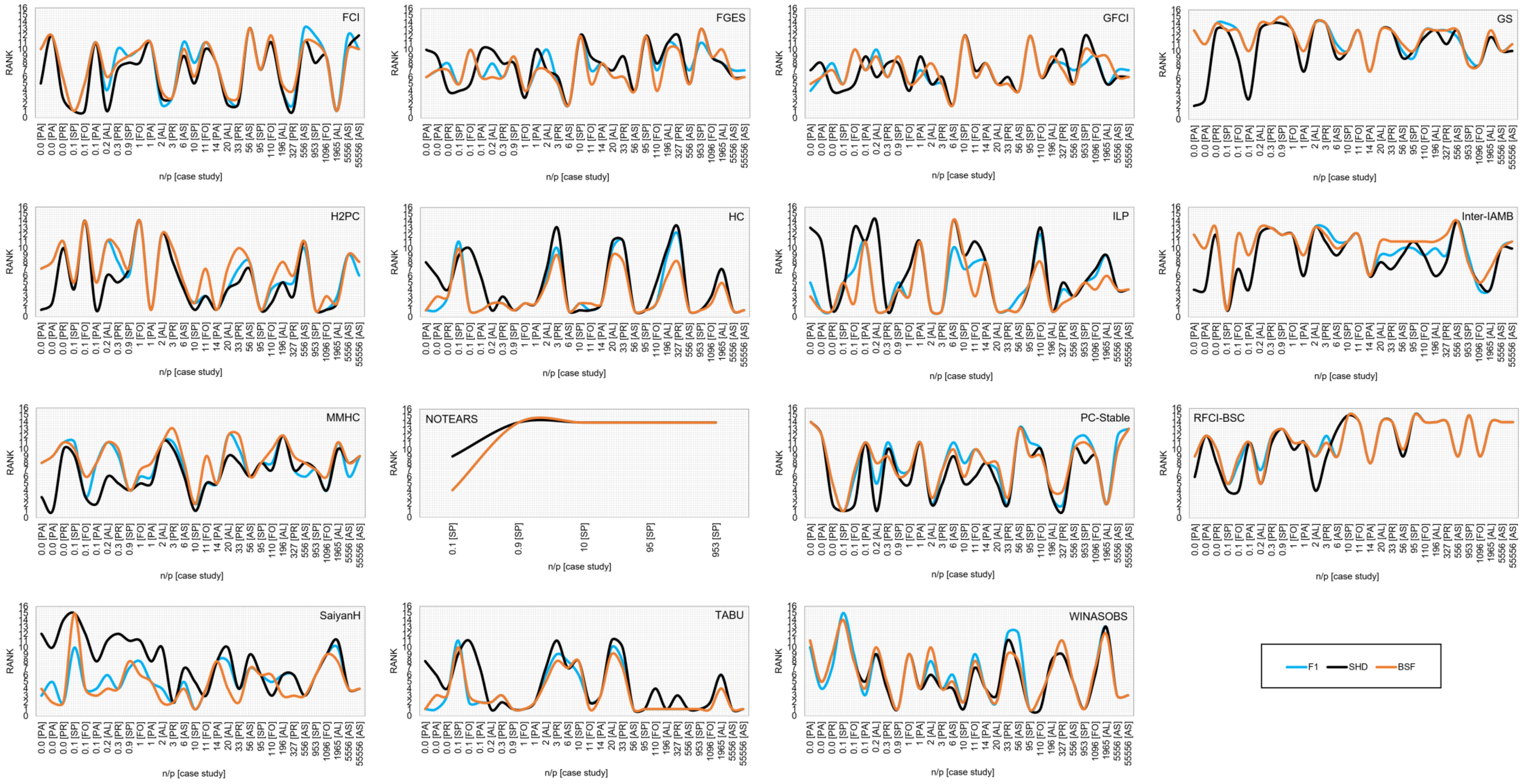

**Fig 1.** Overall ranking of the algorithms (as defined in Table 8) ordered on the $n/p$ scale of $n$ samples over $p$ parameters. These results are based on experiment N (i.e., no data noise). In NOTEARS, the F1 rankings are identical to the BSF rankings.

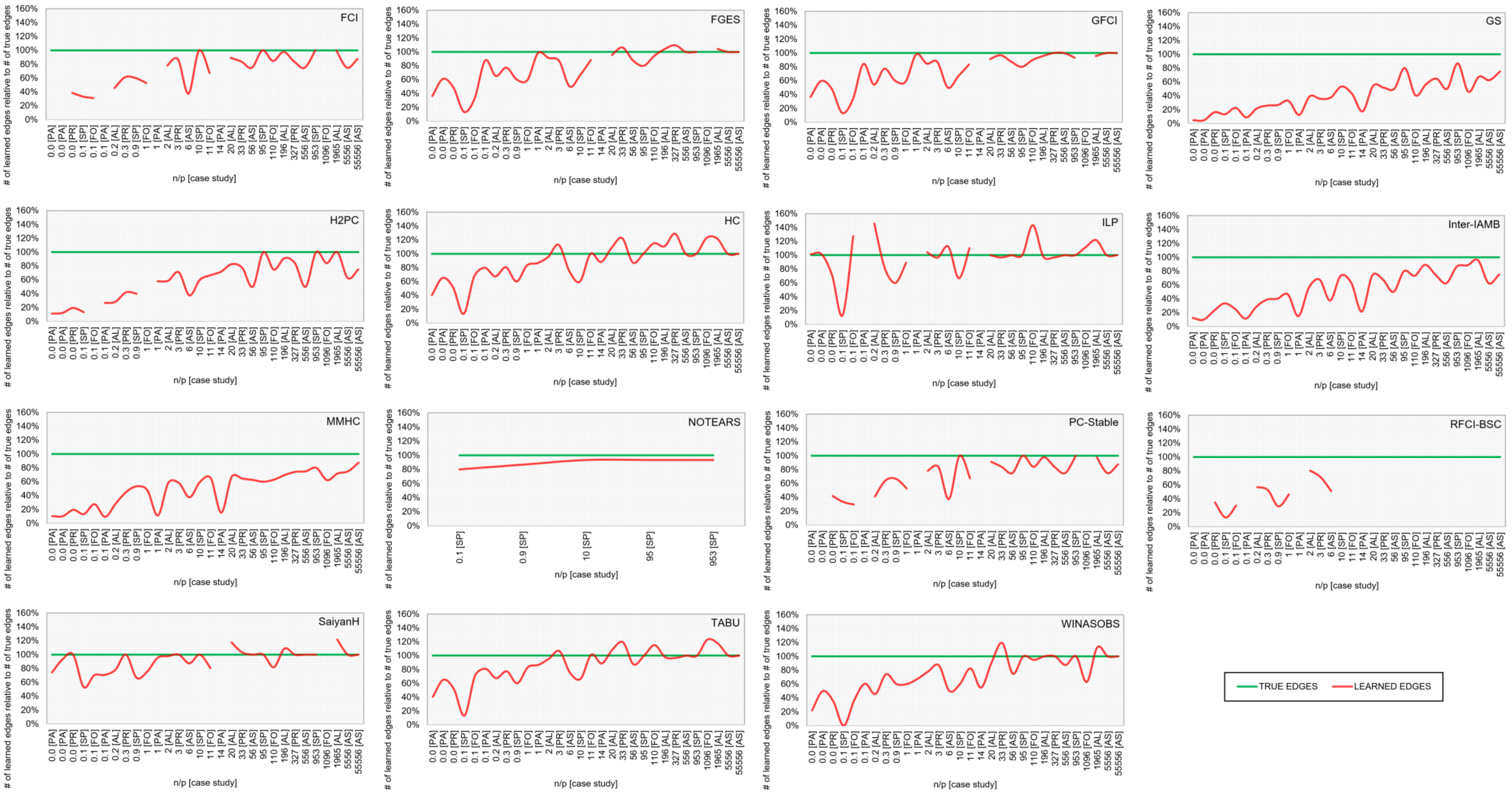

**Fig 2.** The number of learned edges with respect to the number of true edges, for each algorithm, ordered on the $n/p$ scale of $n$ samples over $p$ parameters. These results are based on experiment N (i.e., no data noise). Failed attempts by the algorithms to generate a graph (refer to Appendix C) are excluded and indicated with missing information on the learned edges (red line).

Moreover, Fig 2 shows that RFCI-BSC failed to generate a graph on all high $n/p$ score experiments, and this is likely due to the higher sample size datasets[7]. The results also suggest that GS and MMHC, and to a lesser extent, Inter-IAMB and H2PC, greatly underestimate the number of true edges and, while the underestimation improves with higher $n/p$ scores, these algorithms always underfit the learned graph. In contrast, TABU, HC, ILP, SaiyanH, and WINASOBS, which are the top five algorithms according to the F1 and BSF metrics, are the only algorithms that produce slightly more edges than the number of true edges.

Fig 3 presents the cumulative runtime for each algorithm over all runs in the N experiments. Note that the information presented in this figure has already been adjusted for the different processing powers of the computers used as indicated in Table 7. Moreover, each experiment that did not complete within the 6-hour limit, including failures to produce a graph for other reasons (refer to Appendix C), is assigned the 6-hour limit as runtime for that particular experiment. This means that Fig 3 (as well as Fig 4 shown later) underestimates runtime when an algorithm does not produce a graph within the 6-hour limit, and underestimates or overestimates runtime when an algorithm returns an out-of-memory or an unknown error during the structure learning process. Still, we consider these results to be reasonably accurate. However, one estimate we would like to highlight as being highly uncertain is that of RFCI-BSC (and to a lower extent FCI after adding noise to the data), and this is because this algorithm produced a very high number of failures (refer to Appendix C), each of which is penalised with the 6-hour runtime limit. Lastly, it is worth highlighting the efficiency of the algorithms implemented in the *bnlearn* R package [33]. Specifically, GS, HC, Inter-IAMB, MMHC and TABU, all of which have been tested using *bnlearn*, proved to be considerably faster than other algorithms and never failed to produce a graph; although this observation can also be partly explained by the algorithms themselves.

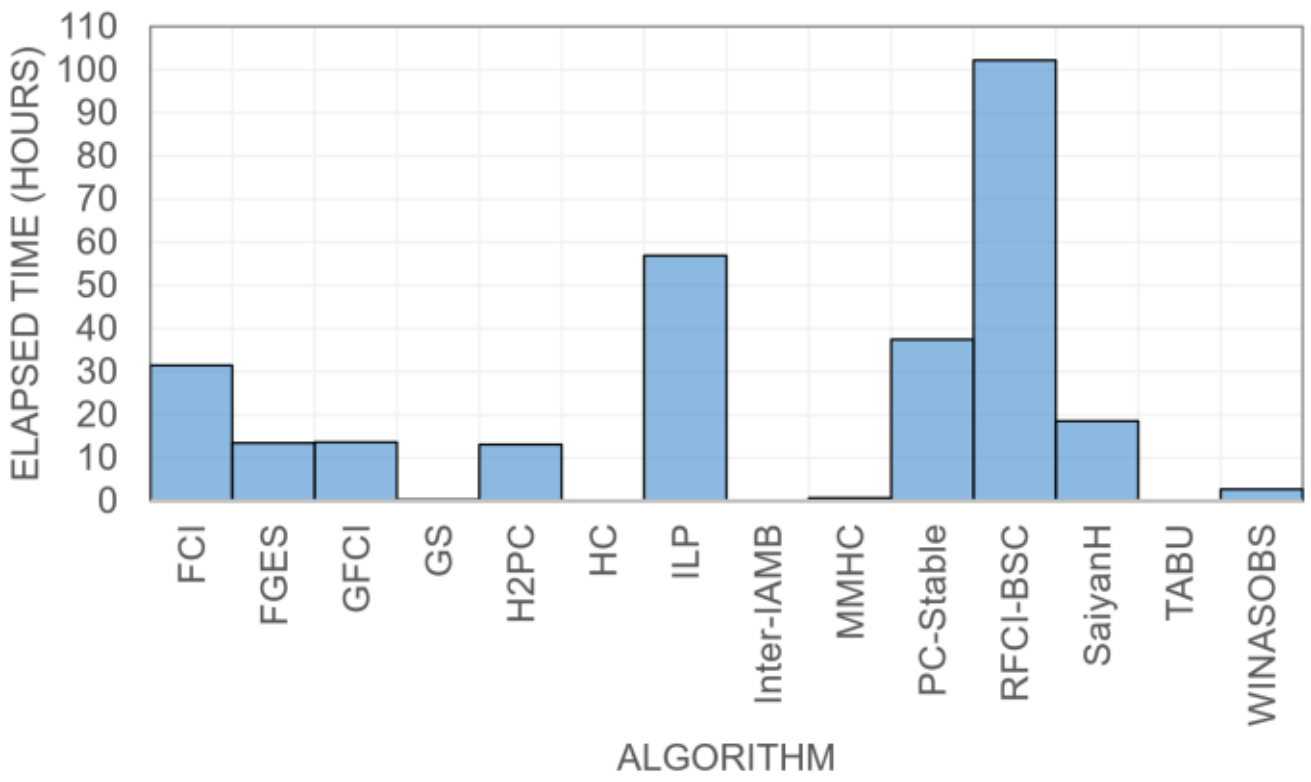

**Fig 3.** The cumulative runtime of the algorithms over all experiments in N. The runtime is adjusted as indicated in Table 7. Failed attempts by the algorithms to generate a graph (refer to Appendix C) are assigned the 6-hour runtime limit. Note that the results exclude NOTEARS since it was applied only to the Sports case study, on which its runtime ranked 13$^{th}$ out of the 15 algorithms.

[7] The likely cause of this may be the bootstrap sampling RFCI-BSC performs for model averaging.

### 6.2. Results with data noise

Table 9 repeats the analysis of Table 8 on the 15 noisy experiments, which are directly compared to the results shown in Table 8 (i.e., the N experiment). Specifically, the analysis highlights how the relative performance of the algorithms has changed after incorporating noise in the data, as determined by each of the metrics.

The results reveal numerous interesting observations. TABU, whose previous overall performance topped all three rankings, has lost significant ground against other algorithms and now ranks 2nd overall by the F1 and BSF metrics, and 4th overall by SHD. In contrast, the HC algorithm, which all metrics previously ranked 2nd, is now ranked 1st by F1 and BSF, and 2nd by SHD. This is an interesting observation because TABU is an improved search version of HC that escapes some of the suboptimal search regions in which HC has the tendency to get stuck in. This result can only suggest that data noise has misled TABU into performing escapes from a local maximum into regions that further deviate from the true graph. However, this observation does not extend to SaiyanH whose score-based learning phase also makes use of tabu search; although the tabu search in SaiyanH plays a less significant role compared to the tabu search in TABU.

The ILP algorithm, which is the only exact learning algorithm tested in this study, has lost some ground in relative performance but not enough to alter its ranking. This result is consistent with that of TABU on the basis that data noise appears to distort model fitting which in turn has a negative effect on algorithms that maximise exploration on curve fitting. This observation provides empirical evidence that while exact learning is expected to work best in theory, which assumes no data noise, it does not work best in practise.

Interestingly, MMHC is the only algorithm with significant gains in performance across all the three metrics. Specifically, it now ranks 6th, 1st, and 9th by F1, SHD, and BSF metrics respectively, up from 10th, 6th, and 11th under no data noise. While MMHC claimed the top spot in terms of SHD under data noise, this result can be largely explained by the low number of edges (i.e., underfitting) the algorithm tends to produce with respect to the number of true edges (refer to Fig 2), which also explains the contradiction with the F1 and BSF rankings. On the other hand, FCI is the algorithm with the highest loss in relative performance. Another interesting contradiction between metrics can be observed in the results of PC-Stable, where F1 and BSF suggest that PC-Stable has improved its performance under data noise, and relative to the other algorithms, whereas the SHD metric suggests otherwise; and this is a good example that demonstrates how classic classification accuracy (i.e., SHD) and balanced accuracy (i.e., BSF and partly F1) can lead to very different conclusions.

Fig 5 presents the increase or decrease in relative ranking between algorithms, for each of the 15 noisy experiments and with respect to experiment N. These results reveal further inconsistencies between metrics, some of which are highly contradictory. For example, the F1 and BSF results on GFCI, HC, and PC-Stable, are in direct disagreement with SHD. Interestingly, the algorithms designed to account for latent variables during structure learning, such as the FCI, GFCI and RFCI-BSC, did not improve their performance relative to other algorithms under experiments which involve the reconstruction of the true MAG (experiments which incorporate code ‘L’).

Fig 4 illustrates the cumulative runtime of the algorithms over all the 15 data noisy experiments. Note that while data noise has influenced the structure learning runtime of algorithms in different ways, the overall result shows that runtime has increased by 3.2% after adding noise to the data. While the results in Fig 4 are largely consistent with those in Fig 3, the results suggest that some algorithms can be considerably slower under data noise, such as FCI and H2PC, while others can be faster, such as ILP.

**Table 9.** Average and overall ranked performance for each algorithm over all case studies and sample sizes, and over all the 15 noisy-based experiments, as determined by each of the three metrics, where Δ is the relative difference in performance with respect to the N experiments (i.e., without noise). Green and red text indicate increased and decreased relative ranked performance respectively. Detailed performance for each of the 15 noisy experiments is provided in Appendix E.

| Algorithm | F1 Average rank | F1 Δ average rank | F1 Overall rank | F1 Δ overall rank | SHD Average rank | SHD Δ average rank | SHD Overall rank | SHD Δ overall rank | BSF Average rank | BSF Δ average rank | BSF Overall rank | BSF Δ overall rank |
|---|---|---|---|---|---|---|---|---|---|---|---|---|
| FCI | 8.67 | -0.97 | 11th | -2 | 8.67 | -2.11 | 12th | -5 | 8.23 | -0.56 | 11th | -2 |
| FGES | 7.15 | +0.35 | 8th | 0 | 7.12 | +0.71 | 8th | +2 | 7.37 | -0.27 | 8th | 0 |
| GFCI | 7.26 | -0.39 | 9th | -2 | 6.91 | -0.04 | 7th | +2 | 7.60 | -0.63 | 10th | -4 |
| GS | 11.74 | +0.12 | 15th | -1 | 9.54 | +0.89 | 13th | 0 | 11.68 | +0.22 | 14th | 0 |
| H2PC | 5.66 | +0.47 | 5th | 0 | 4.96 | +0.14 | 3rd | 0 | 6.26 | +0.71 | 5th | 1 |
| HC | 3.60 | +0.03 | 1st | +1 | 4.92 | -0.15 | 2nd | 0 | 3.03 | +0.13 | 1st | 1 |
| ILP | 5.17 | -0.37 | 3rd | 0 | 6.72 | -0.28 | 6th | -1 | 4.35 | -0.22 | 3rd | 0 |
| Inter-IAMB | 9.79 | +0.21 | 12th | 0 | 7.82 | +0.78 | 9th | +3 | 9.98 | +0.45 | 12th | 0 |
| MMHC | 6.51 | +1.26 | 6th | +4 | 4.66 | +1.81 | 1st | +5 | 7.59 | +1.01 | 9th | +2 |
| NOTEARS | 11.65 | +0.35 | 14th | +1 | 12.83 | +0.17 | 15th | 0 | 12.51 | -0.51 | 15th | 0 |
| PC-Stable | 7.59 | +0.51 | 10th | +1 | 7.87 | -1.03 | 10th | -2 | 7.15 | +0.85 | 7th | +3 |
| RFCI-BSC | 11.50 | 0.00 | 13th | 0 | 11.05 | -0.15 | 14th | 0 | 11.54 | -0.07 | 13th | 0 |
| SaiyanH | 5.27 | +0.06 | 4th | 0 | 7.87 | +0.13 | 11th | 0 | 5.16 | -0.40 | 4th | 0 |
| TABU | 3.62 | -0.35 | 2nd | -1 | 4.99 | -0.56 | 4th | -3 | 3.13 | -0.03 | 2nd | -1 |
| WINASOBS | 6.54 | -0.24 | 7th | -1 | 5.49 | +0.38 | 5th | -1 | 6.77 | -0.61 | 6th | -1 |

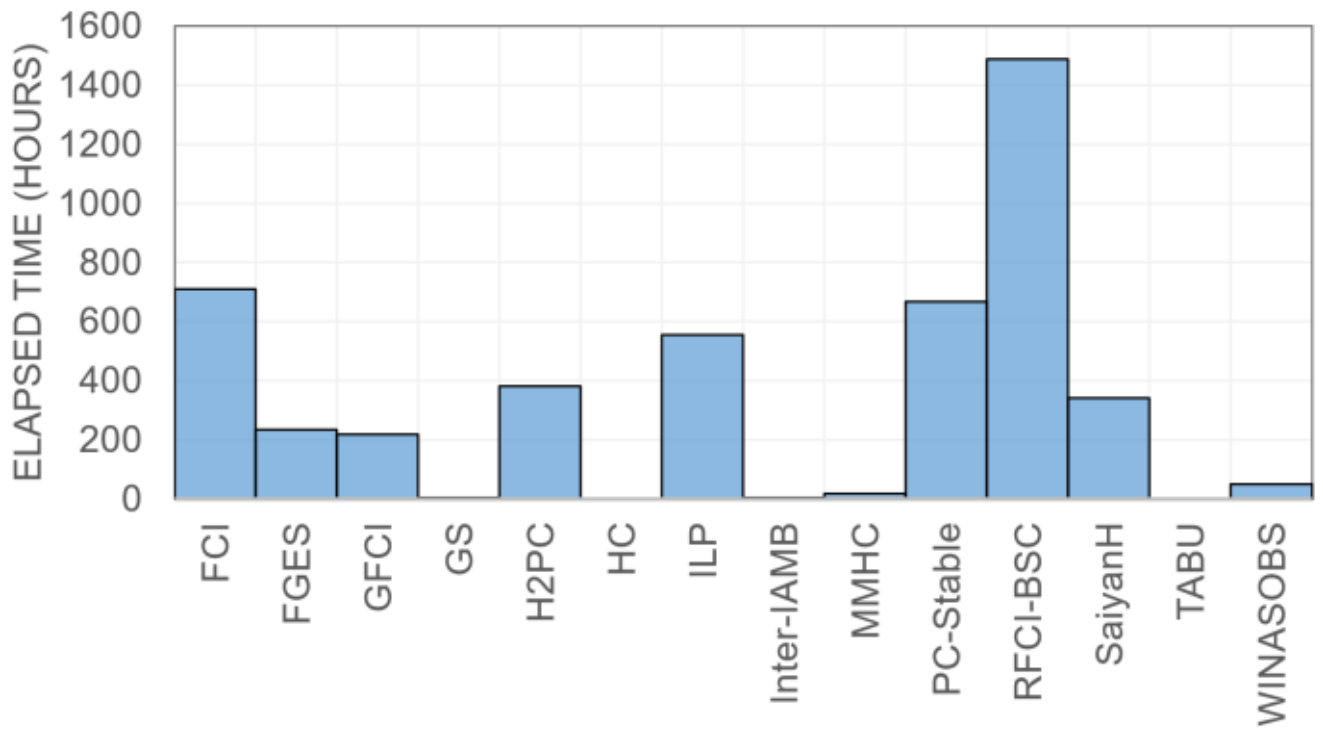

**Fig 4.** The cumulative runtime of the algorithms over all 15 noisy-based experiments. The runtime is adjusted as indicated in Table 7. Failed attempts by the algorithms to generate a graph (refer to Appendix C) are assigned the 6-hour runtime limit. Note that the results exclude NOTEARS since it was applied only to the Sports case study, on which its runtime ranked 9th out of the 15 algorithms.

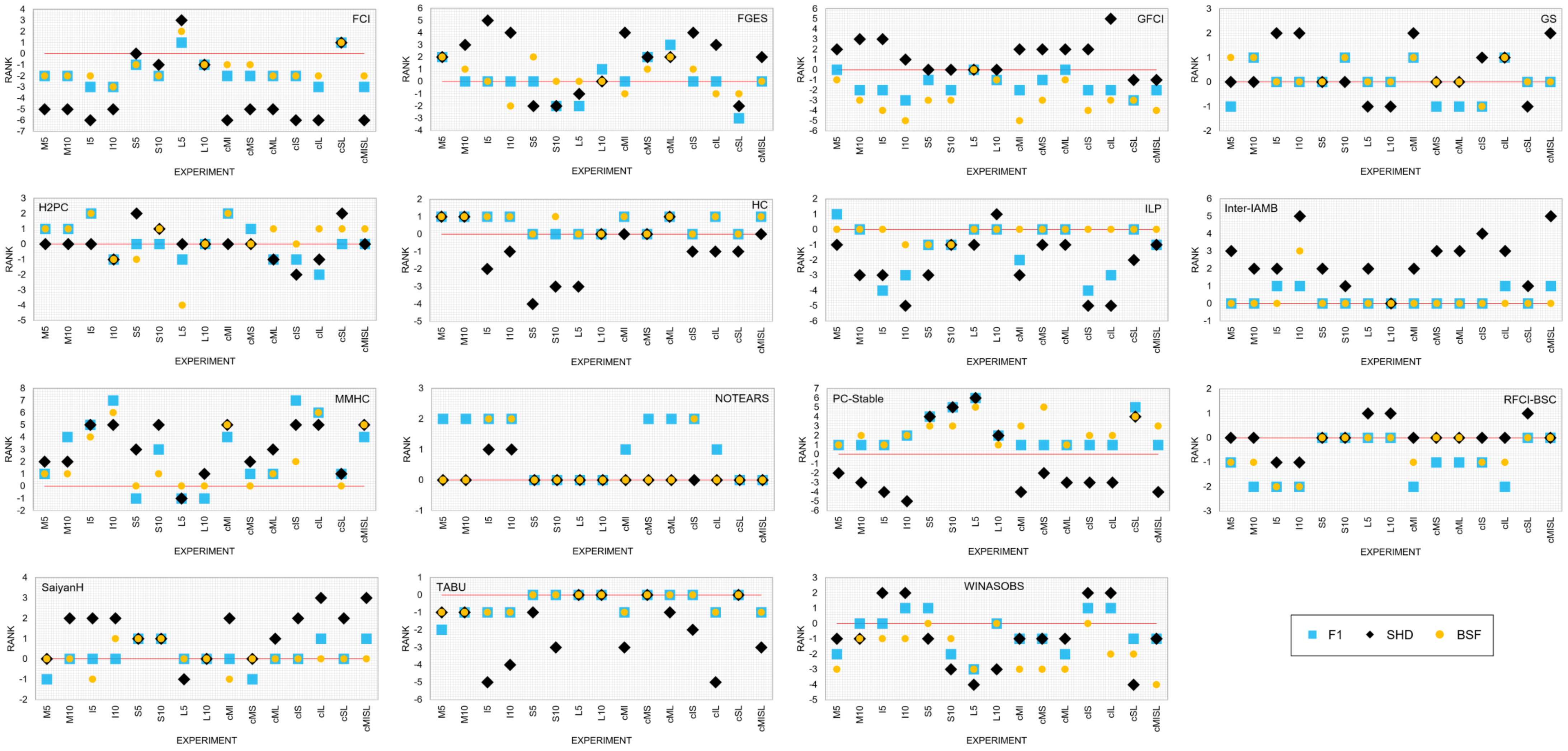

**Fig 5.** The difference in the overall rank achieved by each algorithm for each noisy experiment and over each of the scoring metrics, relative to experiment N (i.e., before adding noise to the data). The superimposed red line represents zero difference in the rankings between experiment N (without noise) and all other experiments (with noise), whereas markers above and below the line indicate increased and decreased performance after adding noise to the data.

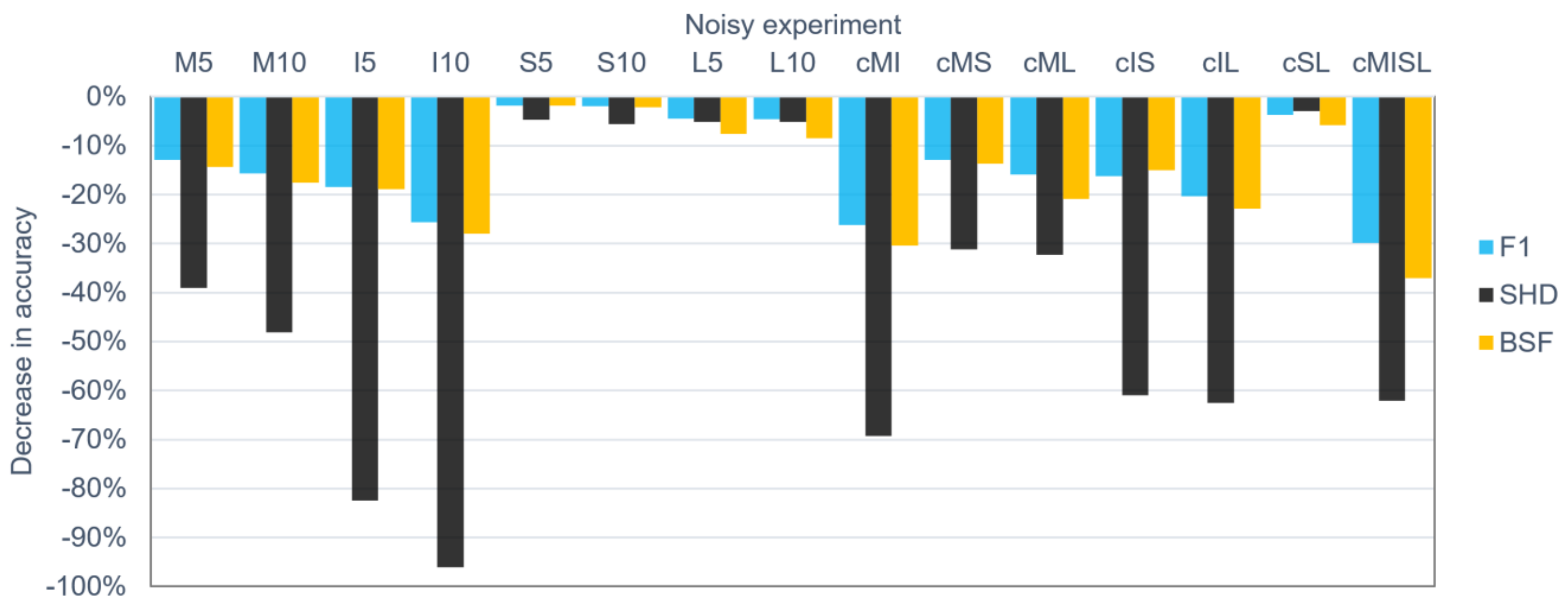

**Fig 6.** The overall decrease in accuracy (F1 and BSF), and increase in error (SHD), over all algorithms and for each type of noise added to the data.

Lastly, Fig 6 presents the overall decrease in accuracy, across all algorithms, for each noisy experiment and with respect to the experiment N (i.e., no data noise). The inconsistency in the conclusions between the F1 and BSF metrics, and with respect to the SHD, is also present in Fig 6, where the imbalance in the SHD score leads to the counterintuitive conclusion that experiments I5, I10, MI, and IL, have decreased structure learning performance more than experiment MISL which incorporates all types of noise and a higher total rate of data noise. On the other hand, the F1 and BSF metrics identify that MISL has had the largest negative impact on structure learning performance, as might be expected. Note that another counterintuitive conclusion, and one which applies to all metrics, is that I5 decreases overall performance larger than IS. However, this conclusion is uncertain and subject to bias because the minor difference between these two results could be explained by the IS experiments not including the Asia case study, unlike the I5 experiments (refer to Table A1).

According to the F1 and BSF metrics, the overall results suggest that data noise of types S and L have had a relatively minor impact on structure learning performance. However, it should be noted that the results from experiments that incorporate L are based on the reconstruction of the true MAGs which incorporate a lower number of variables compared to the true DAGs used in experiments that do not incorporate L, and the difference in the number of variables in the data can influence the result generated by some metrics. Specifically, the SHD error increases with the number of variables and this biases the comparisons between SHD scores that are based on true graphs with different number of variables. On the other hand, data noise of types M and I have had a much stronger impact on decreasing the performance of the algorithms, although this could be because noise of type $M$ and $I$ influence variable values across all variables, whereas noise of type $S$ and $L$ influences only some (up to 10%) of the variables and at the same time reduce model dimensionality. Combining all types of noise in a single dataset (i.e., experiment MISL), which is reasonable to assume better approximates real data, has led to a decrease in structure learning accuracy in the range of 30% to 37%; or an overestimation of real performance in the range of[8] 43% to 59%.

[8] If data noise reduces accuracy by roughly $\frac{1}{3}$, such as from 75% to 50%, then data with no noise increases accuracy by 50% (i.e., from 50% to 75%).

### 6.3. Summary of results with and without data noise

This subsection firstly discusses how the results obtained from noisy experiments compare to experiments that do not incorporate data noise, and then presents detailed performance of the algorithms that takes into consideration all of the experiments, both with and without data noise. Lastly, we summarise the strengths and weaknesses of the algorithms at the end of this subsection.

Focusing on the performance of each algorithm, the results reveal that score-based algorithms are superior to constraint-based algorithms, and this observation is in agreement with [32]. Specifically, the score-based HC and TABU algorithms claimed the top two spots for overall performance, both with and without data noise (although TABU ranked 4th in SHD under data noise). Interestingly, while TABU topped the traditional synthetic tests that did not incorporate data noise, HC was the algorithm that topped the noisy experiments. Both HC and TABU are based on the same implementation and differ only in search strategy, where TABU represents a theoretical improvement in search strategy over HC, by escaping some of the suboptimal graphs which HC fails to do so. However, empirical evidence suggest that data noise has misled TABU into escaping towards less accurate regions despite producing better fit compared to HC (at least in terms of BIC score). Equally interesting is the observation that the HC algorithm, which is the simplest and fastest structure learning algorithm, has outperformed all other, and indeed more sophisticated, algorithms under data noise. These results provide empirical evidence that a higher fitting score does not imply a more accurate causal graph, and support Pearl's view in that curve fitting alone is insufficient for causal discovery [2, 9].

ILP follows at 3rd position (although ranked 5th to 6th by SHD), both with and without data noise. It is important to reiterate that ILP is the only exact learning algorithm investigated in this study and, relative to other approximate algorithms, it requires extensive time to complete its search. As stated in Section 6, while structure learning time was restricted to six hours per run, we took advantage of GOBNILP's option to retrieve the best graph discovered at the end of the 6-hour time limit. This is not an option offered by other implementations and hence, an argument can be made that this decision provided an unfair advantage to ILP. However, since this paper focuses on assessing the usefulness of these algorithms in real-world settings, we considered GOBNILP's feature to be useful in the real world and have taken advantage of it in the experiments. This is a feature that, in theory, can be applied to any score-based method; in contrast to constraint-based methods that are generally (with exceptions [59]) unable to work in anytime fashion. Regardless, it is worth highlighting that even in cases where ILP completes exact learning before the time limit is reached, and specifically on networks with max in-degree 3 (its default hyperparameter) or lower, such as Asia, Sports, and Property, the highest scoring graph is only occasionally the most accurate graph across the algorithms tested. This is because while exact learning guarantees that the algorithm will find a DAG that maximises a score-equivalence function, there is no formal guarantee the learned DAG will be the DAG that maximises the F1, SHD or BSF metrics. Moreover, because we ran ILP with BDeu score and HC and TABU with BIC, some of the differences may be due to the scoring functions rather than the algorithms.

SaiyanH follows at 4th position, both with and without data noise (although ranked 11th by SHD) and claims the top hybrid learning spot according to the F1 and BSF metrics. H2PC and WINASOBS follow with overall ranking ranging between 5th to 6th and 5th to 7th respectively, with and without data noise (although ranked 3rd to 4th and 4th to 5th respectively by SHD). FGES and MMHC, both which are often assumed to be the top structure learning algorithms in this field of research, ranked 8th and 6th to 11th respectively, with and without data noise (although 8th to 10th and 1st to 6th by SHD). While MMHC gained impressive ground against all other algorithms under the noisy experiments, its performance proved to be the most unstable over all algorithms. Overall, our results on MMHC are not in agreement with the results in [23] and support the results in [32].

GFCI and FCI, which are often considered as the best algorithms for causally insufficient systems, ranked 6th to 7th and 11th respectively without data noise, and 6th to 7th and 9th respectively with data noise. While these algorithms have been specifically designed for causal structure learning with latent variables, our results provide no evidence that GFCI and FCI gain advantage over other algorithms under these conditions. On the other hand, FCI is the top constraint-based algorithm in this study, with PC-Stable closely behind. Lastly, Inter-IAMB, GS, RFCI-BSC and NOTEARS have produced disappointing results.

In the previous subsections 6.1 and 6.2, the decision to rank the algorithms in Tables 8 and 9 by their average rank achieved across experiments represents an imperfect approach that hides the actual difference in scoring performance between ranks. The reason we followed this approach, as opposed to ranking the algorithms based on their average metric score, is because the scores incorporate missing information due to failure by some algorithms to produce a graph in all the experiments (refer to Appendix C). Averaging across experiments that incorporate missing values will inevitably bias the scores, especially because failures are not random; i.e., they tend to occur in experiments that involve higher data sample sizes and which tend to produce more accurate scores.

To minimise the uncertainty caused by missing data and ranking, we provide detailed information on the discrepancy in scores between algorithms in Tables 10 and 11. Specifically, Table 10 presents the averaged minimum and maximum performance scores achieved by an algorithm for each case study and sample size combination, as determined by each of the three metrics, and over all algorithms and experiments (both with and without data noise), and Table 11 presents the overall performance of the algorithms for each case study and sample size combination relative to the minimum and maximum scores depicted in Table 10. For example, the minimum averaged F1 score is 0.23 for Asia under sample size 0.1k, because the lowest averaged score across all 15 algorithms averaged across all 16 experiments was 0.23 (in this case, by the GS algorithm). Note that in Table 11, F represents "*at least one failure*" (or at least one missing data point) which can bias the average and hence, the average is not reported for those cases. Similarly, the minimums and maximums in Table 10 are restricted to cases that do not incorporate missing data. These two tables are repeated for cases restricted to experiments $N$ (i.e., without noise - refer to Tables D6 and D7), and all the other 15 experiments with data noise (refer to Tables E6 and E7).

**Table 10.** The averaged minimum and maximum scores for each case study and sample size $n$ combination, as determined by each of the three metrics over all 15 algorithms, and averaged across all 16 experiments (both with and without data noise). Note that a higher SHD score indicates lower performance. The minimums and maximums are based on all the experiments shown in Table 11 that do not include a failure F.

| Min/Max performance | $n$ | F1 | | | | | | SHD | | | | | | BSF | | | | | |
|---|---|---|---|---|---|---|---|---|---|---|---|---|---|---|---|---|---|---|---|
| | | Asia | Spor | Prop | Alar | Form | Path | Asia | Spor | Prop | Alar | Form | Path | Asia | Spor | Prop | Alar | Form | Path |
| Min | 0.1k | 0.23 | 0.00 | 0.12 | 0.13 | 0.12 | 0.03 | 7.2 | 15.7 | 35.8 | 50.0 | 194.3 | 344.0 | 0.12 | -0.03 | 0.07 | 0.08 | 0.07 | 0.01 |
| | 1k | 0.43 | 0.19 | 0.25 | 0.26 | 0.24 | 0.03 | 5.7 | 13.2 | 26.2 | 40.6 | 120.4 | 272.0 | 0.30 | 0.10 | 0.16 | 0.17 | 0.15 | 0.01 |
| | 10k | 0.54 | 0.34 | 0.31 | 0.35 | 0.36 | 0.07 | 5.7 | 13.2 | 25.3 | 39.4 | 107.2 | 283.9 | 0.44 | 0.13 | 0.20 | 0.25 | 0.25 | 0.03 |
| | 100k | 0.51 | 0.34 | 0.41 | 0.42 | 0.35 | 0.10 | 7.0 | 13.2 | 23.3 | 41.0 | 125.0 | 312.2 | 0.37 | 0.13 | 0.30 | 0.33 | 0.25 | 0.06 |
| | 1000k | 0.48 | 0.34 | 0.48 | 0.47 | 0.38 | 0.13 | 9.2 | 13.6 | 28.5 | 44.2 | 151.9 | 200.8 | 0.29 | 0.13 | 0.38 | 0.39 | 0.28 | 0.07 |
| Max | 0.1k | 0.55 | 0.32 | 0.54 | 0.43 | 0.32 | 0.18 | 4.6 | 11.3 | 19.3 | 37.5 | 120.1 | 210.5 | 0.40 | 0.21 | 0.42 | 0.41 | 0.24 | 0.11 |
| | 1k | 0.70 | 0.72 | 0.73 | 0.73 | 0.57 | 0.22 | 3.6 | 6.3 | 12.0 | 19.8 | 88.2 | 205.6 | 0.58 | 0.56 | 0.63 | 0.68 | 0.48 | 0.17 |
| | 10k | 0.79 | 0.75 | 0.85 | 0.78 | 0.69 | 0.33 | 3.0 | 5.6 | 8.2 | 17.8 | 73.3 | 187.7 | 0.73 | 0.61 | 0.82 | 0.79 | 0.67 | 0.27 |
| | 100k | 0.80 | 0.99 | 0.81 | 0.76 | 0.71 | 0.60 | 3.2 | 0.1 | 9.4 | 22.9 | 72.5 | 132.3 | 0.75 | 0.99 | 0.82 | 0.82 | 0.77 | 0.47 |
| | 1000k | 0.81 | 0.97 | 0.81 | 0.80 | 0.75 | 0.69 | 2.4 | 0.9 | 9.2 | 21.4 | 65.8 | 107.3 | 0.79 | 0.95 | 0.84 | 0.86 | 0.82 | 0.60 |

**Table 11.** Relative overall performance of the algorithms for each case study and sample size $n$ combination, as determined by each of the three metrics, and over all 16 experiments (both with and without data noise). The performance is measured relative to the min/max values depicted in Table 10. An F represents *at least one* failed attempt by the algorithm to produce a graph for the particular case study and sample size combination (refer to Appendix C).

| Algorithm | $n$ | F1 | | | | | | SHD | | | | | | BSF | | | | | |
|---|---|---|---|---|---|---|---|---|---|---|---|---|---|---|---|---|---|---|---|
| | | Asia | Spor | Prop | Alar | Form | Path | Asia | Spor | Prop | Alar | Form | Path | Asia | Spor | Prop | Alar | Form | Path |
| FCI | 0.1k | 1% | 100% | 34% | 66% | 55% | 26% | 13% | 100% | 54% | 82% | 91% | 83% | 0% | 100% | 31% | 45% | 47% | 22% |
| | 1k | 1% | 56% | 56% | 74% | 53% | F | 12% | 64% | 39% | 65% | 55% | F | 10% | 63% | 60% | 70% | 49% | F |
| | 10k | 0% | 57% | 72% | 77% | 48% | F | 0% | 75% | 46% | 56% | 33% | F | 2% | 78% | 84% | 83% | 50% | F |
| | 100k | 5% | 50% | F | 61% | F | F | 6% | 36% | F | 0% | F | F | 11% | 43% | F | 77% | F | F |
| | 1000k | 9% | 31% | F | F | F | F | 6% | 0% | F | F | F | F | 11% | 11% | F | F | F | F |
| FGES | 0.1k | 41% | 36% | 32% | 63% | 45% | 50% | 54% | 53% | 53% | 76% | 94% | 86% | 43% | 41% | 29% | 46% | 38% | 49% |
| | 1k | 38% | 48% | 57% | 70% | 56% | 66% | 52% | 54% | 46% | 64% | 65% | 35% | 49% | 52% | 60% | 69% | 52% | 56% |
| | 10k | 34% | 29% | 68% | 76% | 74% | 43% | 58% | 52% | 50% | 71% | 70% | 0% | 41% | 48% | 66% | 74% | 70% | 42% |
| | 100k | 42% | 25% | 70% | 82% | 84% | F | 62% | 28% | 38% | 91% | 76% | F | 53% | 27% | 81% | 77% | 71% | F |
| | 1000k | 38% | 37% | 57% | 88% | F | F | 52% | 29% | 41% | 92% | F | F | 47% | 30% | 74% | 77% | F | F |
| GFCI | 0.1k | 41% | 36% | 32% | 58% | 45% | 52% | 54% | 53% | 53% | 77% | 95% | 87% | 43% | 41% | 29% | 42% | 38% | 51% |
| | 1k | 39% | 48% | 57% | 71% | 54% | 64% | 57% | 54% | 46% | 67% | 62% | 36% | 51% | 52% | 60% | 68% | 50% | 54% |
| | 10k | 35% | 29% | 66% | 77% | 72% | 42% | 62% | 52% | 50% | 74% | 73% | 1% | 44% | 48% | 65% | 74% | 66% | 41% |
| | 100k | 44% | 25% | 68% | 86% | 80% | 15% | 64% | 28% | 40% | 100% | 75% | 0% | 55% | 27% | 77% | 78% | 68% | 18% |
| | 1000k | 36% | 38% | 57% | 92% | F | F | 51% | 31% | 44% | 100% | F | F | 45% | 33% | 71% | 77% | F | F |
| GS | 0.1k | 0% | 34% | 0% | 0% | 0% | 0% | 19% | 52% | 40% | 45% | 80% | 98% | 2% | 39% | 0% | 0% | 0% | 0% |
| | 1k | 12% | 0% | 0% | 0% | 0% | 0% | 24% | 2% | 0% | 0% | 0% | 94% | 13% | 0% | 0% | 0% | 0% | 0% |
| | 10k | 12% | 37% | 0% | 0% | 0% | 0% | 11% | 53% | 0% | 0% | 0% | 91% | 1% | 50% | 0% | 0% | 0% | 0% |
| | 100k | 1% | 31% | 0% | 0% | 0% | 0% | 4% | 35% | 2% | 18% | 0% | 80% | 0% | 33% | 0% | 0% | 0% | 0% |
| | 1000k | 0% | 36% | 0% | 0% | 0% | 0% | 7% | 37% | 35% | 39% | 29% | 0% | 2% | 36% | 0% | 0% | 0% | 0% |
| H2PC | 0.1k | 54% | 62% | F | 51% | F | F | 63% | 69% | F | 81% | F | F | 49% | 61% | F | 29% | F | F |
| | 1k | 96% | 74% | F | 59% | F | F | 100% | 69% | F | 53% | F | F | 89% | 67% | F | 47% | F | F |
| | 10k | 75% | 99% | F | F | F | F | 74% | 99% | F | F | F | F | 67% | 99% | F | F | F | F |
| | 100k | 76% | 96% | 75% | F | 97% | F | 64% | 95% | 64% | F | 98% | F | 72% | 95% | 83% | F | 78% | F |
| | 1000k | 75% | 100% | 71% | 100% | F | F | 52% | 100% | 61% | 76% | F | F | 71% | 100% | 91% | 100% | F | F |
| HC | 0.1k | 100% | 34% | 54% | 90% | 100% | 100% | 100% | 52% | 61% | 100% | 80% | 92% | 100% | 39% | 48% | 65% | 100% | 100% |
| | 1k | 100% | 100% | 74% | 76% | 100% | 96% | 93% | 100% | 62% | 70% | 100% | 44% | 100% | 100% | 75% | 73% | 100% | 86% |
| | 10k | 100% | 100% | 61% | 74% | 99% | 100% | 98% | 100% | 39% | 60% | 99% | 60% | 100% | 100% | 69% | 79% | 98% | 100% |
| | 100k | 100% | 100% | 51% | 85% | 98% | 100% | 87% | 100% | 0% | 72% | 69% | 100% | 100% | 100% | 80% | 93% | 98% | 100% |
| | 1000k | 78% | 99% | 36% | 75% | 99% | 100% | 52% | 98% | 0% | 21% | 0% | 100% | 76% | 99% | 80% | 95% | 99% | 100% |
| ILP | 0.1k | 44% | 72% | 100% | 100% | 68% | 66% | 19% | 75% | 100% | 0% | 0% | 0% | 34% | 69% | 100% | 100% | 96% | 83% |
| | 1k | 82% | 89% | 100% | 100% | 88% | F | 86% | 88% | 100% | 100% | 78% | F | 94% | 88% | 100% | 100% | 93% | F |
| | 10k | 81% | 70% | 100% | 100% | 80% | F | 83% | 79% | 100% | 98% | 57% | F | 84% | 77% | 100% | 100% | 88% | F |
| | 100k | 55% | 88% | 76% | 80% | F | F | 57% | 91% | 34% | 60% | F | F | 62% | 90% | 98% | 87% | F | F |
| | 1000k | 51% | 84% | 58% | 52% | F | F | 44% | 89% | 14% | 0% | F | F | 54% | 88% | 91% | 77% | F | F |
| Inter-IAMB | 0.1k | 1% | 94% | 13% | 18% | 24% | 12% | 23% | 96% | 49% | 58% | 87% | 96% | 4% | 95% | 11% | 11% | 20% | 10% |
| | 1k | 0% | 40% | 19% | 34% | 33% | 26% | 5% | 42% | 15% | 33% | 34% | 90% | 0% | 39% | 17% | 29% | 29% | 18% |
| | 10k | 11% | 43% | 41% | 52% | 38% | 11% | 6% | 58% | 35% | 49% | 31% | 93% | 0% | 58% | 37% | 46% | 36% | 7% |
| | 100k | 0% | 42% | 42% | 58% | 59% | 9% | 0% | 40% | 41% | 70% | 44% | 81% | 0% | 41% | 42% | 55% | 51% | 6% |
| | 1000k | 1% | 41% | 50% | 73% | 68% | 10% | 0% | 34% | 62% | 74% | 100% | 4% | 0% | 36% | 56% | 68% | 56% | 7% |
| MMHC | 0.1k | 66% | 34% | 18% | 46% | 75% | 33% | 73% | 52% | 56% | 79% | 100% | 100% | 61% | 39% | 13% | 26% | 58% | 27% |
| | 1k | 87% | 93% | 52% | 56% | 77% | 54% | 90% | 91% | 47% | 54% | 85% | 100% | 81% | 90% | 43% | 43% | 61% | 37% |
| | 10k | 89% | 99% | 52% | 62% | 81% | 28% | 85% | 99% | 46% | 62% | 84% | 100% | 81% | 99% | 42% | 51% | 65% | 18% |
| | 100k | 82% | 76% | 60% | 58% | 77% | 16% | 69% | 70% | 61% | 74% | 68% | 84% | 77% | 69% | 54% | 43% | 57% | 10% |
| | 1000k | 71% | 91% | 67% | 63% | F | 16% | 48% | 88% | 75% | 90% | F | 16% | 68% | 87% | 65% | 37% | F | 9% |
| NOTEARS | 0.1k | n/a | 98% | n/a | n/a | n/a | n/a | n/a | 33% | n/a | n/a | n/a | n/a | n/a | 41% | n/a | n/a | n/a | n/a |
| | 1k | n/a | 27% | n/a | n/a | n/a | n/a | n/a | 0% | n/a | n/a | n/a | n/a | n/a | 6% | n/a | n/a | n/a | n/a |
| | 10k | n/a | 0% | n/a | n/a | n/a | n/a | n/a | 0% | n/a | n/a | n/a | n/a | n/a | 0% | n/a | n/a | n/a | n/a |
| | 100k | n/a | 0% | n/a | n/a | n/a | n/a | n/a | 0% | n/a | n/a | n/a | n/a | n/a | 0% | n/a | n/a | n/a | n/a |
| | 1000k | n/a | 0% | n/a | n/a | n/a | n/a | n/a | 3% | n/a | n/a | n/a | n/a | n/a | 0% | n/a | n/a | n/a | n/a |
| PC-Stable | 0.1k | 1% | 98% | 37% | 69% | 59% | F | 13% | 91% | 57% | 88% | 91% | F | 0% | 90% | 34% | 46% | 50% | F |
| | 1k | 10% | 68% | 58% | 81% | 65% | F | 19% | 72% | 41% | 73% | 65% | F | 18% | 71% | 62% | 77% | 58% | F |
| | 10k | 31% | 60% | 75% | 75% | 67% | F | 25% | 75% | 49% | 54% | 57% | F | 32% | 78% | 86% | 81% | 65% | F |
| | 100k | 20% | 38% | F | 71% | F | F | 17% | 27% | F | 11% | F | F | 25% | 33% | F | 85% | F | F |
| | 1000k | 12% | 29% | F | F | F | F | 8% | 0% | F | F | F | F | 15% | 10% | F | F | F | F |
| RFCI-BSC | 0.1k | 14% | F | 28% | 61% | 47% | 27% | 23% | F | 51% | 76% | 92% | 87% | 13% | F | 24% | 43% | 39% | 23% |
| | 1k | 18% | F | 22% | 57% | 37% | F | 25% | F | 18% | 51% | 36% | F | 23% | F | 22% | 55% | 31% | F |
| | 10k | 23% | F | 33% | F | F | F | 43% | F | 20% | F | F | F | 27% | F | 32% | F | F | F |
| | 100k | F | F | F | F | F | F | F | F | F | F | F | F | F | F | F | F | F | F |
| | 1000k | F | F | F | F | F | F | F | F | F | F | F | F | F | F | F | F | F | F |
| SaiyanH | 0.1k | 59% | 73% | 52% | 81% | 70% | 64% | 0% | 0% | 0% | 31% | 43% | 45% | 45% | 0% | 59% | 67% | 82% | 72% |
| | 1k | 40% | 74% | 62% | 63% | 70% | 100% | 0% | 70% | 20% | 34% | 28% | 0% | 44% | 71% | 73% | 68% | 78% | 100% |
| | 10k | 50% | 80% | 82% | 73% | 76% | 75% | 49% | 78% | 74% | 54% | 71% | 25% | 56% | 81% | 83% | 79% | 70% | 77% |
| | 100k | 74% | 74% | 100% | 87% | 87% | F | 95% | 74% | 100% | 89% | 100% | F | 84% | 74% | 100% | 87% | 67% | F |
| | 1000k | 100% | 72% | 100% | 93% | F | F | 100% | 78% | 100% | 84% | F | F | 100% | 76% | 95% | 86% | F | F |
| TABU | 0.1k | 89% | 34% | 55% | 85% | 93% | 98% | 87% | 52% | 61% | 94% | 75% | 91% | 87% | 39% | 48% | 62% | 95% | 98% |
| | 1k | 82% | 93% | 75% | 74% | 100% | 96% | 79% | 94% | 65% | 68% | 98% | 43% | 86% | 93% | 75% | 72% | 100% | 85% |
| | 10k | 92% | 81% | 66% | 74% | 100% | 99% | 92% | 84% | 50% | 60% | 100% | 59% | 93% | 84% | 72% | 79% | 100% | 99% |
| | 100k | 75% | 94% | 56% | 100% | 100% | 100% | 66% | 96% | 7% | 96% | 72% | 100% | 77% | 96% | 84% | 100% | 100% | 100% |
| | 1000k | 66% | 91% | 70% | 78% | 100% | 99% | 45% | 92% | 35% | 24% | 9% | 99% | 64% | 92% | 100% | 97% | 100% | 100% |
| WINASOBS | 0.1k | 23% | 0% | 26% | 48% | 52% | 83% | 48% | 32% | 47% | 71% | 91% | 99% | 29% | 14% | 21% | 31% | 44% | 84% |
| | 1k | 14% | 67% | 62% | 62% | 52% | 75% | 38% | 71% | 54% | 59% | 57% | 61% | 21% | 69% | 62% | 55% | 47% | 62% |
| | 10k | 69% | 85% | 76% | 93% | 60% | 80% | 100% | 91% | 74% | 100% | 59% | 65% | 69% | 90% | 73% | 84% | 57% | 73% |
| | 100k | 88% | 88% | 74% | 73% | 74% | 83% | 100% | 90% | 53% | 75% | 63% | 99% | 89% | 89% | 92% | 75% | 67% | 73% |
| | 1000k | 61% | 93% | 45% | 24% | F | F | 62% | 96% | 69% | 43% | F | F | 66% | 95% | 53% | 29% | F | F |

Table 12 summarises the strengths and weaknesses of each algorithm over different categories. Assessments under category *Performance* are based on all three scoring metrics. All comparisons are scaled between 0% and 100%, where 0% represents the worst performance in that category and 100% the best performance. The categories are:

i. *Ranking (Performance)*: determined by the rankings presented in Tables 8 and 9. Because the results in Table 8 are based on a just one of the 16 experiments (i.e., case $N$) and the results in Table 9 on 15 experiments (i.e., all noisy experiments), the results in Table 9 are 15 times more important than the results in Table 8 in determining the scores of this category.

ii. *Small networks (Performance)*: determined by the results presented in Table 11, but restricted to networks Asia, Sports and Property.

iii. *Large networks (Performance)*: determined by the results presented in Table 11, but restricted to networks Alarm, ForMed and Pathfinder.

iv. *Limited data (Performance)*: determined by the results presented in Table 11, but restricted to sample sizes 0.1k and 1k over all case studies, and to sample size 10k for the larger networks of Alarm, ForMed and Pathfinder.

v. *Big data (Performance)*: determined by the results presented in Table 11, but restricted to sample sizes 100k and 1000k over all case studies, and to sample size 10k for the smaller networks of Asia, Sports and Property.

vi. *Variance (Performance)*: determined by the variance in the results presented in Table 11. Note that while we consider consistency in the results to be a positive feature, this outcome should be interpreted with care. This is because algorithms such as GS and Inter-IAMB that rank highly in consistency do so because the scores they produce are consistently very poor.

vii. *Under/Over-fitting*: determined by the average discrepancy between learned and true edges for each case study. A better measure would have been the discrepancy between the number of learned free parameters relative to the number of free parameters in the true graph. However, because some algorithms produce undirected and bidirected edges which complicate the estimation of these parameters, we chose to simplify the measure by relying on the number of edges (refer to Fig 2 for examples).

viii. *Computational speed*: determined by the total elapsed time over both Figures 3 and 4.

ix. *Reliability*: determined by the total number of failures with fail code F2 and F3, as shown in Tables C1 and C2. Note that fail code F1 is not considered here because it already contributes in determining *Computational speed* above.

**Table 12.** The strengths and weaknesses of the algorithms for each of the categories, based on the empirical results presented in this study, where 0% and 100% represent the weakest and strongest performance for each category.

| Algorithm | Performance | | | | | | Under/Over-fitting | Computational speed | Reliability |
|---|---|---|---|---|---|---|---|---|---|
| | Ranking | Smaller networks | Larger networks | Limited data | Big data | Variance | | | |
| FCI | 46% | 35% | 61% | 55% | 33% | 0% | 31% | 52% | 99% |
| FGES | 60% | 52% | 71% | 58% | 67% | 96% | 64% | 84% | 87% |
| GFCI | 60% | 52% | 66% | 58% | 63% | 83% | 62% | 86% | 93% |
| GS | 16% | 1% | 0% | 0% | 16% | 75% | 0% | 100% | 100% |
| H2PC | 79% | 100% | 81% | 74% | 99% | 97% | 36% | 74% | 81% |
| HC | 100% | 100% | 99% | 100% | 95% | 55% | 49% | 100% | 100% |
| ILP | 82% | 98% | 81% | 93% | 83% | 43% | 62% | 63% | 77% |
| Inter-IAMB | 37% | 26% | 41% | 31% | 45% | 29% | 29% | 100% | 100% |
| MMHC | 71% | 84% | 60% | 68% | 75% | 64% | 25% | 99% | 100% |
| NOTEARS | 0% | 0% | 0% | 29% | 0% | 100% | 100% | 100% | 100% |
| PC-Stable | 56% | 42% | 73% | 63% | 39% | 18% | 29% | 55% | 95% |
| RFCI-BSC | 11% | 18% | 52% | 35% | 0% | 99% | 11% | 0% | 0% |
| SaiyanH | 74% | 81% | 76% | 58% | 100% | 36% | 94% | 77% | 100% |
| TABU | 99% | 93% | 100% | 95% | 95% | 78% | 51% | 100% | 100% |
| WINASOBS | 72% | 76% | 73% | 60% | 88% | 64% | 48% | 97% | 97% |

## 7. DISCUSSION AND CONCLUDING REMARKS

This paper presents a new methodology that models the level of difference in structure learning performance between traditional synthetic data and noisy data. The methodology involves applying the algorithms to synthetic data that incorporate different types of noise, and can be used as a knowledge-based method to better estimate the real-world performance of structure learning algorithms under the assumption that similar types and rates of noise may exist in real data. The methodology was applied to 15 BN structure learning algorithms, which are either state-of-the-art, well-established, or recent promising implementations. The performance of the algorithms was assessed in terms of reconstructing the true causal DAG (CBN), or the MAG, under 16 different hypotheses of data noise that are investigated over six diverse case studies, with five different sample sizes per case study. The concluding remarks are derived from thousands of graphs that required approximately seven months of total structure learning runtime to produce.

Nevertheless, this study comes with some limitations. Firstly, part of the analysis is based on different assumptions about the types and levels of synthetic noise that cannot hold for all real datasets. The methodology presented in this paper aims to approximate real performance under those, or similar, conditions. Secondly, because the paper aims to measure the impact of data noise on structure learning performance, this required that the hyperparameters for each algorithm remain static across experiments, while the noise varies in the data. While we chose to test the algorithms based on their hyperparameter defaults as implemented in each software, some algorithms may be more sensitive to their hyperparameters than others.

The results are based on three scoring criteria; the F1, SHD, and BSF metrics. Our results show that the F1 and BSF metrics are largely in agreement, whereas the SHD often deviates considerably from them and. When this happens, the SHD tends to lead to conflicting and counterintuitive conclusions; an observation consistent with what has been reported in [55]. This is because the SHD score represents pure classification accuracy (i.e., summation of false positives and false negatives), whereas the F1 and BSF metrics represent partly and fully balanced scores respectively. While classification accuracy is widely considered to be misleading in other machine learning fields, the SHD metric remains popular in the field of BN structure learning. The results below are discussed across all the three metrics, but emphasis is given to the F1 and BSF metrics in deriving conclusions.

Oddly, the results suggest that score-based methods that are not designed to find causal structures (although often used for this purpose) since they represent a score-fitting exercise, are better at finding causal structures than constraint-based methods which have been designed for causal discovery by exploiting causal classes not considered by traditional machine learning methods. This is an observation that requires further investigation. Furthermore, the results suggest that data noise can have a considerable impact on the accuracy of the learned graph. Specifically, latent variables (experiments L) and the merging of states (experiments S) have had relatively minor impact on structure learning accuracy. This is not necessarily a surprising result since both these manipulations reduce the dimensionality of the input data with the residual data values not being subject to noise. On the other hand, missing and incorrect data values (experiments M and I respectively) have had a major impact on the accuracy of the learned graphs. Specifically, the reduction in structure learning accuracy occurring due to 5% or 10% missing data values ranged between 13% and 18% (i.e., accuracy increases by 15% to 21% without data noise), whereas the reduction in accuracy due to 5% or 10% incorrect data values ranged between 18% to 28% (i.e., accuracy increases by 23% to 39% without data noise). When both these types on noise are combined in a single dataset, the decrease in accuracy ranges between 26% and 30% (i.e., accuracy increases by 34% to 44% without data noise). Incorporating all four types of noise in a single dataset decreases accuracy in the range of 30% to 37% (i.e., accuracy increases by 43% to 59% without data noise). These results have major implications since they suggest that BN structure learning accuracy presented in the literature, on the basis of traditional synthetic data, overestimates real-world performance to higher degree than maybe was previously assumed. Still, traditional synthetic experiments remain important in evaluating BN structure learning algorithms under various hypothetical assumptions. All the datasets used in this study, including the raw results, graphs and BN models, are freely available online [60].

Lastly, we have not explored the option to incorporate knowledge-based constraints into the structure learning process of the algorithms. While we initially considered the possibility of exploring such constraints, not all algorithms support knowledge, nor do all algorithms which support knowledge support the same types of constraint, and this caused difficulties in setting up the experiments in a fair manner. On the other hand, disregarding knowledge-based constraints enabled us to increase the number of experiments under data noise and to provide a broader picture on its impact on the accuracy of the learned graphs. Still, knowledge-based constraints represent a desired feature when applying these algorithms to real-world problems, and it is an area for future investigation.

## ACKNOWLEDGEMENTS

This research was supported by the ERSRC Fellowship project EP/S001646/1 on *Bayesian Artificial Intelligence for Decision Making under Uncertainty*, by The Alan Turing Institute in the UK under the EPSRC grant EP/N510129/1, and by the Royal Thai Government Scholarship offered by the Office of Civil Service Commission (OCSC) in Thailand.

## APPENDIX A: SYNTHETIC DATA DETAILS

Table A1 indicates which experiments were not possible to be performed. Specifically, in the case of Asia, experiments S (i.e., merging states) could not be performed because all variables in Asia are Boolean. In the cases of Asia and Sports, experiment L5 (i.e., 5% latent variables) could not be performed because both Asia and Sports consist of less than 10 nodes; i.e., a single latent variable corresponds to a rate of approximately 10% (i.e., L10) and hence, L5 becomes redundant.

**Table A1.** Experiments not performed indicated with an X.

| | Experiment | | | | | | | | | | | | | | | |
|---|---|---|---|---|---|---|---|---|---|---|---|---|---|---|---|---|
| Case study | N | M5 | M10 | I5 | I10 | S5 | S10 | L5 | L10 | cMI | cMS | cML | cIS | cIL | cSL | cMISL |
| Alarm | | | | | | | | | | | | | | | | |
| Asia | | | | | | X | X | X | | | X | | X | | X | |
| Property | | | | | | | | | | | | | | | | |
| ForMed | | | | | | | | | | | | | | | | |
| Sports | | | | | | | | X | | | | | | | | |
| Pathfinder | | | | | | | | | | | | | | | | |

Supplementary details:

i. All noisy experiments were performed by manipulating the initially generated dataset N (i.e. no noise), for each case study. For example, experiment M5 is dataset N with approximately 5% missing data values.

ii. Lower sample size datasets are sub-datasets of the 1000k dataset. For example, the dataset N with 100k samples corresponds to the first 100k samples of the dataset N with 1000k samples.

iii. In experiments M (i.e., missing values) and I (i.e., incorrect states), each data value has a chance of being randomised (i.e., 5% or 10%). This means that some variability is expected between datasets over the total rate of noise, and the variability increases with lower sample size and fewer variables. The most extreme example involves experiment M10 in Asia, where the rates of missing values are 6.63%, 9.54%, 9.91%, 9.98%, and 10.02% for sample sizes 0.1k, 1k, 10k, 100k, and 1000k respectively.

iv. Experiments S (i.e., merged states) and L (i.e., latent variable) involve manipulation of variables rather than data values. In these experiments, approximately 5% and 10% of the variables are manipulated. Since experiment S cannot be performed on Boolean variables, the manipulation is restricted to multinomial variables. Table A2 presents the number of variables manipulated under each S and L experiments.

**Table A2.** The number of variables manipulated under each S and L experiments. Experiments that could not be performed are indicated as 'n/a' (refer to Table A1).

| | Experiment | | | | | | | | | |
|---|---|---|---|---|---|---|---|---|---|---|
| Case study | S5 | S10 | L5 | L10 | cMS | cML | cIS | cIL | cSL | cMISL |
| Alarm | 2/37 [5.4%] | 4/37 [10.8%] | 2/37 [5.4%] | 4/37 [10.8%] | 2/37 [5.4%] | 2/37 [5.4%] | 2/37 [5.4%] | 2/37 [5.4%] | 2/37 [5.4%] | 2/37 [5.4%] |
| Asia | n/a | n/a | n/a | 1/8 [12.5%] | n/a | 1/8 [12.5%] | n/a | 1/8 [12.5%] | n/a | 1/8 [12.5%] |
| Property | 1/27 [3.7%] | 3/27 [11.1%] | 1/27 [3.7%] | 3/27 [11.1%] | 1/27 [3.7%] | 1/27 [3.7%] | 1/27 [3.7%] | 1/27 [3.7%] | 1/27 [3.7%] | 1/27 [3.7%] |
| ForMed | 4/88 [4.5%] | 9/88 [10.2%] | 4/88 [4.5%] | 9/88 [10.2%] | 4/88 [4.5%] | 4/88 [4.5%] | 4/88 [4.5%] | 4/88 [4.5%] | 4/88 [4.5%] | 4/88 [4.5%] |
| Sports | n/a | 1/9 [11.1%] | n/a | 1/9 [11.1%] | 1/9 [11.1%] | 1/9 [11.1%] | 1/9 [11.1%] | 1/9 [11.1%] | 1/9 [11.1%] | 1/9 [11.1%] |
| Pathfinder | 5/109 [4.6%] | 11/109 [10.1%] | 5/109 [4.6%] | 11/109 [10.1%] | 5/109 [4.6%] | 5/109 [4.6%] | 5/109 [4.6%] | 5/109 [4.6%] | 5/109 [4.6%] | 5/109 [4.6%] |

v. Datasets that involve more than one type of noise have had each type of noise simulated in the following order: L → S → I → M. For example, the experiment cMS is dataset N is first manipulated with S and then with M.

vi. Experiment cMISL incorporates all types on noise at their default rate of 5%. However, for case Asia, cMISL does not include experiment S (refer to Table A1). Moreover, for cases Asia and Sports, cMISL includes experiments L10 instead of L5 (refer to Table A1)

## APPENDIX B: MAXIMAL ANCESTRAL GRAPHS (MAGs) – ALARM EXAMPLE

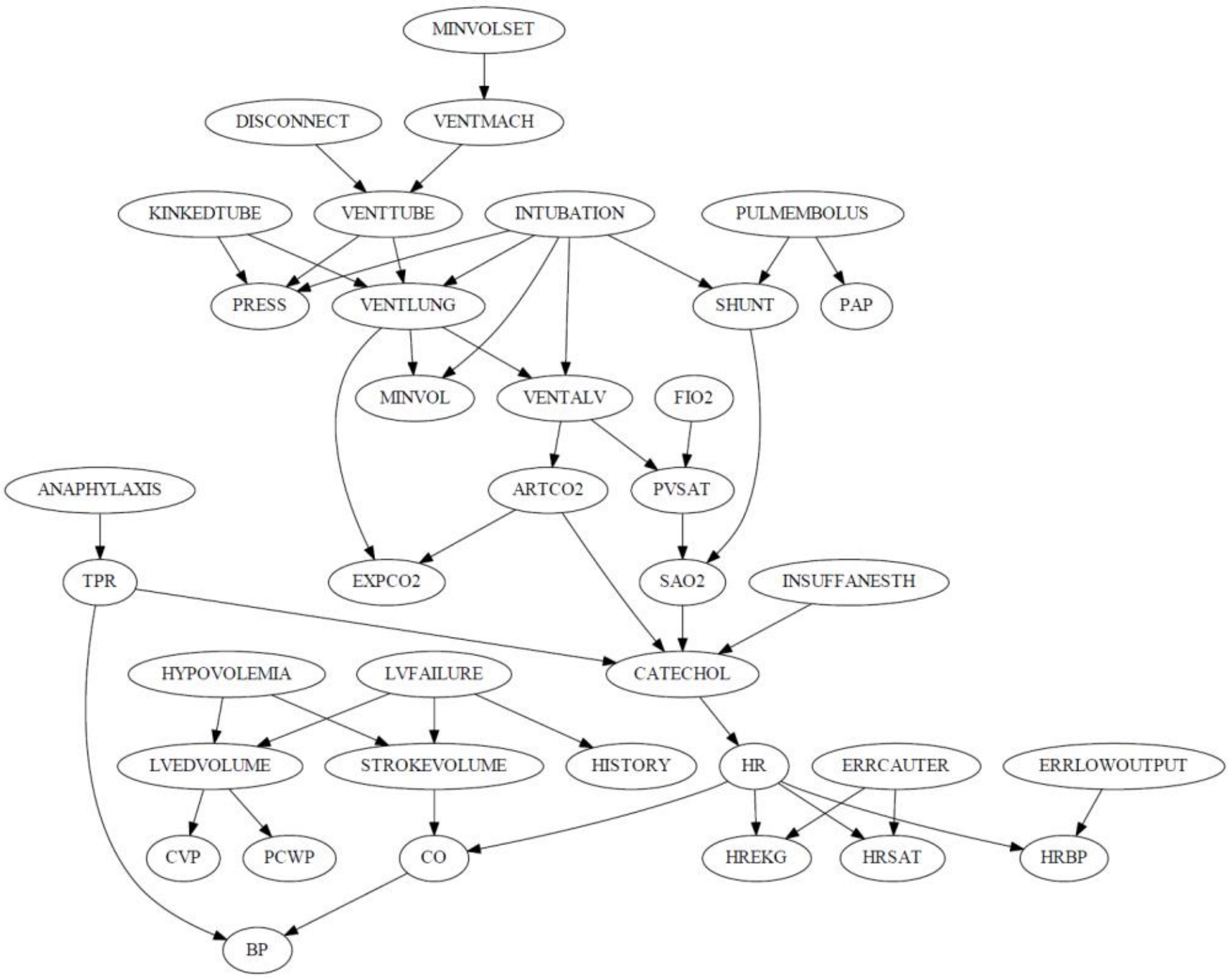

**Fig B1.** The true graph of ALARM (generated using Bayesys). Total variables: 37. Total edges: 46.

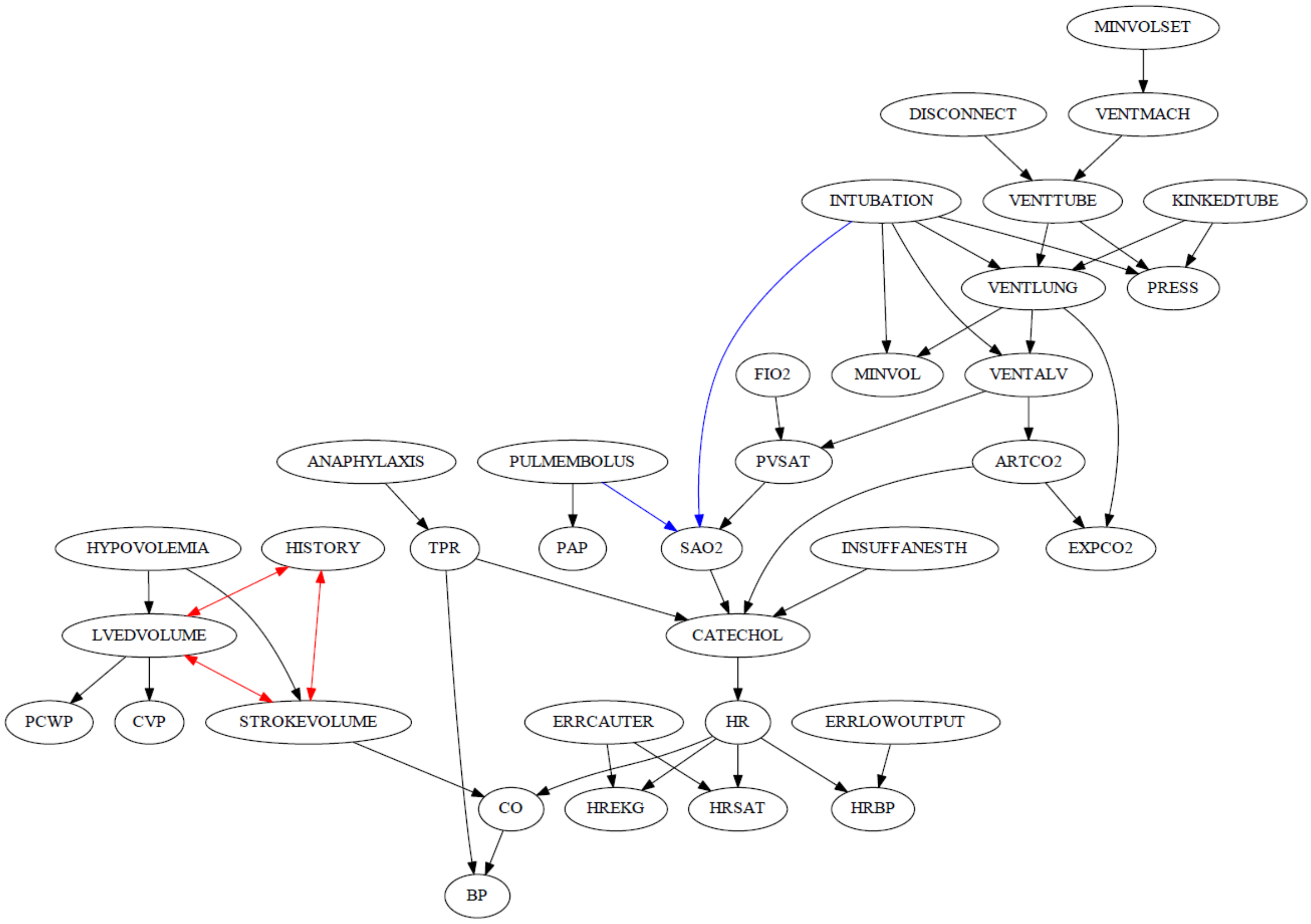

**Fig B2.** The true MAG-5 of ALARM (generated using Bayesys). Total variables: 35. Total edges: 46. Latent variables: LVFAILURE, SHUNT. Blue and red edges represent arcs and bi-directed edges in MAG, respectively, that are not present in the ground truth DAG.

## APPENDIX C: STRUCTURE LEARNING FAILED OCCURRENCES

**Table C1.** Failure information for the first eight experiments (from experiment N to L5), where F1 is represents fail event "*Algorithm does not complete within 6 hours*", F2 represents fail event "*Algorithm runtime error for unknown reason*", and F3 represents fail event "*Algorithm runtime error: Out of memory*". Higher fail occurrences are represented with a darker red backcolour.

| Algorithm | N | | | M5 | | | M10 | | | I5 | | | I10 | | | S5 | | | S10 | | | L5 | | |
|---|---|---|---|---|---|---|---|---|---|---|---|---|---|---|---|---|---|---|---|---|---|---|---|---|
| | F1 | F2 | F3 | F1 | F2 | F3 | F1 | F2 | F3 | F1 | F2 | F3 | F1 | F2 | F3 | F1 | F2 | F3 | F1 | F2 | F3 | F1 | F2 | F3 |
| FCI | 4 | 0 | 1 | 4 | 0 | 1 | 8 | 0 | 0 | 9 | 0 | 0 | 9 | 0 | 0 | 7 | 0 | 0 | 7 | 0 | 0 | 5 | 0 | 0 |
| FGES | 0 | 0 | 2 | 0 | 0 | 2 | 0 | 0 | 2 | 0 | 0 | 2 | 0 | 0 | 2 | 0 | 0 | 2 | 0 | 0 | 2 | 1 | 0 | 2 |
| GFCI | 0 | 0 | 2 | 0 | 0 | 2 | 1 | 0 | 1 | 1 | 0 | 1 | 1 | 0 | 1 | 1 | 0 | 1 | 1 | 0 | 1 | 2 | 0 | 0 |
| GS | 0 | 0 | 0 | 0 | 0 | 0 | 0 | 0 | 0 | 0 | 0 | 0 | 0 | 0 | 0 | 0 | 0 | 0 | 0 | 0 | 0 | 0 | 0 | 0 |
| H2PC | 0 | 2 | 0 | 0 | 2 | 0 | 0 | 1 | 1 | 2 | 3 | 0 | 2 | 7 | 0 | 0 | 1 | 0 | 0 | 2 | 0 | 0 | 2 | 0 |
| HC | 0 | 0 | 0 | 0 | 0 | 0 | 0 | 0 | 0 | 0 | 0 | 0 | 0 | 0 | 0 | 0 | 0 | 0 | 0 | 0 | 0 | 0 | 0 | 0 |
| ILP | 0 | 3 | 0 | 0 | 2 | 0 | 0 | 2 | 0 | 0 | 2 | 0 | 0 | 3 | 0 | 0 | 6 | 0 | 0 | 5 | 0 | 0 | 5 | 0 |
| Inter-IAMB | 0 | 0 | 0 | 0 | 0 | 0 | 0 | 0 | 0 | 0 | 0 | 0 | 0 | 0 | 0 | 0 | 0 | 0 | 0 | 0 | 0 | 0 | 0 | 0 |
| MMHC | 0 | 0 | 0 | 0 | 0 | 0 | 0 | 0 | 0 | 0 | 0 | 0 | 1 | 0 | 0 | 0 | 0 | 0 | 0 | 0 | 0 | 0 | 0 | 0 |
| NOTEARS | 0 | 0 | 0 | 0 | 0 | 0 | 0 | 0 | 0 | 0 | 0 | 0 | 0 | 0 | 0 | 0 | 0 | 0 | 0 | 0 | 0 | 0 | 0 | 0 |
| PC-Stable | 6 | 0 | 0 | 4 | 0 | 1 | 7 | 0 | 1 | 5 | 0 | 2 | 9 | 0 | 0 | 4 | 0 | 1 | 4 | 0 | 1 | 4 | 1 | 0 |
| RFCI-BSC | 0 | 16 | 1 | 0 | 18 | 0 | 0 | 18 | 0 | 0 | 17 | 0 | 0 | 16 | 0 | 0 | 12 | 0 | 1 | 15 | 0 | 0 | 11 | 0 |
| SaiyanH | 2 | 0 | 0 | 3 | 0 | 0 | 3 | 0 | 0 | 3 | 0 | 0 | 3 | 0 | 0 | 2 | 0 | 0 | 2 | 0 | 0 | 2 | 0 | 0 |
| TABU | 0 | 0 | 0 | 0 | 0 | 0 | 0 | 0 | 0 | 0 | 0 | 0 | 0 | 0 | 0 | 0 | 0 | 0 | 0 | 0 | 0 | 0 | 0 | 0 |
| WINASOBS | 0 | 0 | 0 | 0 | 0 | 0 | 0 | 0 | 0 | 0 | 0 | 0 | 0 | 0 | 0 | 0 | 0 | 0 | 0 | 0 | 0 | 0 | 1 | 0 |

**Table C2.** Failure information for the last eight experiments (from experiment L10 to cMISL), where F1 is represents fail event "*Algorithm does not complete within 6 hours*", F2 represents fail event "*Algorithm runtime error for unknown reason*", and F3 represents fail event "*Algorithm runtime error: Out of memory*". Higher fail occurrences are represented with a darker red backcolour.

| Algorithm | L10 | | | cMI | | | cMS | | | cML | | | cIS | | | cIL | | | cSL | | | cMISL | | |
|---|---|---|---|---|---|---|---|---|---|---|---|---|---|---|---|---|---|---|---|---|---|---|---|---|
| | F1 | F2 | F3 | F1 | F2 | F3 | F1 | F2 | F3 | F1 | F2 | F3 | F1 | F2 | F3 | F1 | F2 | F3 | F1 | F2 | F3 | F1 | F2 | F3 |
| FCI | 5 | 0 | 0 | 8 | 0 | 0 | 8 | 0 | 0 | 6 | 0 | 0 | 8 | 0 | 0 | 8 | 0 | 0 | 5 | 0 | 0 | 9 | 0 | 0 |
| FGES | 0 | 0 | 2 | 0 | 0 | 2 | 0 | 0 | 2 | 0 | 0 | 2 | 0 | 0 | 2 | 0 | 0 | 2 | 0 | 0 | 2 | 1 | 0 | 1 |
| GFCI | 1 | 0 | 1 | 1 | 0 | 1 | 1 | 0 | 1 | 1 | 0 | 1 | 1 | 0 | 1 | 1 | 0 | 1 | 1 | 0 | 1 | 1 | 0 | 1 |
| GS | 0 | 0 | 0 | 0 | 0 | 0 | 0 | 0 | 0 | 0 | 0 | 0 | 0 | 0 | 0 | 0 | 0 | 0 | 0 | 0 | 0 | 0 | 0 | 0 |
| H2PC | 0 | 2 | 0 | 1 | 3 | 0 | 0 | 2 | 0 | 2 | 2 | 0 | 2 | 4 | 0 | 1 | 6 | 1 | 0 | 2 | 0 | 0 | 3 | 1 |
| HC | 0 | 0 | 0 | 0 | 0 | 0 | 0 | 0 | 0 | 0 | 0 | 0 | 0 | 0 | 0 | 0 | 0 | 0 | 0 | 0 | 0 | 0 | 0 | 0 |
| ILP | 0 | 5 | 0 | 0 | 2 | 0 | 0 | 4 | 0 | 0 | 3 | 0 | 0 | 4 | 0 | 0 | 3 | 0 | 0 | 5 | 0 | 0 | 2 | 0 |
| Inter-IAMB | 0 | 0 | 0 | 0 | 0 | 0 | 0 | 0 | 0 | 0 | 0 | 0 | 0 | 0 | 0 | 0 | 0 | 0 | 0 | 0 | 0 | 0 | 0 | 0 |
| MMHC | 0 | 0 | 0 | 0 | 0 | 0 | 0 | 0 | 0 | 0 | 0 | 0 | 0 | 0 | 0 | 0 | 0 | 0 | 0 | 0 | 0 | 0 | 0 | 0 |
| NOTEARS | 0 | 0 | 0 | 0 | 0 | 0 | 0 | 0 | 0 | 0 | 0 | 0 | 0 | 0 | 0 | 0 | 0 | 0 | 0 | 0 | 0 | 0 | 0 | 0 |
| PC-Stable | 4 | 0 | 1 | 9 | 0 | 0 | 4 | 0 | 1 | 4 | 0 | 0 | 5 | 0 | 1 | 4 | 0 | 1 | 4 | 0 | 1 | 9 | 0 | 0 |
| RFCI-BSC | 0 | 15 | 0 | 5 | 13 | 0 | 0 | 16 | 0 | 0 | 16 | 0 | 0 | 16 | 0 | 0 | 14 | 0 | 0 | 13 | 0 | 0 | 15 | 0 |
| SaiyanH | 2 | 0 | 0 | 2 | 0 | 0 | 3 | 0 | 0 | 3 | 0 | 0 | 3 | 0 | 0 | 3 | 0 | 0 | 2 | 0 | 0 | 2 | 0 | 0 |
| TABU | 0 | 0 | 0 | 0 | 0 | 0 | 0 | 0 | 0 | 0 | 0 | 0 | 0 | 0 | 0 | 0 | 0 | 0 | 0 | 0 | 0 | 0 | 0 | 0 |
| WINASOBS | 0 | 1 | 0 | 0 | 0 | 0 | 0 | 0 | 0 | 0 | 1 | 0 | 0 | 0 | 0 | 0 | 1 | 0 | 0 | 2 | 0 | 0 | 1 | 0 |

## APPENDIX D: SUPPLEMENTARY RESULTS, WITHOUT DATA NOISE

**Table D1.** F1 scores of the algorithms over all experiments N, where F represents a failed attempt by the algorithm to produce a graph (refer to Appendix C). The results are presented per case study per sample size, where the sample size of 0.1 corresponds to 0.1k data samples and so forth. Darker green backcolour corresponds to higher accuracy whereas darker red backcolour corresponds to lower accuracy.

| | ALARM | | | | | ASIA | | | | | PATHFINDER | | | | | PROPERTY | | | | | SPORTS | | | | | FORMED | | | | |
|---|---|---|---|---|---|---|---|---|---|---|---|---|---|---|---|---|---|---|---|---|---|---|---|---|---|---|---|---|---|---|
| Algorithm | 0.1 | 1 | 10 | 100 | 1000 | 0.1 | 1 | 10 | 100 | 1000 | 0.1 | 1 | 10 | 100 | 1000 | 0.1 | 1 | 10 | 100 | 1000 | 0.1 | 1 | 10 | 100 | 1000 | 0.1 | 1 | 10 | 100 | 1000 |
| FCI | 0.45 | 0.78 | 0.87 | 0.89 | 0.95 | 0.27 | 0.43 | 0.57 | 0.57 | 0.67 | 0.08 | F | F | F | F | 0.33 | 0.48 | 0.76 | 0.84 | 0.84 | 0.35 | 0.46 | 0.57 | 0.83 | 0.57 | 0.32 | 0.49 | 0.58 | 0.57 | F |
| FGES | 0.41 | 0.69 | 0.79 | 0.82 | 0.82 | 0.58 | 0.80 | 0.81 | 0.81 | 0.81 | 0.17 | 0.19 | 0.18 | 0.16 | F | 0.30 | 0.56 | 0.72 | 0.75 | 0.66 | 0.24 | 0.46 | 0.44 | 0.48 | 0.67 | 0.30 | 0.60 | 0.68 | 0.71 | F |
| GFCI | 0.38 | 0.71 | 0.83 | 0.86 | 0.86 | 0.58 | 0.80 | 0.81 | 0.81 | 0.81 | 0.17 | 0.21 | 0.17 | 0.16 | F | 0.30 | 0.56 | 0.72 | 0.79 | 0.71 | 0.24 | 0.46 | 0.44 | 0.48 | 0.69 | 0.30 | 0.60 | 0.68 | 0.72 | F |
| GS | 0.16 | 0.30 | 0.45 | 0.60 | 0.61 | 0.27 | 0.58 | 0.58 | 0.62 | 0.64 | 0.04 | 0.04 | 0.08 | 0.13 | 0.17 | 0.14 | 0.26 | 0.29 | 0.49 | 0.59 | 0.12 | 0.16 | 0.52 | 0.63 | 0.68 | 0.17 | 0.27 | 0.40 | 0.35 | 0.41 |
| H2PC | 0.36 | 0.59 | 0.83 | 0.89 | 0.92 | 0.55 | 0.67 | 0.67 | 0.77 | 0.86 | 0.13 | 0.15 | 0.32 | 0.67 | 0.76 | 0.22 | 0.52 | 0.66 | 0.73 | 0.81 | 0.24 | 0.57 | 0.75 | 1.00 | 1.00 | F | F | 0.74 | 0.78 | 0.82 |
| HC | 0.52 | 0.70 | 0.73 | 0.84 | 0.82 | 0.79 | 0.93 | 1.00 | 1.00 | 1.00 | 0.24 | 0.27 | 0.38 | 0.56 | 0.67 | 0.38 | 0.66 | 0.61 | 0.64 | 0.62 | 0.18 | 0.75 | 0.75 | 1.00 | 1.00 | 0.42 | 0.71 | 0.79 | 0.78 | 0.79 |
| ILP | 0.54 | 0.86 | 0.95 | 0.96 | 0.80 | 0.35 | 0.88 | 0.88 | 0.88 | 0.88 | 0.17 | 0.27 | F | F | F | 0.59 | 0.88 | 0.98 | 0.95 | 0.84 | 0.24 | 0.67 | 0.64 | 0.93 | 0.93 | 0.31 | 0.64 | 0.68 | 0.52 | 0.59 |
| Inter-IAMB | 0.24 | 0.47 | 0.74 | 0.83 | 0.87 | 0.27 | 0.58 | 0.54 | 0.62 | 0.64 | 0.07 | 0.07 | 0.08 | 0.16 | 0.21 | 0.18 | 0.35 | 0.56 | 0.65 | 0.70 | 0.35 | 0.38 | 0.46 | 0.59 | 0.68 | 0.21 | 0.37 | 0.52 | 0.64 | 0.64 |
| MMHC | 0.36 | 0.63 | 0.70 | 0.73 | 0.72 | 0.46 | 0.77 | 0.86 | 0.86 | 0.80 | 0.11 | 0.14 | 0.14 | 0.19 | 0.26 | 0.22 | 0.51 | 0.57 | 0.65 | 0.76 | 0.18 | 0.70 | 0.75 | 0.75 | 0.89 | 0.33 | 0.57 | 0.71 | 0.68 | 0.64 |
| NOTEARS | n/a | n/a | n/a | n/a | n/a | n/a | n/a | n/a | n/a | n/a | n/a | n/a | n/a | n/a | n/a | n/a | n/a | n/a | n/a | n/a | 0.33 | 0.32 | 0.35 | 0.35 | 0.35 | n/a | n/a | n/a | n/a | n/a |
| PC-Stable | 0.43 | 0.78 | 0.78 | 0.90 | 0.93 | 0.27 | 0.43 | 0.64 | 0.57 | 0.60 | F | F | F | F | F | 0.34 | 0.47 | 0.72 | 0.84 | 0.84 | 0.35 | 0.56 | 0.57 | 0.57 | 0.57 | 0.32 | 0.56 | 0.62 | 0.63 | F |
| RFCI-BSC | 0.42 | 0.69 | F | F | F | 0.36 | 0.58 | 0.71 | F | F | 0.09 | F | F | F | F | 0.26 | 0.43 | 0.56 | F | F | 0.24 | 0.33 | F | F | F | 0.30 | 0.43 | F | F | F |
| SaiyanH | 0.43 | 0.73 | 0.75 | 0.88 | 0.78 | 0.53 | 0.69 | 0.88 | 0.88 | 0.88 | 0.18 | 0.25 | 0.32 | 0.38 | F | 0.39 | 0.58 | 0.81 | 0.81 | 0.79 | 0.22 | 0.56 | 0.77 | 0.90 | 0.90 | 0.32 | 0.53 | 0.73 | 0.75 | F |
| TABU | 0.52 | 0.70 | 0.73 | 0.96 | 0.85 | 0.43 | 0.93 | 1.00 | 1.00 | 1.00 | 0.24 | 0.27 | 0.37 | 0.56 | 0.67 | 0.38 | 0.66 | 0.64 | 0.68 | 0.85 | 0.18 | 0.75 | 0.68 | 1.00 | 1.00 | 0.38 | 0.71 | 0.79 | 0.80 | 0.80 |
| WINASOBS | 0.39 | 0.70 | 0.88 | 0.86 | 0.59 | 0.50 | 0.57 | 0.87 | 0.94 | 0.94 | 0.08 | 0.25 | 0.33 | 0.49 | 0.57 | 0.31 | 0.57 | 0.74 | 0.62 | 0.66 | 0.00 | 0.75 | 0.75 | 1.00 | 1.00 | 0.30 | 0.50 | 0.66 | 0.78 | 0.51 |

**Table D2.** SHD scores of the algorithms over all experiments N, where F represents a failed attempt by the algorithm to produce a graph (refer to Appendix C). The results are presented per case study per sample size, where the sample size of 0.1 corresponds to 0.1k data samples and so forth. Darker green backcolour corresponds to higher accuracy whereas darker red backcolour corresponds to lower accuracy.

| | ALARM | | | | | ASIA | | | | | PATHFINDER | | | | | PROPERTY | | | | | SPORTS | | | | | FORMED | | | | |
|---|---|---|---|---|---|---|---|---|---|---|---|---|---|---|---|---|---|---|---|---|---|---|---|---|---|---|---|---|---|---|
| Algorithm | 0.1 | 1 | 10 | 100 | 1000 | 0.1 | 1 | 10 | 100 | 1000 | 0.1 | 1 | 10 | 100 | 1000 | 0.1 | 1 | 10 | 100 | 1000 | 0.1 | 1 | 10 | 100 | 1000 | 0.1 | 1 | 10 | 100 | 1000 |
| FCI | 32 | 15 | 8 | 7 | 3 | 7 | 5 | 4 | 4 | 4 | 199 | F | F | F | F | 24 | 20 | 11 | 7 | 7 | 12 | 10 | 7 | 3 | 7 | 111 | 87 | 72 | 86 | F |
| FGES | 39 | 21 | 16 | 15 | 15 | 5 | 2 | 2 | 2 | 2 | 211 | 246 | 292 | 318 | F | 26 | 21 | 14 | 15 | 21 | 13 | 10 | 11 | 11 | 8 | 116 | 72 | 71 | 68 | F |
| GFCI | 37 | 19 | 13 | 11 | 11 | 5 | 2 | 2 | 2 | 2 | 208 | 240 | 288 | 317 | F | 26 | 21 | 14 | 12 | 16 | 13 | 10 | 11 | 11 | 7 | 116 | 73 | 68 | 62 | F |
| GS | 43 | 39 | 33 | 25 | 26 | 7 | 5 | 5 | 4 | 4 | 191 | 191 | 188 | 181 | 176 | 29 | 26 | 26 | 21 | 17 | 14 | 15 | 9 | 8 | 7 | 130 | 117 | 100 | 108 | 102 |
| H2PC | 37 | 26 | 11 | 7 | 4 | 5 | 4 | 4 | 3 | 2 | 189 | 187 | 167 | 101 | 75 | 27 | 20 | 15 | 11 | 8 | 13 | 9 | 6 | 0 | 0 | F | F | 53 | 48 | 42 |
| HC | 32 | 21 | 20 | 13 | 14 | 3 | 1 | 0 | 0 | 0 | 209 | 233 | 214 | 156 | 115 | 26 | 17 | 21 | 20 | 22 | 14 | 6 | 6 | 0 | 0 | 130 | 65 | 47 | 55 | 55 |
| ILP | 48 | 12 | 5 | 3 | 15 | 9 | 1 | 1 | 1 | 1 | 318 | 280 | F | F | F | 19 | 7 | 1 | 2 | 9 | 13 | 7 | 8 | 1 | 1 | 205 | 81 | 77 | 133 | 94 |
| Inter-IAMB | 40 | 29 | 17 | 10 | 7 | 7 | 5 | 6 | 4 | 4 | 197 | 192 | 190 | 179 | 171 | 28 | 24 | 19 | 14 | 12 | 12 | 11 | 10 | 8 | 7 | 122 | 104 | 85 | 69 | 78 |
| MMHC | 37 | 23 | 19 | 18 | 18 | 6 | 3 | 2 | 2 | 3 | 191 | 185 | 183 | 176 | 166 | 27 | 20 | 18 | 15 | 11 | 14 | 7 | 6 | 6 | 3 | 112 | 80 | 60 | 66 | 75 |
| NOTEARS | n/a | n/a | n/a | n/a | n/a | n/a | n/a | n/a | n/a | n/a | n/a | n/a | n/a | n/a | n/a | n/a | n/a | n/a | n/a | n/a | 14 | 15 | 14 | 14 | 14 | n/a | n/a | n/a | n/a | n/a |
| PC-Stable | 32 | 14 | 13 | 6 | 4 | 7 | 5 | 4 | 4 | 5 | F | F | F | F | F | 24 | 21 | 13 | 7 | 7 | 12 | 8 | 7 | 7 | 7 | 112 | 80 | 68 | 77 | F |
| RFCI-BSC | 35 | 19 | F | F | F | 7 | 5 | 3 | F | F | 207 | F | F | F | F | 26 | 21 | 17 | F | F | 13 | 12 | F | F | F | 114 | 95 | F | F | F |
| SaiyanH | 40 | 21 | 20 | 10 | 20 | 6 | 5 | 1 | 1 | 1 | 271 | 277 | 219 | 225 | F | 35 | 23 | 10 | 10 | 10 | 17 | 10 | 7 | 2 | 2 | 146 | 96 | 53 | 52 | F |
| TABU | 32 | 21 | 20 | 3 | 12 | 6 | 1 | 0 | 0 | 0 | 209 | 233 | 218 | 156 | 116 | 26 | 16 | 19 | 18 | 8 | 14 | 6 | 8 | 0 | 0 | 139 | 64 | 48 | 52 | 54 |
| WINASOBS | 38 | 20 | 9 | 11 | 30 | 5 | 4 | 2 | 1 | 1 | 218 | 215 | 203 | 160 | 128 | 27 | 19 | 12 | 20 | 15 | 15 | 6 | 6 | 0 | 0 | 122 | 90 | 68 | 47 | 84 |

**Table D3.** BSF scores of the algorithms over all experiments N, where F represents a failed attempt by the algorithm to produce a graph (refer to Appendix C). The results are presented per case study per sample size, where the sample size of 0.1 corresponds to 0.1k data samples and so forth. Darker green backcolour corresponds to higher accuracy whereas darker red backcolour corresponds to lower accuracy.

| | ALARM | | | | | ASIA | | | | | PATHFINDER | | | | | PROPERTY | | | | | SPORTS | | | | | FORMED | | | | |
|---|---|---|---|---|---|---|---|---|---|---|---|---|---|---|---|---|---|---|---|---|---|---|---|---|---|---|---|---|---|---|
| Algorithm | 0.1 | 1 | 10 | 100 | 1000 | 0.1 | 1 | 10 | 100 | 1000 | 0.1 | 1 | 10 | 100 | 1000 | 0.1 | 1 | 10 | 100 | 1000 | 0.1 | 1 | 10 | 100 | 1000 | 0.1 | 1 | 10 | 100 | 1000 |
| FCI | 0.32 | 0.69 | 0.83 | 0.88 | 0.95 | 0.19 | 0.38 | 0.50 | 0.50 | 0.58 | 0.04 | F | F | F | F | 0.23 | 0.38 | 0.70 | 0.77 | 0.77 | 0.23 | 0.37 | 0.57 | 0.83 | 0.57 | 0.21 | 0.37 | 0.49 | 0.52 | F |
| FGES | 0.32 | 0.66 | 0.76 | 0.83 | 0.83 | 0.44 | 0.75 | 0.81 | 0.81 | 0.81 | 0.11 | 0.14 | 0.14 | 0.13 | F | 0.22 | 0.48 | 0.67 | 0.75 | 0.66 | 0.13 | 0.37 | 0.32 | 0.34 | 0.52 | 0.19 | 0.48 | 0.63 | 0.68 | F |
| GFCI | 0.29 | 0.65 | 0.79 | 0.83 | 0.83 | 0.44 | 0.75 | 0.81 | 0.81 | 0.81 | 0.11 | 0.15 | 0.14 | 0.14 | F | 0.22 | 0.48 | 0.67 | 0.76 | 0.69 | 0.13 | 0.37 | 0.32 | 0.34 | 0.57 | 0.19 | 0.48 | 0.62 | 0.68 | F |
| GS | 0.10 | 0.20 | 0.34 | 0.47 | 0.51 | 0.19 | 0.44 | 0.44 | 0.50 | 0.56 | 0.02 | 0.02 | 0.04 | 0.07 | 0.10 | 0.08 | 0.16 | 0.19 | 0.37 | 0.48 | 0.07 | 0.05 | 0.40 | 0.52 | 0.59 | 0.10 | 0.18 | 0.28 | 0.25 | 0.30 |
| H2PC | 0.23 | 0.47 | 0.76 | 0.85 | 0.92 | 0.38 | 0.50 | 0.50 | 0.63 | 0.75 | 0.07 | 0.08 | 0.20 | 0.52 | 0.65 | 0.13 | 0.37 | 0.56 | 0.65 | 0.74 | 0.13 | 0.40 | 0.60 | 1.00 | 1.00 | F | F | 0.62 | 0.68 | 0.76 |
| HC | 0.43 | 0.68 | 0.75 | 0.87 | 0.90 | 0.69 | 0.88 | 1.00 | 1.00 | 1.00 | 0.16 | 0.21 | 0.33 | 0.51 | 0.62 | 0.28 | 0.58 | 0.61 | 0.68 | 0.67 | 0.10 | 0.60 | 0.60 | 1.00 | 1.00 | 0.34 | 0.64 | 0.79 | 0.83 | 0.87 |
| ILP | 0.61 | 0.87 | 0.94 | 0.95 | 0.88 | 0.18 | 0.88 | 0.88 | 0.88 | 0.88 | 0.14 | 0.25 | F | F | F | 0.49 | 0.79 | 0.97 | 0.94 | 0.81 | 0.13 | 0.53 | 0.49 | 0.93 | 0.93 | 0.32 | 0.59 | 0.70 | 0.61 | 0.61 |
| Inter-IAMB | 0.15 | 0.37 | 0.64 | 0.78 | 0.85 | 0.19 | 0.44 | 0.39 | 0.50 | 0.56 | 0.04 | 0.04 | 0.05 | 0.09 | 0.13 | 0.11 | 0.24 | 0.46 | 0.55 | 0.61 | 0.23 | 0.27 | 0.35 | 0.49 | 0.59 | 0.13 | 0.27 | 0.42 | 0.55 | 0.60 |
| MMHC | 0.23 | 0.50 | 0.59 | 0.62 | 0.62 | 0.31 | 0.63 | 0.75 | 0.75 | 0.70 | 0.06 | 0.08 | 0.08 | 0.11 | 0.15 | 0.13 | 0.37 | 0.45 | 0.53 | 0.66 | 0.10 | 0.53 | 0.60 | 0.60 | 0.80 | 0.21 | 0.42 | 0.58 | 0.55 | 0.52 |
| NOTEARS | n/a | n/a | n/a | n/a | n/a | n/a | n/a | n/a | n/a | n/a | n/a | n/a | n/a | n/a | n/a | n/a | n/a | n/a | n/a | n/a | 0.16 | 0.11 | 0.14 | 0.14 | 0.14 | n/a | n/a | n/a | n/a | n/a |
| PC-Stable | 0.30 | 0.70 | 0.75 | 0.89 | 0.92 | 0.19 | 0.38 | 0.56 | 0.50 | 0.51 | F | F | F | F | F | 0.24 | 0.38 | 0.66 | 0.77 | 0.77 | 0.23 | 0.47 | 0.57 | 0.57 | 0.57 | 0.21 | 0.42 | 0.52 | 0.58 | F |
| RFCI-BSC | 0.33 | 0.62 | F | F | F | 0.23 | 0.49 | 0.68 | F | F | 0.05 | F | F | F | F | 0.17 | 0.33 | 0.48 | F | F | 0.13 | 0.21 | F | F | F | 0.19 | 0.32 | F | F | F |
| SaiyanH | 0.36 | 0.70 | 0.80 | 0.90 | 0.84 | 0.40 | 0.59 | 0.88 | 0.88 | 0.88 | 0.14 | 0.22 | 0.26 | 0.36 | F | 0.34 | 0.55 | 0.79 | 0.81 | 0.78 | -0.02 | 0.37 | 0.62 | 0.90 | 0.90 | 0.26 | 0.46 | 0.66 | 0.68 | F |
| TABU | 0.43 | 0.68 | 0.75 | 0.95 | 0.91 | 0.33 | 0.88 | 1.00 | 1.00 | 1.00 | 0.16 | 0.21 | 0.32 | 0.51 | 0.62 | 0.28 | 0.57 | 0.64 | 0.71 | 0.83 | 0.10 | 0.60 | 0.52 | 1.00 | 1.00 | 0.31 | 0.65 | 0.79 | 0.85 | 0.88 |
| WINASOBS | 0.28 | 0.62 | 0.84 | 0.85 | 0.61 | 0.38 | 0.50 | 0.81 | 0.94 | 0.94 | 0.04 | 0.18 | 0.26 | 0.41 | 0.44 | 0.20 | 0.49 | 0.69 | 0.65 | 0.65 | 0.00 | 0.60 | 0.60 | 1.00 | 1.00 | 0.20 | 0.40 | 0.60 | 0.75 | 0.41 |

**Table D4.** Precision scores of the algorithms over all experiments N, where F represents a failed attempt by the algorithm to produce a graph (refer to Appendix C). The results are presented per case study per sample size, where the sample size of 0.1 corresponds to 0.1k data samples and so forth. Darker green backcolour corresponds to higher accuracy whereas darker red backcolour corresponds to lower accuracy.

| | ALARM | | | | | ASIA | | | | | PATHFINDER | | | | | PROPERTY | | | | | SPORTS | | | | | FORMED | | | | |
|---|---|---|---|---|---|---|---|---|---|---|---|---|---|---|---|---|---|---|---|---|---|---|---|---|---|---|---|---|---|---|
| Algorithm | 0.1 | 1 | 10 | 100 | 1000 | 0.1 | 1 | 10 | 100 | 1000 | 0.1 | 1 | 10 | 100 | 1000 | 0.1 | 1 | 10 | 100 | 1000 | 0.1 | 1 | 10 | 100 | 1000 | 0.1 | 1 | 10 | 100 | 1000 |
| FCI | 0.71 | 0.89 | 0.93 | 0.90 | 0.95 | 0.50 | 0.50 | 0.67 | 0.67 | 0.71 | 0.31 | F | F | F | F | 0.58 | 0.63 | 0.82 | 0.92 | 0.92 | 0.70 | 0.61 | 0.57 | 0.83 | 0.57 | 0.67 | 0.71 | 0.73 | 0.62 | F |
| FGES | 0.52 | 0.73 | 0.81 | 0.80 | 0.80 | 0.88 | 0.86 | 0.81 | 0.81 | 0.81 | 0.31 | 0.25 | 0.19 | 0.16 | F | 0.47 | 0.65 | 0.78 | 0.73 | 0.63 | 1.00 | 0.61 | 0.55 | 0.54 | 0.67 | 0.60 | 0.81 | 0.72 | 0.73 | F |
| GFCI | 0.54 | 0.77 | 0.87 | 0.88 | 0.88 | 0.88 | 0.86 | 0.81 | 0.81 | 0.81 | 0.32 | 0.28 | 0.19 | 0.17 | F | 0.47 | 0.65 | 0.78 | 0.80 | 0.71 | 1.00 | 0.61 | 0.55 | 0.54 | 0.71 | 0.60 | 0.81 | 0.75 | 0.76 | F |
| GS | 0.45 | 0.53 | 0.64 | 0.83 | 0.76 | 0.50 | 0.88 | 0.88 | 0.80 | 0.75 | 0.50 | 0.50 | 0.47 | 0.58 | 0.56 | 0.50 | 0.63 | 0.55 | 0.72 | 0.75 | 0.50 | 0.38 | 0.75 | 0.71 | 0.73 | 0.47 | 0.56 | 0.66 | 0.61 | 0.65 |
| H2PC | 0.81 | 0.80 | 0.92 | 0.93 | 0.92 | 1.00 | 1.00 | 1.00 | 1.00 | 1.00 | 0.64 | 0.67 | 0.75 | 0.91 | 0.91 | 0.67 | 0.89 | 0.80 | 0.83 | 0.89 | 1.00 | 1.00 | 1.00 | 1.00 | 1.00 | F | F | 0.92 | 0.91 | 0.90 |
| HC | 0.65 | 0.72 | 0.70 | 0.79 | 0.75 | 0.92 | 1.00 | 1.00 | 1.00 | 1.00 | 0.41 | 0.34 | 0.43 | 0.60 | 0.72 | 0.56 | 0.74 | 0.57 | 0.58 | 0.55 | 0.75 | 1.00 | 1.00 | 1.00 | 1.00 | 0.52 | 0.78 | 0.79 | 0.73 | 0.72 |
| ILP | 0.46 | 0.84 | 0.95 | 0.97 | 0.73 | 0.33 | 0.88 | 0.88 | 0.88 | 0.88 | 0.17 | 0.27 | F | F | F | 0.71 | 0.98 | 1.00 | 0.97 | 0.85 | 1.00 | 0.89 | 0.80 | 0.93 | 0.93 | 0.28 | 0.68 | 0.65 | 0.44 | 0.56 |
| Inter-IAMB | 0.54 | 0.65 | 0.87 | 0.88 | 0.89 | 0.50 | 0.88 | 0.70 | 0.80 | 0.75 | 0.31 | 0.40 | 0.41 | 0.60 | 0.58 | 0.50 | 0.63 | 0.69 | 0.81 | 0.83 | 0.70 | 0.67 | 0.55 | 0.67 | 0.73 | 0.53 | 0.59 | 0.67 | 0.76 | 0.68 |
| MMHC | 0.81 | 0.85 | 0.87 | 0.89 | 0.86 | 0.83 | 1.00 | 1.00 | 1.00 | 0.86 | 0.60 | 0.75 | 0.83 | 0.93 | 0.97 | 0.67 | 0.82 | 0.78 | 0.83 | 0.89 | 0.75 | 1.00 | 1.00 | 1.00 | 1.00 | 0.76 | 0.89 | 0.89 | 0.88 | 0.83 |
| NOTEARS | n/a | n/a | n/a | n/a | n/a | n/a | n/a | n/a | n/a | n/a | n/a | n/a | n/a | n/a | n/a | n/a | n/a | n/a | n/a | n/a | 0.38 | 0.35 | 0.36 | 0.36 | 0.36 | n/a | n/a | n/a | n/a | n/a |
| PC-Stable | 0.74 | 0.89 | 0.82 | 0.91 | 0.94 | 0.50 | 0.50 | 0.75 | 0.67 | 0.64 | F | F | F | F | F | 0.58 | 0.60 | 0.79 | 0.92 | 0.92 | 0.70 | 0.70 | 0.57 | 0.57 | 0.57 | 0.70 | 0.80 | 0.77 | 0.69 | F |
| RFCI-BSC | 0.58 | 0.78 | F | F | F | 0.53 | 0.66 | 0.75 | F | F | 0.26 | F | F | F | F | 0.49 | 0.62 | 0.68 | F | F | 1.00 | 0.73 | F | F | F | 0.64 | 0.68 | F | F | F |
| SaiyanH | 0.49 | 0.73 | 0.69 | 0.84 | 0.71 | 0.57 | 0.69 | 0.88 | 0.88 | 0.88 | 0.21 | 0.26 | 0.39 | 0.39 | F | 0.39 | 0.58 | 0.81 | 0.80 | 0.79 | 0.31 | 0.70 | 0.77 | 0.90 | 0.90 | 0.39 | 0.62 | 0.82 | 0.83 | F |
| TABU | 0.65 | 0.72 | 0.70 | 0.97 | 0.79 | 0.50 | 1.00 | 1.00 | 1.00 | 1.00 | 0.41 | 0.34 | 0.41 | 0.60 | 0.71 | 0.56 | 0.75 | 0.62 | 0.62 | 0.87 | 0.75 | 1.00 | 0.85 | 1.00 | 1.00 | 0.46 | 0.78 | 0.78 | 0.74 | 0.72 |
| WINASOBS | 0.62 | 0.79 | 0.92 | 0.86 | 0.56 | 0.75 | 0.67 | 0.93 | 0.94 | 0.94 | 0.22 | 0.38 | 0.44 | 0.61 | 0.80 | 0.59 | 0.67 | 0.80 | 0.57 | 0.66 | 0.00 | 1.00 | 1.00 | 1.00 | 1.00 | 0.55 | 0.67 | 0.73 | 0.80 | 0.66 |

**Table D5.** Recall scores of the algorithms over all experiments N, where F represents a failed attempt by the algorithm to produce a graph (refer to Appendix C). The results are presented per case study per sample size, where the sample size of 0.1 corresponds to 0.1k data samples and so forth. Darker green backcolour corresponds to higher accuracy whereas darker red backcolour corresponds to lower accuracy.

| | ALARM | | | | | ASIA | | | | | PATHFINDER | | | | | PROPERTY | | | | | SPORTS | | | | | FORMED | | | | |
|---|---|---|---|---|---|---|---|---|---|---|---|---|---|---|---|---|---|---|---|---|---|---|---|---|---|---|---|---|---|---|
| Algorithm | 0.1 | 1 | 10 | 100 | 1000 | 0.1 | 1 | 10 | 100 | 1000 | 0.1 | 1 | 10 | 100 | 1000 | 0.1 | 1 | 10 | 100 | 1000 | 0.1 | 1 | 10 | 100 | 1000 | 0.1 | 1 | 10 | 100 | 1000 |
| FCI | 0.33 | 0.70 | 0.83 | 0.88 | 0.95 | 0.19 | 0.38 | 0.50 | 0.50 | 0.63 | 0.05 | F | F | F | F | 0.23 | 0.39 | 0.71 | 0.77 | 0.77 | 0.23 | 0.37 | 0.57 | 0.83 | 0.57 | 0.21 | 0.37 | 0.49 | 0.53 | F |
| FGES | 0.34 | 0.66 | 0.77 | 0.84 | 0.84 | 0.44 | 0.75 | 0.81 | 0.81 | 0.81 | 0.11 | 0.15 | 0.17 | 0.16 | F | 0.23 | 0.50 | 0.68 | 0.77 | 0.69 | 0.13 | 0.37 | 0.37 | 0.43 | 0.67 | 0.20 | 0.48 | 0.64 | 0.69 | F |
| GFCI | 0.29 | 0.65 | 0.79 | 0.84 | 0.84 | 0.44 | 0.75 | 0.81 | 0.81 | 0.81 | 0.12 | 0.17 | 0.16 | 0.16 | F | 0.23 | 0.50 | 0.68 | 0.77 | 0.71 | 0.13 | 0.37 | 0.37 | 0.43 | 0.67 | 0.20 | 0.48 | 0.62 | 0.68 | F |
| GS | 0.10 | 0.21 | 0.35 | 0.47 | 0.51 | 0.19 | 0.44 | 0.44 | 0.50 | 0.56 | 0.02 | 0.02 | 0.04 | 0.07 | 0.10 | 0.08 | 0.16 | 0.19 | 0.37 | 0.48 | 0.07 | 0.10 | 0.40 | 0.57 | 0.63 | 0.11 | 0.18 | 0.28 | 0.25 | 0.30 |
| H2PC | 0.23 | 0.47 | 0.76 | 0.85 | 0.92 | 0.38 | 0.50 | 0.50 | 0.63 | 0.75 | 0.07 | 0.08 | 0.20 | 0.53 | 0.65 | 0.13 | 0.37 | 0.57 | 0.65 | 0.74 | 0.13 | 0.40 | 0.60 | 1.00 | 1.00 | F | F | 0.62 | 0.68 | 0.76 |
| HC | 0.44 | 0.69 | 0.76 | 0.88 | 0.91 | 0.69 | 0.88 | 1.00 | 1.00 | 1.00 | 0.17 | 0.22 | 0.34 | 0.52 | 0.63 | 0.29 | 0.60 | 0.65 | 0.71 | 0.71 | 0.10 | 0.60 | 0.60 | 1.00 | 1.00 | 0.35 | 0.65 | 0.79 | 0.84 | 0.88 |
| ILP | 0.66 | 0.88 | 0.95 | 0.95 | 0.89 | 0.38 | 0.88 | 0.88 | 0.88 | 0.88 | 0.17 | 0.27 | F | F | F | 0.50 | 0.79 | 0.97 | 0.94 | 0.82 | 0.13 | 0.53 | 0.53 | 0.93 | 0.93 | 0.36 | 0.60 | 0.71 | 0.63 | 0.62 |
| Inter-IAMB | 0.15 | 0.37 | 0.64 | 0.78 | 0.85 | 0.19 | 0.44 | 0.44 | 0.50 | 0.56 | 0.04 | 0.04 | 0.05 | 0.09 | 0.13 | 0.11 | 0.24 | 0.47 | 0.55 | 0.61 | 0.23 | 0.27 | 0.40 | 0.53 | 0.63 | 0.13 | 0.27 | 0.42 | 0.55 | 0.60 |
| MMHC | 0.23 | 0.50 | 0.59 | 0.62 | 0.62 | 0.31 | 0.63 | 0.75 | 0.75 | 0.75 | 0.06 | 0.08 | 0.08 | 0.11 | 0.15 | 0.13 | 0.37 | 0.45 | 0.53 | 0.66 | 0.10 | 0.53 | 0.60 | 0.60 | 0.80 | 0.21 | 0.42 | 0.58 | 0.55 | 0.52 |
| NOTEARS | n/a | n/a | n/a | n/a | n/a | n/a | n/a | n/a | n/a | n/a | n/a | n/a | n/a | n/a | n/a | n/a | n/a | n/a | n/a | n/a | 0.30 | 0.30 | 0.33 | 0.33 | 0.33 | n/a | n/a | n/a | n/a | n/a |
| PC-Stable | 0.30 | 0.70 | 0.75 | 0.89 | 0.92 | 0.19 | 0.38 | 0.56 | 0.50 | 0.56 | F | F | F | F | F | 0.24 | 0.39 | 0.66 | 0.77 | 0.77 | 0.23 | 0.47 | 0.57 | 0.57 | 0.57 | 0.21 | 0.42 | 0.52 | 0.58 | F |
| RFCI-BSC | 0.33 | 0.63 | F | F | F | 0.27 | 0.53 | 0.68 | F | F | 0.05 | F | F | F | F | 0.17 | 0.33 | 0.48 | F | F | 0.13 | 0.21 | F | F | F | 0.20 | 0.32 | F | F | F |
| SaiyanH | 0.38 | 0.72 | 0.82 | 0.91 | 0.86 | 0.50 | 0.69 | 0.88 | 0.88 | 0.88 | 0.16 | 0.24 | 0.27 | 0.37 | F | 0.39 | 0.58 | 0.81 | 0.82 | 0.79 | 0.17 | 0.47 | 0.77 | 0.90 | 0.90 | 0.28 | 0.47 | 0.66 | 0.68 | F |
| TABU | 0.44 | 0.69 | 0.76 | 0.95 | 0.92 | 0.38 | 0.88 | 1.00 | 1.00 | 1.00 | 0.17 | 0.22 | 0.33 | 0.52 | 0.63 | 0.29 | 0.58 | 0.66 | 0.74 | 0.84 | 0.10 | 0.60 | 0.57 | 1.00 | 1.00 | 0.33 | 0.65 | 0.79 | 0.86 | 0.89 |
| WINASOBS | 0.28 | 0.62 | 0.84 | 0.86 | 0.63 | 0.38 | 0.50 | 0.81 | 0.94 | 0.94 | 0.05 | 0.19 | 0.27 | 0.41 | 0.44 | 0.21 | 0.50 | 0.69 | 0.68 | 0.66 | 0.00 | 0.60 | 0.60 | 1.00 | 1.00 | 0.20 | 0.40 | 0.60 | 0.76 | 0.41 |

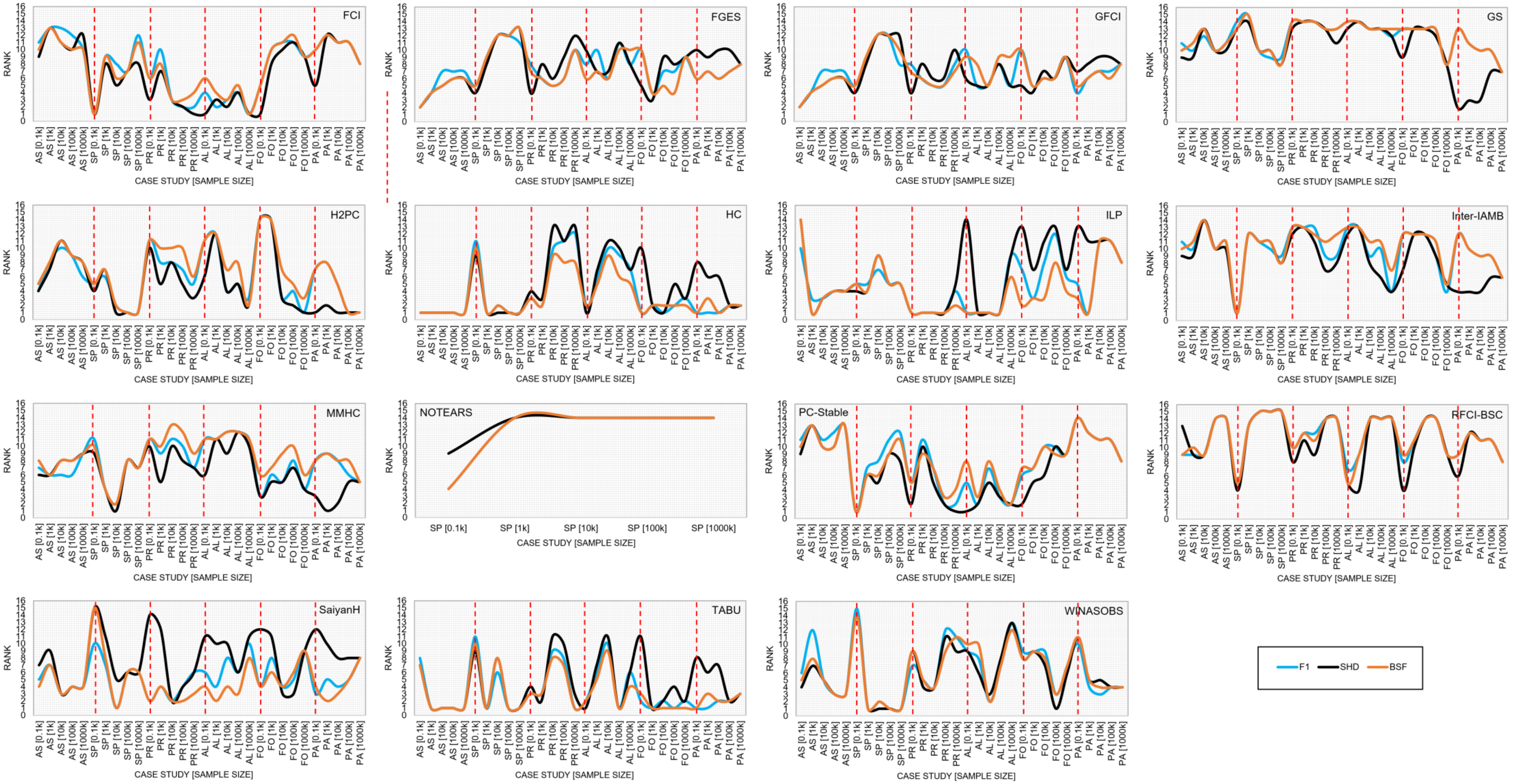

**Fig D1.** Overall ranking of the algorithms (as defined in Table 8) ordered by case study (separated by a red dotted line). Each case study is ordered by the number of nodes in the true graph, and each experiment within a case study is ordered by sample size. A lower Rank indicates better performance. These results are based on experiment N (i.e., no noise). In NOTEARS, the F1 rankings are identical to the BSF rankings.

**Table D6.** The minimum and maximum scores for each case study and sample size $n$ combination, as determined by each of the three metrics over all 15 algorithms for the $N$ experiments (without data noise). Note that a higher SHD score indicates lower performance. The minimums and maximums are based on all the experiments shown in Table 11 that do not include a failure F.

| Min/Max performance | $n$ | F1 | | | | | | SHD | | | | | | BSF | | | | | |
|---|---|---|---|---|---|---|---|---|---|---|---|---|---|---|---|---|---|---|---|
| | | Asia | Spor | Prop | Alar | Form | Path | Asia | Spor | Prop | Alar | Form | Path | Asia | Spor | Prop | Alar | Form | Path |
| Min | 0.1k | 0.27 | 0.00 | 0.14 | 0.16 | 0.17 | 0.04 | 9 | 16.5 | 35 | 47.5 | 205 | 317.5 | 0.18 | -0.02 | 0.08 | 0.10 | 0.10 | 0.02 |
| | 1k | 0.43 | 0.16 | 0.26 | 0.30 | 0.27 | 0.04 | 5 | 14.5 | 26 | 38.5 | 117 | 279.5 | 0.38 | 0.05 | 0.16 | 0.20 | 0.18 | 0.02 |
| | 10k | 0.54 | 0.35 | 0.29 | 0.45 | 0.40 | 0.08 | 5.5 | 14 | 26 | 33 | 100 | 291.5 | 0.39 | 0.14 | 0.19 | 0.34 | 0.28 | 0.04 |
| | 100k | 0.57 | 0.35 | 0.49 | 0.60 | 0.35 | 0.13 | 4 | 14 | 20.5 | 24.5 | 132.5 | 317.5 | 0.50 | 0.14 | 0.37 | 0.47 | 0.25 | 0.07 |
| | 1000k | 0.60 | 0.35 | 0.59 | 0.59 | 0.41 | 0.17 | 4.5 | 14 | 22 | 30 | 102 | 176 | 0.51 | 0.14 | 0.48 | 0.51 | 0.30 | 0.10 |
| Max | 0.1k | 0.79 | 0.35 | 0.59 | 0.54 | 0.42 | 0.24 | 2.5 | 11.5 | 18.5 | 32 | 111 | 189 | 0.69 | 0.23 | 0.49 | 0.61 | 0.34 | 0.16 |
| | 1k | 0.93 | 0.75 | 0.88 | 0.86 | 0.71 | 0.27 | 1 | 6 | 6.5 | 11.5 | 64 | 185 | 0.88 | 0.60 | 0.79 | 0.87 | 0.65 | 0.25 |
| | 10k | 1.00 | 0.77 | 0.98 | 0.95 | 0.79 | 0.38 | 0 | 6 | 1 | 4.5 | 47 | 167 | 1.00 | 0.62 | 0.97 | 0.94 | 0.79 | 0.33 |
| | 100k | 1.00 | 1.00 | 0.95 | 0.96 | 0.80 | 0.67 | 0 | 0 | 2 | 2.5 | 46.5 | 100.5 | 1.00 | 1.00 | 0.94 | 0.95 | 0.85 | 0.52 |
| | 1000k | 1.00 | 1.00 | 0.85 | 0.95 | 0.82 | 0.76 | 0 | 0 | 7 | 2.5 | 41.5 | 74.5 | 1.00 | 1.00 | 0.83 | 0.95 | 0.88 | 0.65 |

**Table D7.** Relative overall performance of the algorithms for each case study and sample size $n$ combination, as determined by each of the three metrics over the $N$ experiments (without data noise). The performance is measured relative to the min/max values depicted in Table 10. An F represents *at least one* failed attempt by the algorithm to produce a graph for the particular case study and sample size combination (refer to Appendix C).

| Algorithm | $n$ | F1 | | | | | | SHD | | | | | | BSF | | | | | |
|---|---|---|---|---|---|---|---|---|---|---|---|---|---|---|---|---|---|---|---|
| | | Asia | Spor | Prop | Alar | Form | Path | Asia | Spor | Prop | Alar | Form | Path | Asia | Spor | Prop | Alar | Form | Path |
| FCI | 0.1k | 0% | 100% | 42% | 76% | 61% | 19% | 38% | 100% | 67% | 100% | 100% | 92% | 3% | 100% | 35% | 44% | 46% | 15% |
| | 1k | 0% | 51% | 36% | 85% | 49% | F | 0% | 59% | 31% | 87% | 58% | F | 0% | 57% | 35% | 74% | 41% | F |
| | 10k | 7% | 53% | 68% | 85% | 48% | F | 27% | 94% | 60% | 88% | 54% | F | 18% | 88% | 66% | 81% | 41% | F |
| | 100k | 0% | 75% | 76% | 82% | 49% | F | 0% | 82% | 73% | 82% | 55% | F | 0% | 81% | 72% | 86% | 46% | F |
| | 1000k | 17% | 34% | 96% | 100% | F | F | 11% | 54% | 100% | 100% | F | F | 13% | 49% | 84% | 100% | F | F |
| FGES | 0.1k | 60% | 67% | 37% | 65% | 50% | 63% | 69% | 70% | 55% | 58% | 95% | 83% | 51% | 61% | 34% | 44% | 39% | 61% |
| | 1k | 74% | 51% | 50% | 70% | 75% | 65% | 75% | 59% | 28% | 67% | 85% | 36% | 75% | 57% | 51% | 68% | 64% | 51% |
| | 10k | 60% | 23% | 63% | 68% | 71% | 34% | 73% | 44% | 48% | 61% | 55% | 0% | 69% | 37% | 61% | 70% | 69% | 36% |
| | 100k | 56% | 21% | 56% | 62% | 80% | 7% | 63% | 25% | 30% | 45% | 76% | 0% | 63% | 23% | 67% | 75% | 72% | 14% |
| | 1000k | 53% | 49% | 28% | 64% | F | F | 67% | 43% | 10% | 56% | F | F | 62% | 44% | 51% | 73% | F | F |
| GFCI | 0.1k | 60% | 67% | 37% | 58% | 50% | 67% | 69% | 70% | 55% | 71% | 95% | 85% | 51% | 61% | 34% | 37% | 39% | 65% |
| | 1k | 74% | 51% | 50% | 72% | 74% | 74% | 75% | 59% | 28% | 72% | 84% | 42% | 75% | 57% | 51% | 66% | 63% | 59% |
| | 10k | 60% | 23% | 63% | 77% | 72% | 31% | 73% | 44% | 48% | 72% | 60% | 3% | 69% | 37% | 61% | 74% | 67% | 33% |
| | 100k | 56% | 21% | 65% | 72% | 83% | 7% | 63% | 25% | 46% | 64% | 82% | 0% | 63% | 23% | 69% | 76% | 72% | 14% |
| | 1000k | 53% | 53% | 46% | 75% | F | F | 67% | 50% | 40% | 71% | F | F | 62% | 50% | 59% | 74% | F | F |
| GS | 0.1k | 0% | 34% | 0% | 0% | 0% | 0% | 38% | 50% | 39% | 32% | 80% | 99% | 3% | 35% | 0% | 0% | 0% | 0% |
| | 1k | 31% | 0% | 0% | 0% | 0% | 0% | 13% | 0% | 0% | 0% | 0% | 94% | 13% | 0% | 0% | 0% | 0% | 0% |
| | 10k | 10% | 42% | 0% | 0% | 0% | 0% | 18% | 63% | 0% | 0% | 0% | 83% | 8% | 53% | 0% | 0% | 0% | 0% |
| | 100k | 10% | 44% | 0% | 0% | 0% | 0% | 0% | 46% | 0% | 0% | 28% | 63% | 0% | 44% | 0% | 0% | 0% | 0% |
| | 1000k | 11% | 51% | 0% | 5% | 0% | 0% | 22% | 54% | 33% | 16% | 0% | 0% | 10% | 52% | 0% | 0% | 0% | 0% |
| H2PC | 0.1k | 53% | 67% | 17% | 51% | F | 44% | 62% | 70% | 48% | 71% | F | 100% | 39% | 61% | 12% | 25% | F | 35% |
| | 1k | 47% | 70% | 43% | 52% | F | 46% | 25% | 65% | 33% | 48% | F | 98% | 25% | 64% | 33% | 39% | F | 26% |
| | 10k | 28% | 96% | 54% | 77% | 87% | 79% | 27% | 100% | 46% | 77% | 89% | 100% | 18% | 95% | 48% | 70% | 66% | 55% |
| | 100k | 46% | 100% | 52% | 81% | 96% | 100% | 25% | 100% | 51% | 80% | 99% | 100% | 25% | 100% | 49% | 80% | 72% | 100% |
| | 1000k | 64% | 100% | 83% | 94% | 100% | 100% | 56% | 100% | 93% | 96% | 100% | 100% | 49% | 100% | 75% | 95% | 79% | 100% |
| HC | 0.1k | 100% | 50% | 55% | 94% | 100% | 100% | 100% | 60% | 55% | 100% | 80% | 85% | 100% | 48% | 48% | 64% | 100% | 100% |
| | 1k | 100% | 100% | 65% | 71% | 99% | 100% | 100% | 100% | 49% | 67% | 99% | 49% | 100% | 100% | 67% | 71% | 99% | 82% |
| | 10k | 100% | 96% | 46% | 56% | 100% | 100% | 100% | 100% | 20% | 46% | 100% | 62% | 100% | 95% | 54% | 67% | 99% | 100% |
| | 100k | 100% | 100% | 32% | 66% | 97% | 80% | 100% | 100% | 3% | 55% | 90% | 75% | 100% | 100% | 54% | 84% | 98% | 97% |
| | 1000k | 100% | 100% | 12% | 66% | 92% | 85% | 100% | 100% | 0% | 58% | 79% | 60% | 100% | 100% | 54% | 89% | 99% | 95% |
| ILP | 0.1k | 16% | 67% | 100% | 100% | 57% | 65% | 0% | 70% | 100% | 0% | 0% | 0% | 0% | 61% | 100% | 100% | 94% | 89% |
| | 1k | 88% | 86% | 100% | 100% | 83% | 100% | 100% | 88% | 100% | 100% | 68% | 0% | 100% | 88% | 100% | 100% | 88% | 100% |
| | 10k | 73% | 70% | 100% | 100% | 71% | F | 82% | 75% | 100% | 100% | 43% | F | 80% | 71% | 100% | 100% | 83% | F |
| | 100k | 71% | 90% | 100% | 100% | 38% | F | 75% | 93% | 100% | 100% | 0% | F | 75% | 92% | 100% | 100% | 61% | F |
| | 1000k | 69% | 90% | 94% | 60% | 43% | F | 78% | 93% | 90% | 55% | 14% | F | 74% | 92% | 95% | 84% | 54% | F |
| Inter-IAMB | 0.1k | 0% | 100% | 10% | 20% | 17% | 12% | 38% | 100% | 45% | 48% | 89% | 94% | 3% | 100% | 8% | 11% | 13% | 10% |
| | 1k | 31% | 38% | 15% | 31% | 22% | 12% | 13% | 41% | 13% | 35% | 25% | 93% | 13% | 39% | 13% | 25% | 19% | 7% |
| | 10k | 0% | 28% | 39% | 58% | 31% | 3% | 0% | 50% | 30% | 58% | 29% | 82% | 0% | 43% | 35% | 50% | 28% | 1% |
| | 100k | 10% | 38% | 36% | 64% | 65% | 5% | 0% | 43% | 35% | 66% | 74% | 64% | 0% | 40% | 32% | 66% | 51% | 4% |
| | 1000k | 11% | 51% | 44% | 78% | 55% | 7% | 22% | 54% | 67% | 84% | 40% | 5% | 10% | 52% | 38% | 78% | 51% | 5% |
| MMHC | 0.1k | 35% | 50% | 17% | 51% | 65% | 35% | 54% | 60% | 48% | 71% | 99% | 98% | 27% | 48% | 12% | 25% | 45% | 27% |
| | 1k | 67% | 91% | 41% | 59% | 69% | 43% | 50% | 88% | 33% | 57% | 71% | 100% | 50% | 88% | 33% | 44% | 52% | 24% |
| | 10k | 69% | 96% | 41% | 51% | 79% | 21% | 64% | 100% | 32% | 49% | 76% | 87% | 59% | 95% | 33% | 41% | 59% | 12% |
| | 100k | 67% | 62% | 34% | 37% | 74% | 11% | 50% | 57% | 32% | 32% | 78% | 65% | 50% | 53% | 29% | 32% | 51% | 7% |
| | 1000k | 50% | 83% | 65% | 37% | 55% | 15% | 33% | 79% | 77% | 45% | 45% | 10% | 38% | 77% | 52% | 26% | 38% | 9% |
| NOTEARS | 0.1k | n/a | 95% | n/a | n/a | n/a | n/a | n/a | 60% | n/a | n/a | n/a | n/a | n/a | 70% | n/a | n/a | n/a | n/a |
| | 1k | n/a | 28% | n/a | n/a | n/a | n/a | n/a | 0% | n/a | n/a | n/a | n/a | n/a | 11% | n/a | n/a | n/a | n/a |
| | 10k | n/a | 0% | n/a | n/a | n/a | n/a | n/a | 0% | n/a | n/a | n/a | n/a | n/a | 0% | n/a | n/a | n/a | n/a |
| | 100k | n/a | 0% | n/a | n/a | n/a | n/a | n/a | 0% | n/a | n/a | n/a | n/a | n/a | 0% | n/a | n/a | n/a | n/a |
| | 1000k | n/a | 0% | n/a | n/a | n/a | n/a | n/a | 0% | n/a | n/a | n/a | n/a | n/a | 0% | n/a | n/a | n/a | n/a |
| PC-Stable | 0.1k | 0% | 100% | 45% | 71% | 60% | F | 38% | 100% | 70% | 100% | 99% | F | 3% | 100% | 39% | 40% | 44% | F |
| | 1k | 0% | 68% | 35% | 85% | 64% | F | 0% | 76% | 26% | 91% | 71% | F | 0% | 76% | 35% | 74% | 52% | F |
| | 10k | 23% | 53% | 62% | 67% | 57% | F | 36% | 94% | 54% | 72% | 61% | F | 29% | 88% | 60% | 68% | 47% | F |
| | 100k | 0% | 34% | 76% | 85% | 64% | F | 0% | 54% | 73% | 84% | 65% | F | 0% | 49% | 72% | 88% | 55% | F |
| | 1000k | 0% | 34% | 96% | 97% | F | F | 0% | 54% | 100% | 96% | F | F | 0% | 49% | 84% | 95% | F | F |
| RFCI-BSC | 0.1k | 16% | 67% | 26% | 69% | 52% | 21% | 38% | 70% | 52% | 82% | 97% | 86% | 12% | 61% | 22% | 45% | 39% | 18% |
| | 1k | 31% | 29% | 28% | 70% | 36% | F | 13% | 32% | 25% | 74% | 41% | F | 23% | 29% | 27% | 63% | 29% | F |
| | 10k | 38% | F | 40% | F | F | F | 52% | F | 34% | F | F | F | 47% | F | 37% | F | F | F |
| | 100k | F | F | F | F | F | F | F | F | F | F | F | F | F | F | F | F | F | F |
| | 1000k | F | F | F | F | F | F | F | F | F | F | F | F | F | F | F | F | F | F |
| SaiyanH | 0.1k | 51% | 62% | 56% | 70% | 62% | 72% | 46% | 0% | 0% | 52% | 63% | 36% | 44% | 0% | 62% | 52% | 68% | 86% |
| | 1k | 51% | 68% | 53% | 76% | 59% | 92% | 13% | 53% | 15% | 65% | 41% | 3% | 42% | 58% | 62% | 75% | 60% | 88% |
| | 10k | 73% | 100% | 74% | 60% | 85% | 80% | 82% | 94% | 64% | 47% | 89% | 59% | 80% | 100% | 78% | 76% | 74% | 77% |
| | 100k | 71% | 85% | 69% | 77% | 89% | 47% | 75% | 89% | 59% | 66% | 94% | 43% | 75% | 88% | 78% | 91% | 72% | 63% |
| | 1000k | 69% | 85% | 77% | 52% | F | F | 78% | 89% | 83% | 38% | F | F | 74% | 88% | 86% | 75% | F | F |
| TABU | 0.1k | 30% | 50% | 55% | 94% | 86% | 100% | 46% | 60% | 55% | 100% | 70% | 85% | 29% | 48% | 48% | 64% | 90% | 100% |
| | 1k | 100% | 100% | 64% | 71% | 100% | 100% | 100% | 100% | 51% | 67% | 100% | 49% | 100% | 100% | 65% | 71% | 100% | 82% |
| | 10k | 100% | 79% | 51% | 56% | 99% | 95% | 100% | 81% | 30% | 46% | 99% | 59% | 100% | 78% | 57% | 67% | 100% | 97% |
| | 100k | 100% | 100% | 40% | 100% | 100% | 80% | 100% | 100% | 14% | 100% | 94% | 75% | 100% | 100% | 60% | 100% | 100% | 97% |
| | 1000k | 100% | 100% | 100% | 73% | 93% | 84% | 100% | 100% | 93% | 67% | 80% | 60% | 100% | 100% | 100% | 92% | 100% | 94% |
| WINASOBS | 0.1k | 44% | 0% | 38% | 60% | 51% | 19% | 62% | 30% | 52% | 61% | 88% | 78% | 39% | 9% | 30% | 35% | 41% | 15% |
| | 1k | 28% | 100% | 51% | 70% | 52% | 92% | 25% | 100% | 38% | 70% | 52% | 69% | 25% | 100% | 52% | 62% | 47% | 69% |
| | 10k | 71% | 96% | 65% | 86% | 67% | 84% | 73% | 100% | 58% | 86% | 60% | 71% | 69% | 95% | 64% | 82% | 62% | 75% |
| | 100k | 86% | 100% | 28% | 73% | 96% | 68% | 88% | 100% | 3% | 64% | 100% | 73% | 88% | 100% | 49% | 80% | 85% | 74% |
| | 1000k | 85% | 100% | 28% | 0% | 24% | 67% | 89% | 100% | 50% | 0% | 30% | 48% | 87% | 100% | 48% | 23% | 20% | 61% |

## APPENDIX E: SUPPLEMENTARY RESULTS, WITH DATA NOISE

**Table E1.** Average (A) and overall (O) ranked performance for each algorithm (highlighted in yellow backcolour), over all case studies and sample sizes in noisy-based experiments M5, M10, I5, I10, and S5, as determined by each of the three metrics.

| | M5 | | | | | | M10 | | | | | | I5 | | | | | | I10 | | | | | | S5 | | | | | |
|---|---|---|---|---|---|---|---|---|---|---|---|---|---|---|---|---|---|---|---|---|---|---|---|---|---|---|---|---|---|---|
| | F1 | | SHD | | BSF | | F1 | | SHD | | BSF | | F1 | | SHD | | BSF | | F1 | | SHD | | BSF | | F1 | | SHD | | BSF | |
| Algorithm | A | O | A | O | A | O | A | O | A | O | A | O | A | O | A | O | A | O | A | O | A | O | A | O | A | O | A | O | A | O |
| FCI | 9.0 | 11 | 9.5 | 12 | 8.3 | 11 | 9.1 | 11 | 9.3 | 12 | 8.3 | 11 | 10 | 12 | 10.4 | 13 | 9.5 | 11 | 9.5 | 12 | 8.7 | 12 | 9.2 | 12 | 7.9 | 10 | 6.8 | 7 | 8.1 | 10 |
| FGES | 6.5 | 6 | 6.5 | 8 | 6.6 | 6 | 6.8 | 8 | 6.7 | 7 | 7.2 | 7 | 7.2 | 8 | 6.3 | 5 | 7.5 | 8 | 7.9 | 8 | 6.8 | 6 | 8.7 | 10 | 7.2 | 8 | 8.4 | 12 | 6.9 | 6 |
| GFCI | 6.5 | 7 | 6.3 | 7 | 6.8 | 7 | 7.0 | 9 | 6.7 | 6 | 7.4 | 9 | 7.4 | 9 | 6.4 | 6 | 7.9 | 10 | 8.3 | 10 | 7 | 8 | 8.9 | 11 | 7.2 | 8 | 7.9 | 9 | 7.7 | 9 |
| GS | 12.0 | 15 | 9.9 | 13 | 11.9 | 13 | 11.6 | 13 | 9.4 | 13 | 11.5 | 13 | 11.6 | 14 | 8.6 | 11 | 11.5 | 14 | 11.1 | 14 | 8.6 | 11 | 11.1 | 14 | 12.2 | 14 | 10.2 | 13 | 12.2 | 14 |
| H2PC | 5.3 | 4 | 4.6 | 3 | 6.2 | 5 | 5.3 | 4 | 4.7 | 3 | 5.9 | 5 | 4.6 | 3 | 4.7 | 3 | 5.2 | 4 | 6.0 | 6 | 6.5 | 4 | 6.2 | 7 | 5.9 | 5 | 4.0 | 1 | 7.0 | 7 |
| HC | 3.6 | 1 | 4.1 | 1 | 3.2 | 1 | 3.0 | 1 | 3.6 | 1 | 2.8 | 1 | 3.7 | 1 | 5.5 | 4 | 2.9 | 1 | 3.4 | 1 | 5.9 | 3 | 2.4 | 1 | 4.4 | 2 | 6.6 | 6 | 3.6 | 2 |
| ILP | 3.9 | 2 | 5.8 | 6 | 3.4 | 3 | 5.3 | 3 | 7.1 | 8 | 4.2 | 3 | 5.8 | 7 | 7.1 | 8 | 4.4 | 3 | 6.0 | 6 | 7.5 | 10 | 5.4 | 4 | 5.4 | 4 | 6.8 | 8 | 4.6 | 4 |
| Inter-IAMB | 10.8 | 12 | 8.4 | 9 | 10.8 | 12 | 10.8 | 12 | 8.4 | 10 | 10.8 | 12 | 9.8 | 11 | 7.6 | 10 | 10.1 | 12 | 8.3 | 11 | 6.9 | 7 | 8.6 | 9 | 10.2 | 12 | 8.2 | 10 | 10.5 | 12 |
| MMHC | 7.1 | 9 | 5.2 | 4 | 8.1 | 10 | 6.6 | 6 | 4.9 | 4 | 7.7 | 10 | 5 | 5 | 3.5 | 1 | 6.4 | 7 | 4.2 | 3 | 3.4 | 1 | 5.7 | 5 | 8.3 | 11 | 5.7 | 3 | 9.0 | 11 |
| NOTEARS | 11.4 | 13 | 13.4 | 15 | 13.6 | 15 | 11.6 | 13 | 12.0 | 15 | 12.2 | 15 | 11.2 | 13 | 10.8 | 14 | 10.8 | 13 | 11 | 13 | 10 | 14 | 9.6 | 13 | n/a | n/a | n/a | n/a | n/a | n/a |
| PC-Stable | 8.1 | 10 | 8.6 | 10 | 7.2 | 9 | 7.7 | 10 | 8.7 | 11 | 7.4 | 8 | 8.4 | 10 | 9.2 | 12 | 7.8 | 9 | 8.2 | 9 | 8.9 | 13 | 8 | 8 | 6.9 | 7 | 5.9 | 4 | 7.0 | 7 |
| RFCI-BSC | 11.7 | 14 | 11.7 | 14 | 11.9 | 14 | 11.8 | 15 | 11.9 | 14 | 11.9 | 14 | 12 | 15 | 11 | 15 | 12 | 15 | 11.5 | 15 | 10.8 | 15 | 11.4 | 15 | 11.4 | 13 | 10.2 | 14 | 11.4 | 13 |
| SaiyanH | 5.9 | 5 | 8.9 | 11 | 6.1 | 4 | 5.3 | 4 | 8.2 | 9 | 5.2 | 4 | 4.9 | 4 | 7.2 | 9 | 5.4 | 5 | 5 | 4 | 7.1 | 9 | 4.7 | 3 | 4.9 | 3 | 8.2 | 10 | 4.1 | 3 |
| TABU | 4.0 | 3 | 4.4 | 2 | 3.3 | 2 | 3.3 | 2 | 3.9 | 2 | 3.1 | 2 | 4.2 | 2 | 6.4 | 6 | 3.5 | 2 | 4.2 | 2 | 6.6 | 5 | 3.5 | 2 | 3.3 | 1 | 5.2 | 2 | 2.8 | 1 |
| WINASOBS | 6.8 | 8 | 5.5 | 5 | 7.1 | 8 | 6.6 | 6 | 5.0 | 5 | 6.9 | 6 | 5.6 | 6 | 4.6 | 2 | 6.1 | 6 | 5.2 | 5 | 3.7 | 2 | 5.7 | 6 | 5.9 | 5 | 6.0 | 5 | 5.6 | 5 |

**Table E2.** Average (A) and overall (O) ranked performance for each algorithm, over all case studies and sample sizes in noisy-based experiments S10, L5, L10, cMI, and cMS, as determined by each of the three metrics.

| | S10 | | | | | | L5 | | | | | | L10 | | | | | | cMI | | | | | | cMS | | | | | |
|---|---|---|---|---|---|---|---|---|---|---|---|---|---|---|---|---|---|---|---|---|---|---|---|---|---|---|---|---|---|---|
| | F1 | | SHD | | BSF | | F1 | | SHD | | BSF | | F1 | | SHD | | BSF | | F1 | | SHD | | BSF | | F1 | | SHD | | BSF | |
| Algorithm | A | O | A | O | A | O | A | O | A | O | A | O | A | O | A | O | A | O | A | O | A | O | A | O | A | O | A | O | A | O |
| FCI | 8.3 | 11 | 7.2 | 8 | 8.3 | 11 | 6.9 | 8 | 5.8 | 4 | 7.1 | 7 | 7.7 | 10 | 6.7 | 8 | 7.4 | 10 | 8.9 | 11 | 10.7 | 13 | 8.0 | 10 | 8.9 | 11 | 9.4 | 12 | 8.1 | 10 |
| FGES | 7.6 | 10 | 8.2 | 12 | 7.5 | 8 | 7.2 | 10 | 8.4 | 11 | 7.2 | 8 | 6.9 | 7 | 7.3 | 10 | 7.1 | 8 | 7.7 | 8 | 6.3 | 6 | 7.8 | 9 | 6.8 | 6 | 7.3 | 8 | 7.2 | 7 |
| GFCI | 7.4 | 9 | 7.4 | 9 | 7.6 | 9 | 6.7 | 7 | 7.3 | 9 | 7.0 | 6 | 6.9 | 8 | 7.0 | 9 | 6.9 | 7 | 7.9 | 9 | 6.5 | 7 | 8.1 | 11 | 7.0 | 8 | 7.2 | 7 | 7.6 | 9 |
| GS | 12.0 | 13 | 10.2 | 13 | 12.0 | 13 | 12.2 | 14 | 10.4 | 14 | 12.2 | 14 | 11.9 | 14 | 10.8 | 14 | 12.0 | 14 | 11.4 | 13 | 9.3 | 11 | 11.2 | 13 | 12.0 | 15 | 9.5 | 13 | 11.8 | 14 |
| H2PC | 5.8 | 5 | 4.6 | 2 | 6.5 | 5 | 6.3 | 6 | 5.4 | 3 | 7.6 | 10 | 6.4 | 5 | 5.2 | 3 | 6.8 | 6 | 4.7 | 3 | 4.2 | 3 | 5.2 | 4 | 5.8 | 4 | 4.5 | 3 | 6.6 | 6 |
| HC | 4.1 | 2 | 5.9 | 5 | 3.3 | 1 | 4.4 | 2 | 6.0 | 5 | 3.5 | 2 | 3.6 | 2 | 5.0 | 2 | 3.1 | 2 | 3.3 | 1 | 4.1 | 2 | 3.0 | 1 | 3.3 | 2 | 3.8 | 2 | 3.0 | 2 |
| ILP | 5.1 | 4 | 6.8 | 6 | 4.4 | 4 | 4.9 | 3 | 6.4 | 6 | 4.2 | 3 | 4.6 | 3 | 5.7 | 4 | 4.0 | 3 | 5.1 | 5 | 6.6 | 8 | 4.5 | 3 | 5.3 | 3 | 7.2 | 6 | 4.1 | 3 |
| Inter-IAMB | 9.7 | 12 | 8.1 | 11 | 9.9 | 12 | 10.0 | 12 | 7.9 | 10 | 10.4 | 12 | 10.1 | 12 | 8.7 | 12 | 10.1 | 12 | 9.6 | 12 | 7.9 | 10 | 9.8 | 12 | 10.2 | 12 | 7.8 | 9 | 10.4 | 12 |
| MMHC | 6.9 | 7 | 4.4 | 1 | 8.2 | 10 | 8.7 | 11 | 6.8 | 7 | 9.5 | 11 | 7.8 | 11 | 6.2 | 5 | 8.1 | 11 | 5.3 | 6 | 3.4 | 1 | 6.6 | 6 | 7.1 | 9 | 4.9 | 4 | 8.4 | 11 |
| NOTEARS | 12.2 | 15 | 14.2 | 15 | 13.8 | 15 | n/a | n/a | n/a | n/a | n/a | n/a | 12.2 | 15 | 14.2 | 15 | 14.0 | 15 | 11.4 | 14 | 12.6 | 15 | 12.4 | 15 | 11.6 | 13 | 13.0 | 15 | 12.0 | 15 |
| PC-Stable | 6.8 | 6 | 5.2 | 3 | 7.2 | 7 | 6.2 | 5 | 5.3 | 2 | 6.7 | 5 | 7.4 | 9 | 6.3 | 6 | 7.2 | 9 | 8.0 | 10 | 9.6 | 12 | 7.0 | 7 | 7.4 | 10 | 8.5 | 10 | 6.3 | 5 |
| RFCI-BSC | 12.0 | 13 | 11.5 | 14 | 12.0 | 13 | 10.9 | 13 | 10.1 | 13 | 11.0 | 13 | 11.1 | 13 | 10.6 | 13 | 11.3 | 13 | 11.6 | 15 | 11.7 | 14 | 11.8 | 14 | 11.7 | 14 | 11.4 | 14 | 11.5 | 13 |
| SaiyanH | 4.7 | 3 | 7.6 | 10 | 3.9 | 3 | 5.9 | 4 | 8.5 | 12 | 4.6 | 4 | 5.3 | 4 | 7.4 | 11 | 5.2 | 4 | 4.9 | 4 | 7.3 | 9 | 5.2 | 5 | 6.3 | 5 | 9.1 | 11 | 6.1 | 4 |
| TABU | 3.9 | 1 | 5.6 | 4 | 3.3 | 1 | 3.3 | 1 | 4.9 | 1 | 2.7 | 1 | 3.2 | 1 | 4.6 | 1 | 3.0 | 1 | 3.9 | 2 | 4.8 | 4 | 3.4 | 2 | 2.8 | 1 | 3.5 | 1 | 2.8 | 1 |
| WINASOBS | 7.0 | 8 | 6.8 | 7 | 6.9 | 6 | 7.2 | 9 | 6.9 | 8 | 7.2 | 8 | 6.8 | 6 | 6.4 | 7 | 6.6 | 5 | 7.2 | 7 | 5.3 | 5 | 7.5 | 8 | 6.8 | 7 | 5.7 | 5 | 7.5 | 8 |

**Table E3.** Average (A) and overall (O) ranked performance for each algorithm, over all case studies and sample sizes in noisy-based experiments cML, cIS, cIL, cSL, and cMISL, as determined by each of the three metrics.

| | cML | | | | | | cIS | | | | | | cIL | | | | | | cSL | | | | | | cMISL | | | | | |
|---|---|---|---|---|---|---|---|---|---|---|---|---|---|---|---|---|---|---|---|---|---|---|---|---|---|---|---|---|---|---|
| | F1 | | SHD | | BSF | | F1 | | SHD | | BSF | | F1 | | SHD | | BSF | | F1 | | SHD | | BSF | | F1 | | SHD | | BSF | |
| Algorithm | A | O | A | O | A | O | A | O | A | O | A | O | A | O | A | O | A | O | A | O | A | O | A | O | A | O | A | O | A | O |
| FCI | 8.8 | 11 | 9.4 | 12 | 8.0 | 11 | 9.4 | 11 | 10.0 | 13 | 8.5 | 11 | 9.4 | 12 | 9.8 | 13 | 8.7 | 11 | 7.0 | 8 | 6.2 | 6 | 7.2 | 8 | 9.3 | 12 | 10.4 | 13 | 8.9 | 11 |
| FGES | 6.0 | 5 | 6.5 | 8 | 6.5 | 6 | 6.6 | 8 | 6.0 | 6 | 7.2 | 7 | 7.2 | 8 | 6.4 | 7 | 7.8 | 9 | 8.1 | 11 | 8.6 | 12 | 7.6 | 9 | 7.6 | 8 | 7.2 | 8 | 7.8 | 8 |
| GFCI | 6.2 | 7 | 6.4 | 7 | 6.7 | 7 | 7.1 | 9 | 6.2 | 7 | 7.7 | 10 | 7.4 | 9 | 6.0 | 4 | 7.8 | 9 | 7.8 | 10 | 7.9 | 10 | 7.6 | 9 | 8.1 | 9 | 7.5 | 10 | 8.3 | 10 |
| GS | 12.0 | 15 | 9.8 | 13 | 11.9 | 14 | 12.0 | 15 | 9.0 | 12 | 12.0 | 15 | 11.4 | 13 | 8.8 | 12 | 11.2 | 13 | 11.8 | 14 | 10.4 | 14 | 12.0 | 14 | 11.0 | 14 | 8.4 | 11 | 10.9 | 14 |
| H2PC | 6.2 | 6 | 5.1 | 4 | 6.5 | 5 | 5.9 | 6 | 5.8 | 5 | 6.6 | 6 | 5.9 | 7 | 6.0 | 4 | 6.1 | 5 | 5.6 | 5 | 4.8 | 1 | 6.3 | 5 | 5.1 | 5 | 4.3 | 3 | 5.3 | 5 |
| HC | 3.4 | 1 | 3.7 | 1 | 3.3 | 1 | 3.6 | 2 | 5.4 | 3 | 2.6 | 2 | 3.3 | 1 | 5.0 | 3 | 2.6 | 1 | 3.9 | 2 | 5.0 | 3 | 3.5 | 2 | 3.0 | 1 | 4.2 | 2 | 2.8 | 1 |
| ILP | 4.2 | 3 | 5.7 | 6 | 3.5 | 3 | 6.2 | 7 | 8.1 | 10 | 5.1 | 3 | 5.8 | 6 | 7.4 | 10 | 4.9 | 3 | 5.0 | 3 | 6.3 | 7 | 4.2 | 3 | 4.9 | 4 | 6.3 | 6 | 4.2 | 3 |
| Inter-IAMB | 10.0 | 12 | 7.5 | 9 | 10.0 | 12 | 9.4 | 12 | 7.3 | 8 | 10.0 | 12 | 9.3 | 11 | 7.3 | 9 | 9.4 | 12 | 9.7 | 12 | 8.3 | 11 | 10.1 | 12 | 9.0 | 11 | 7.0 | 7 | 9.1 | 12 |
| MMHC | 7.3 | 9 | 4.9 | 3 | 7.9 | 10 | 5.4 | 3 | 3.7 | 1 | 7.3 | 9 | 5.0 | 4 | 3.2 | 1 | 6.1 | 5 | 7.5 | 9 | 6.0 | 5 | 8.3 | 11 | 5.6 | 6 | 3.4 | 1 | 6.5 | 6 |
| NOTEARS | 11.2 | 13 | 13.2 | 15 | 13.2 | 15 | 11.4 | 13 | 12.0 | 15 | 11.4 | 13 | 11.8 | 14 | 13.0 | 15 | 12.8 | 15 | 12.4 | 15 | 14.6 | 15 | 14.4 | 15 | 12.0 | 15 | 13.8 | 15 | 12.4 | 15 |
| PC-Stable | 7.9 | 10 | 9.2 | 11 | 7.5 | 9 | 8.0 | 10 | 8.9 | 11 | 7.2 | 8 | 7.8 | 10 | 8.8 | 11 | 6.8 | 8 | 6.7 | 6 | 5.3 | 4 | 6.5 | 6 | 8.4 | 10 | 9.7 | 12 | 7.7 | 7 |
| RFCI-BSC | 11.5 | 14 | 11.7 | 14 | 11.6 | 13 | 11.8 | 14 | 11.1 | 14 | 11.7 | 14 | 11.8 | 15 | 11.2 | 14 | 11.7 | 14 | 11.1 | 13 | 10.2 | 13 | 11.1 | 13 | 10.7 | 13 | 10.8 | 14 | 10.7 | 13 |
| SaiyanH | 5.8 | 4 | 8.7 | 10 | 5.8 | 4 | 5.5 | 4 | 7.6 | 9 | 5.8 | 4 | 4.7 | 3 | 7.1 | 8 | 5.6 | 4 | 5.4 | 4 | 7.8 | 9 | 4.9 | 4 | 4.6 | 3 | 7.2 | 8 | 4.9 | 4 |
| TABU | 3.4 | 1 | 4.0 | 2 | 3.3 | 1 | 3.5 | 1 | 5.4 | 3 | 2.5 | 1 | 4.3 | 2 | 6.2 | 6 | 3.4 | 2 | 3.6 | 1 | 4.8 | 1 | 3.2 | 1 | 3.4 | 2 | 4.7 | 4 | 3.3 | 2 |
| WINASOBS | 7.2 | 8 | 5.6 | 5 | 7.2 | 8 | 5.6 | 5 | 4.6 | 2 | 6.0 | 5 | 5.7 | 5 | 4.1 | 2 | 6.5 | 7 | 7.0 | 7 | 6.8 | 8 | 7.0 | 7 | 7.5 | 7 | 5.4 | 5 | 8.0 | 9 |

**Table E4.** Performance of the algorithms on each noisy experiment relative to the N case (i.e., no noise). The results are presented per case study per sample size (where the sample size of 0.1 corresponds to 0.1k data samples and so forth.), per scoring metric S and per experiment E. The table values represent relative percentage change with respect to the results of experiment N (i.e., after adding each type of data noise in the data). Note that in the case of SHD, a decrease in performance corresponds to an increase in the SHD score (which can increase by more than 100%), and vice versa.

| | | ALARM | | | | | ASIA | | | | | PATHFINDER | | | | | PROPERTY | | | | | SPORTS | | | | | FORMED | | | | |
|---|---|---|---|---|---|---|---|---|---|---|---|---|---|---|---|---|---|---|---|---|---|---|---|---|---|---|---|---|---|---|---|
| E | S | 0.1 | 1 | 10 | 100 | 1000 | 0.1 | 1 | 10 | 100 | 1000 | 0.1 | 1 | 10 | 100 | 1000 | 0.1 | 1 | 10 | 100 | 1000 | 0.1 | 1 | 10 | 100 | 1000 | 0.1 | 1 | 10 | 100 | 1000 |
| M5 | F1 | -20 | -18 | -14 | -22 | -24 | -8 | -19 | -5 | -10 | -19 | -25 | -8 | -3 | -3 | -7 | -19 | -5 | -3 | -14 | -16 | -44 | 0 | -5 | -3 | -5 | -23 | -21 | -11 | -4 | -8 |
| | SHD | -8 | -32 | -53 | -173 | -178 | -6 | -46 | -53 | -102 | -219 | -1 | 6 | -1 | 3 | -10 | -6 | -2 | -6 | -52 | -88 | -8 | 1 | -3 | -13 | -40 | 1 | -20 | -23 | -12 | -30 |
| | BSF | -27 | -22 | -14 | -15 | -14 | -14 | -26 | -8 | -10 | -23 | -32 | -17 | -2 | -5 | -1 | -25 | -7 | -3 | -6 | -3 | -56 | 2 | -5 | -5 | -11 | -31 | -29 | -14 | -2 | -2 |
| M10 | F1 | -30 | -18 | -17 | -26 | -26 | -29 | -35 | -6 | -13 | -21 | -22 | -3 | -3 | -3 | -18 | -19 | -7 | -11 | -17 | -16 | -36 | 1 | -2 | -5 | -4 | -32 | -23 | -13 | -6 | -10 |
| | SHD | -15 | -36 | -67 | -218 | -178 | -20 | -91 | -57 | -140 | -242 | 1 | 8 | -2 | 2 | -18 | -4 | -5 | -29 | -74 | -80 | -7 | 2 | -1 | -24 | -48 | 0 | -21 | -26 | -16 | -42 |
| | BSF | -39 | -24 | -15 | -17 | -18 | -42 | -47 | -9 | -14 | -26 | -29 | -12 | -2 | -4 | -14 | -25 | -8 | -9 | -8 | -5 | -44 | 3 | -2 | -9 | -13 | -42 | -31 | -16 | -4 | -2 |
| I5 | F1 | -26 | -24 | -27 | -37 | -35 | -23 | -26 | -14 | -28 | -34 | -37 | -28 | 3 | -2 | -17 | -19 | -9 | -14 | -15 | -18 | -16 | 6 | 1 | -7 | -11 | -30 | -17 | -13 | -14 | -25 |
| | SHD | -18 | -60 | -130 | -369 | -282 | -28 | -81 | -135 | -294 | -415 | -7 | -2 | -2 | 2 | -25 | -8 | -12 | -42 | -79 | -98 | -3 | 6 | -2 | -47 | -89 | -15 | -24 | -43 | -61 | -107 |
| | BSF | -33 | -27 | -20 | -23 | -22 | -40 | -35 | -18 | -39 | -50 | -43 | -35 | 4 | -1 | -9 | -25 | -10 | -10 | -4 | -5 | -21 | 9 | -2 | -17 | -25 | -33 | -19 | -7 | -1 | -5 |
| I10 | F1 | -49 | -34 | -34 | -40 | -41 | -35 | -36 | -29 | -30 | -38 | -9 | -24 | -13 | -25 | -35 | -26 | -19 | -16 | -21 | -23 | -32 | 2 | 1 | -7 | -15 | -43 | -34 | -21 | -17 | -27 |
| | SHD | -28 | -80 | -170 | -412 | -328 | -29 | -97 | -221 | -308 | -452 | -1 | 15 | 7 | 9 | -13 | -10 | -19 | -48 | -88 | -129 | -6 | 1 | -1 | -49 | -130 | -19 | -42 | -59 | -67 | -110 |
| | BSF | -58 | -39 | -29 | -28 | -29 | -50 | -50 | -42 | -42 | -63 | -9 | -32 | -14 | -27 | -30 | -34 | -20 | -13 | -12 | -10 | -38 | 2 | -1 | -17 | -37 | -48 | -38 | -17 | -5 | -7 |
| S5 | F1 | -3 | -2 | -3 | -4 | -6 | n/a | n/a | n/a | n/a | n/a | 2 | -10 | -2 | -2 | -1 | 0 | -1 | 0 | 1 | 1 | n/a | n/a | n/a | n/a | n/a | -2 | -2 | -1 | -2 | -1 |
| | SHD | -4 | -5 | -15 | -34 | -36 | n/a | n/a | n/a | n/a | n/a | 1 | 1 | -1 | -1 | -1 | -1 | 0 | 1 | 5 | 2 | n/a | n/a | n/a | n/a | n/a | -1 | -2 | -1 | -1 | -2 |
| | BSF | -4 | 0 | -1 | -1 | -3 | n/a | n/a | n/a | n/a | n/a | 1 | -14 | -3 | -3 | -1 | 1 | -1 | 0 | 2 | 2 | n/a | n/a | n/a | n/a | n/a | -2 | -3 | -1 | -4 | -1 |
| S10 | F1 | -4 | -6 | -4 | -6 | -5 | n/a | n/a | n/a | n/a | n/a | 1 | -16 | -7 | -4 | 0 | 5 | 3 | 1 | 0 | 1 | 6 | 0 | -1 | 3 | -6 | -7 | -2 | -1 | 0 | -1 |
| | SHD | -5 | -17 | -20 | -42 | -33 | n/a | n/a | n/a | n/a | n/a | 1 | -1 | -2 | -2 | -1 | 7 | 6 | 6 | 4 | 4 | 1 | 0 | -3 | 8 | -28 | -4 | -5 | -7 | -2 | -4 |
| | BSF | -5 | -7 | -3 | -5 | -4 | n/a | n/a | n/a | n/a | n/a | 1 | -21 | -9 | -6 | -1 | 6 | 3 | 1 | -1 | 1 | 6 | 0 | -3 | 4 | -9 | -6 | 0 | 1 | 2 | -1 |
| L5 | F1 | 0 | -4 | -4 | -4 | -2 | n/a | n/a | n/a | n/a | n/a | -15 | -24 | -12 | -9 | -16 | 2 | 4 | 2 | 5 | 6 | n/a | n/a | n/a | n/a | n/a | -8 | -6 | -4 | -2 | 0 |
| | SHD | 3 | -6 | -8 | -19 | -7 | n/a | n/a | n/a | n/a | n/a | -14 | -14 | -16 | -10 | -28 | -1 | 4 | 3 | 13 | 18 | n/a | n/a | n/a | n/a | n/a | -4 | -6 | -8 | -4 | 1 |
| | BSF | -2 | -7 | -5 | -5 | -2 | n/a | n/a | n/a | n/a | n/a | -22 | -34 | -20 | -18 | -22 | 0 | 3 | 1 | 5 | 7 | n/a | n/a | n/a | n/a | n/a | -11 | -8 | -6 | -4 | -2 |
| L10 | F1 | -14 | -13 | -13 | -9 | -5 | 17 | 3 | 1 | 0 | 0 | -29 | -28 | -23 | -23 | -28 | 13 | 12 | 8 | 11 | 13 | 5 | -1 | 0 | 1 | 0 | -10 | -6 | -7 | -5 | -7 |
| | SHD | -23 | -49 | -68 | -81 | -44 | 32 | 34 | 30 | 28 | 35 | -43 | -39 | -44 | -45 | -73 | 21 | 29 | 32 | 44 | 50 | 14 | 13 | 14 | 14 | 15 | -4 | -9 | -16 | -11 | -11 |
| | BSF | -23 | -23 | -23 | -16 | -13 | 21 | 6 | 2 | 1 | 2 | -41 | -43 | -37 | -37 | -39 | 15 | 15 | 12 | 16 | 18 | -1 | -3 | 0 | -1 | 0 | -15 | -11 | -12 | -10 | -13 |
| cMI | F1 | -49 | -31 | -24 | -39 | -38 | -45 | -25 | -27 | -20 | -26 | -38 | -42 | -12 | -18 | -23 | -31 | -21 | -12 | -17 | -20 | -56 | -7 | -4 | -8 | -15 | -42 | -36 | -22 | -16 | -19 |
| | SHD | -21 | -61 | -96 | -328 | -260 | -29 | -59 | -124 | -208 | -306 | 2 | 6 | 2 | -1 | -20 | -8 | -20 | -29 | -57 | -95 | -11 | -4 | -5 | -46 | -107 | -4 | -35 | -45 | -42 | -68 |
| | BSF | -59 | -40 | -23 | -30 | -29 | -61 | -35 | -33 | -28 | -39 | -46 | -52 | -16 | -23 | -21 | -39 | -24 | -12 | -12 | -9 | -72 | -8 | -6 | -17 | -32 | -51 | -44 | -26 | -15 | -8 |
| cMS | F1 | -27 | -22 | -15 | -24 | -24 | n/a | n/a | n/a | n/a | n/a | -24 | -12 | -8 | -6 | -7 | -9 | -4 | -4 | -13 | -15 | -19 | 1 | -3 | -3 | -6 | -30 | -22 | -12 | -6 | -8 |
| | SHD | -13 | -44 | -59 | -182 | -169 | n/a | n/a | n/a | n/a | n/a | -2 | 7 | 0 | 1 | -10 | -1 | -1 | -10 | -50 | -79 | -4 | 3 | -1 | -15 | -51 | -2 | -24 | -26 | -16 | -30 |
| | BSF | -35 | -28 | -15 | -17 | -15 | n/a | n/a | n/a | n/a | n/a | -30 | -22 | -10 | -8 | -1 | -14 | -5 | -4 | -4 | -2 | -26 | 4 | -3 | -6 | -14 | -38 | -29 | -15 | -5 | 0 |
| cML | F1 | -23 | -19 | -17 | -24 | -23 | -14 | -3 | -4 | -14 | -19 | -33 | -22 | -19 | -16 | -34 | -32 | -7 | -3 | -7 | -14 | -13 | -7 | -8 | -7 | -9 | -26 | -27 | -17 | -7 | -9 |
| | SHD | -6 | -32 | -57 | -169 | -153 | 15 | 13 | -8 | -77 | -127 | -16 | -9 | -17 | -15 | -47 | -14 | -6 | -10 | -34 | -77 | 8 | 6 | 6 | -11 | -33 | -1 | -27 | -33 | -17 | -21 |
| | BSF | -31 | -27 | -19 | -19 | -16 | -25 | -7 | -5 | -15 | -22 | -43 | -35 | -26 | -25 | -37 | -41 | -11 | -6 | -2 | -5 | -41 | -14 | -11 | -13 | -20 | -35 | -35 | -22 | -9 | -8 |
| cIS | F1 | -28 | -19 | -27 | -36 | -38 | n/a | n/a | n/a | n/a | n/a | -30 | -24 | 2 | -5 | -21 | -5 | -11 | -8 | -19 | -21 | -2 | -5 | -1 | -6 | -11 | -25 | -15 | -13 | -15 | -21 |
| | SHD | -18 | -48 | -132 | -364 | -301 | n/a | n/a | n/a | n/a | n/a | -5 | -1 | 0 | 0 | -27 | -1 | -13 | -28 | -98 | -112 | -2 | -8 | -6 | -38 | -93 | -11 | -21 | -44 | -64 | -90 |
| | BSF | -33 | -21 | -21 | -23 | -26 | n/a | n/a | n/a | n/a | n/a | -36 | -31 | 2 | -5 | -14 | -10 | -11 | -4 | -7 | -9 | -10 | -12 | -6 | -13 | -27 | -28 | -17 | -7 | -2 | -3 |
| cIL | F1 | -30 | -18 | -24 | -35 | -31 | -25 | -24 | -22 | -28 | -29 | -40 | -35 | -9 | -13 | -31 | -26 | -4 | -7 | -15 | -17 | -19 | -7 | 0 | -1 | -10 | -35 | -21 | -17 | -16 | -23 |
| | SHD | -15 | -41 | -111 | -313 | -247 | 14 | -37 | -101 | -172 | -244 | -21 | -16 | -18 | -13 | -51 | -14 | -7 | -26 | -83 | -95 | 10 | 9 | 14 | -4 | -45 | -16 | -27 | -51 | -64 | -92 |
| | BSF | -38 | -22 | -20 | -25 | -20 | -34 | -30 | -29 | -39 | -41 | -50 | -46 | -14 | -20 | -28 | -34 | -6 | -5 | -6 | -9 | -32 | -10 | -1 | -8 | -25 | -39 | -25 | -15 | -7 | -8 |
| cSL | F1 | 3 | -2 | -3 | -4 | -1 | n/a | n/a | n/a | n/a | n/a | -15 | -23 | -12 | -14 | -17 | 3 | 4 | 2 | 4 | 4 | 10 | 3 | -1 | 3 | -5 | -10 | -6 | -6 | -7 | -1 |
| | SHD | 5 | -2 | -6 | -22 | 1 | n/a | n/a | n/a | n/a | n/a | -15 | -14 | -16 | -17 | -30 | 0 | 4 | 2 | 9 | 13 | 15 | 16 | 13 | 24 | 3 | -5 | -9 | -15 | -17 | -11 |
| | BSF | 3 | -5 | -5 | -5 | -2 | n/a | n/a | n/a | n/a | n/a | -21 | -33 | -20 | -22 | -22 | 3 | 3 | 1 | 4 | 4 | 7 | 3 | -2 | 4 | -6 | -12 | -8 | -8 | -8 | 1 |
| cMISL | F1 | -63 | -39 | -26 | -35 | -35 | -48 | -40 | -32 | -29 | -33 | -50 | -51 | -24 | -23 | -34 | -36 | -18 | -11 | -16 | -21 | -44 | -11 | 1 | -11 | -14 | -42 | -43 | -28 | -20 | -19 |
| | SHD | -25 | -68 | -98 | -283 | -219 | -2 | -46 | -115 | -158 | -223 | -15 | -9 | -13 | -16 | -47 | -13 | -18 | -35 | -57 | -105 | 6 | 5 | 10 | -31 | -77 | -5 | -43 | -59 | -49 | -58 |
| | BSF | -72 | -48 | -27 | -30 | -29 | -66 | -53 | -42 | -39 | -49 | -60 | -62 | -35 | -33 | -37 | -46 | -24 | -14 | -14 | -15 | -60 | -17 | -5 | -23 | -38 | -52 | -52 | -34 | -22 | -13 |

**Table E5.** Precision (P), Recall (R), F1, SHD and BSF scores of the algorithms over all experiments cMISL, where F represents a failed attempt by the algorithm to produce a graph (refer to Appendix C). The results are presented per case study per sample size $n$.

| Algorithm | Data | $n$ | P | R | F1 | SHD | BSF | Algorithm | Data | $n$ | P | R | F1 | SHD | BSF |
|---|---|---|---|---|---|---|---|---|---|---|---|---|---|---|---|
| FCI | Alarm | 0.1k | 0.32 | 0.12 | 0.18 | 48.5 | 0.11 | Inter-IAMB | Alarm | 0.1k | 0.33 | 0.07 | 0.11 | 45 | 0.06 |
| FCI | Alarm | 1k | 0.50 | 0.42 | 0.46 | 37 | 0.40 | Inter-IAMB | Alarm | 1k | 0.55 | 0.26 | 0.35 | 36.5 | 0.25 |
| FCI | Alarm | 10k | 0.45 | 0.61 | 0.52 | 41.5 | 0.57 | Inter-IAMB | Alarm | 10k | 0.69 | 0.48 | 0.57 | 29.5 | 0.47 |
| FCI | Alarm | 100k | 0.35 | 0.66 | 0.45 | 62.5 | 0.57 | Inter-IAMB | Alarm | 100k | 0.54 | 0.54 | 0.54 | 35.5 | 0.52 |
| FCI | Alarm | 1000k | F | F | F | F | F | Inter-IAMB | Alarm | 1000k | 0.40 | 0.51 | 0.45 | 49 | 0.46 |
| FCI | Asia | 0.1k | 0.25 | 0.08 | 0.13 | 6.5 | 0.02 | Inter-IAMB | Asia | 0.1k | 0.33 | 0.17 | 0.22 | 6 | 0.10 |
| FCI | Asia | 1k | 0.38 | 0.25 | 0.30 | 5.5 | 0.18 | Inter-IAMB | Asia | 1k | 0.38 | 0.25 | 0.30 | 5.5 | 0.18 |
| FCI | Asia | 10k | 0.39 | 0.58 | 0.47 | 6.5 | 0.32 | Inter-IAMB | Asia | 10k | 0.44 | 0.58 | 0.50 | 5.5 | 0.38 |
| FCI | Asia | 100k | 0.39 | 0.58 | 0.47 | 6.5 | 0.32 | Inter-IAMB | Asia | 100k | 0.44 | 0.67 | 0.53 | 6 | 0.40 |
| FCI | Asia | 1000k | 0.31 | 0.67 | 0.42 | 9 | 0.20 | Inter-IAMB | Asia | 1000k | 0.33 | 0.67 | 0.44 | 9 | 0.20 |
| FCI | Formed | 0.1k | 0.41 | 0.13 | 0.19 | 138.5 | 0.12 | Inter-IAMB | Formed | 0.1k | 0.46 | 0.08 | 0.14 | 130.5 | 0.08 |
| FCI | Formed | 1k | 0.54 | 0.24 | 0.33 | 119.5 | 0.24 | Inter-IAMB | Formed | 1k | 0.62 | 0.21 | 0.31 | 115.5 | 0.21 |
| FCI | Formed | 10k | 0.47 | 0.40 | 0.43 | 121.5 | 0.39 | Inter-IAMB | Formed | 10k | 0.55 | 0.33 | 0.41 | 111 | 0.32 |
| FCI | Formed | 100k | F | F | F | F | F | Inter-IAMB | Formed | 100k | 0.52 | 0.42 | 0.47 | 115 | 0.41 |
| FCI | Formed | 1000k | F | F | F | F | F | Inter-IAMB | Formed | 1000k | 0.48 | 0.50 | 0.49 | 122.5 | 0.48 |
| FCI | Pathfinder | 0.1k | 0.09 | 0.02 | 0.04 | 274.5 | 0.01 | Inter-IAMB | Pathfinder | 0.1k | 0.25 | 0.02 | 0.03 | 233.5 | 0.01 |
| FCI | Pathfinder | 1k | F | F | F | F | F | Inter-IAMB | Pathfinder | 1k | 0.32 | 0.04 | 0.07 | 237.5 | 0.04 |
| FCI | Pathfinder | 10k | F | F | F | F | F | Inter-IAMB | Pathfinder | 10k | 0.53 | 0.05 | 0.08 | 221.5 | 0.05 |
| FCI | Pathfinder | 100k | F | F | F | F | F | Inter-IAMB | Pathfinder | 100k | 0.48 | 0.07 | 0.12 | 221 | 0.06 |
| FCI | Pathfinder | 1000k | F | F | F | F | F | Inter-IAMB | Pathfinder | 1000k | 0.43 | 0.09 | 0.14 | 221 | 0.09 |
| FCI | Property | 0.1k | 0.29 | 0.11 | 0.16 | 33.5 | 0.09 | Inter-IAMB | Property | 0.1k | 0.50 | 0.09 | 0.16 | 29 | 0.09 |
| FCI | Property | 1k | 0.65 | 0.47 | 0.55 | 20 | 0.46 | Inter-IAMB | Property | 1k | 0.63 | 0.23 | 0.34 | 24.5 | 0.23 |
| FCI | Property | 10k | 0.48 | 0.70 | 0.57 | 30.5 | 0.63 | Inter-IAMB | Property | 10k | 0.61 | 0.34 | 0.44 | 23 | 0.34 |
| FCI | Property | 100k | F | F | F | F | F | Inter-IAMB | Property | 100k | 0.61 | 0.42 | 0.50 | 21.5 | 0.41 |
| FCI | Property | 1000k | F | F | F | F | F | Inter-IAMB | Property | 1000k | 0.50 | 0.41 | 0.45 | 24 | 0.39 |
| FCI | Sports | 0.1k | 0.75 | 0.23 | 0.35 | 10 | 0.23 | Inter-IAMB | Sports | 0.1k | 0.50 | 0.12 | 0.19 | 11.5 | 0.12 |
| FCI | Sports | 1k | 0.71 | 0.39 | 0.50 | 8 | 0.39 | Inter-IAMB | Sports | 1k | 0.80 | 0.31 | 0.44 | 9 | 0.31 |
| FCI | Sports | 10k | 0.50 | 0.58 | 0.54 | 9.5 | 0.31 | Inter-IAMB | Sports | 10k | 0.70 | 0.54 | 0.61 | 7 | 0.47 |
| FCI | Sports | 100k | 0.42 | 0.62 | 0.50 | 11 | 0.22 | Inter-IAMB | Sports | 100k | 0.61 | 0.65 | 0.63 | 7.5 | 0.45 |
| FCI | Sports | 1000k | 0.32 | 0.62 | 0.42 | 17 | -0.19 | Inter-IAMB | Sports | 1000k | 0.50 | 0.65 | 0.57 | 10.5 | 0.25 |
| FGES | Alarm | 0.1k | 0.27 | 0.07 | 0.11 | 47 | 0.06 | MMHC | Alarm | 0.1k | 0.75 | 0.07 | 0.12 | 43 | 0.07 |
| FGES | Alarm | 1k | 0.42 | 0.23 | 0.30 | 39.5 | 0.22 | MMHC | Alarm | 1k | 0.78 | 0.28 | 0.41 | 33.5 | 0.28 |
| FGES | Alarm | 10k | 0.61 | 0.57 | 0.59 | 28.5 | 0.55 | MMHC | Alarm | 10k | 0.67 | 0.49 | 0.56 | 29 | 0.48 |
| FGES | Alarm | 100k | 0.59 | 0.66 | 0.62 | 31.5 | 0.63 | MMHC | Alarm | 100k | 0.59 | 0.50 | 0.54 | 34.5 | 0.48 |
| FGES | Alarm | 1000k | 0.56 | 0.67 | 0.61 | 33 | 0.63 | MMHC | Alarm | 1000k | 0.57 | 0.48 | 0.52 | 36.5 | 0.45 |
| FGES | Asia | 0.1k | 0.25 | 0.08 | 0.13 | 6.5 | 0.02 | MMHC | Asia | 0.1k | 1.00 | 0.33 | 0.50 | 4 | 0.33 |
| FGES | Asia | 1k | 0.38 | 0.25 | 0.30 | 5.5 | 0.18 | MMHC | Asia | 1k | 1.00 | 0.50 | 0.67 | 3 | 0.50 |
| FGES | Asia | 10k | 0.36 | 0.42 | 0.39 | 5.5 | 0.28 | MMHC | Asia | 10k | 0.57 | 0.67 | 0.62 | 4 | 0.53 |
| FGES | Asia | 100k | 0.50 | 0.67 | 0.57 | 5 | 0.47 | MMHC | Asia | 100k | 0.56 | 0.75 | 0.64 | 4.5 | 0.55 |
| FGES | Asia | 1000k | 0.44 | 0.67 | 0.53 | 6 | 0.40 | MMHC | Asia | 1000k | 0.56 | 0.83 | 0.67 | 5 | 0.57 |
| FGES | Formed | 0.1k | 0.46 | 0.08 | 0.13 | 131.5 | 0.07 | MMHC | Formed | 0.1k | 0.92 | 0.13 | 0.22 | 122.5 | 0.13 |
| FGES | Formed | 1k | 0.43 | 0.15 | 0.23 | 128.5 | 0.15 | MMHC | Formed | 1k | 0.81 | 0.24 | 0.37 | 110 | 0.24 |
| FGES | Formed | 10k | 0.59 | 0.41 | 0.48 | 102 | 0.40 | MMHC | Formed | 10k | 0.70 | 0.40 | 0.51 | 98.5 | 0.39 |
| FGES | Formed | 100k | 0.65 | 0.57 | 0.61 | 88 | 0.56 | MMHC | Formed | 100k | 0.63 | 0.46 | 0.53 | 105 | 0.45 |
| FGES | Formed | 1000k | F | F | F | F | F | MMHC | Formed | 1000k | 0.66 | 0.53 | 0.59 | 98 | 0.52 |
| FGES | Pathfinder | 0.1k | 0.29 | 0.05 | 0.08 | 239.5 | 0.04 | MMHC | Pathfinder | 0.1k | 0.67 | 0.02 | 0.03 | 228 | 0.02 |
| FGES | Pathfinder | 1k | 0.20 | 0.05 | 0.08 | 254 | 0.04 | MMHC | Pathfinder | 1k | 0.39 | 0.04 | 0.08 | 236 | 0.04 |
| FGES | Pathfinder | 10k | 0.28 | 0.13 | 0.17 | 269 | 0.11 | MMHC | Pathfinder | 10k | 0.88 | 0.06 | 0.11 | 218 | 0.06 |
| FGES | Pathfinder | 100k | 0.19 | 0.12 | 0.14 | 311 | 0.10 | MMHC | Pathfinder | 100k | 0.73 | 0.08 | 0.14 | 217.5 | 0.08 |
| FGES | Pathfinder | 1000k | F | F | F | F | F | MMHC | Pathfinder | 1000k | 0.73 | 0.08 | 0.15 | 218 | 0.08 |
| FGES | Property | 0.1k | 0.50 | 0.13 | 0.20 | 28 | 0.13 | MMHC | Property | 0.1k | 1.00 | 0.09 | 0.17 | 29 | 0.09 |
| FGES | Property | 1k | 0.44 | 0.23 | 0.31 | 26.5 | 0.23 | MMHC | Property | 1k | 0.83 | 0.31 | 0.46 | 22 | 0.31 |
| FGES | Property | 10k | 0.72 | 0.56 | 0.63 | 19 | 0.55 | MMHC | Property | 10k | 0.76 | 0.45 | 0.57 | 19.5 | 0.45 |
| FGES | Property | 100k | 0.65 | 0.63 | 0.64 | 20 | 0.60 | MMHC | Property | 100k | 0.67 | 0.55 | 0.60 | 19.5 | 0.53 |
| FGES | Property | 1000k | 0.61 | 0.69 | 0.65 | 22 | 0.65 | MMHC | Property | 1000k | 0.69 | 0.58 | 0.63 | 18.5 | 0.56 |

| | | | | | | | |
|---|---|---|---|---|---|---|---|
| FGES | Sports | 0.1k | 0.00 | 0.00 | 0.00 | 13 | 0.00 |
| FGES | Sports | 1k | 0.75 | 0.23 | 0.35 | 10 | 0.23 |
| FGES | Sports | 10k | 0.63 | 0.39 | 0.48 | 8 | 0.39 |
| FGES | Sports | 100k | 0.55 | 0.42 | 0.48 | 9.5 | 0.29 |
| FGES | Sports | 1000k | 0.50 | 0.54 | 0.52 | 11 | 0.21 |
| GFCI | Alarm | 0.1k | 0.27 | 0.07 | 0.11 | 47 | 0.06 |
| GFCI | Alarm | 1k | 0.42 | 0.23 | 0.30 | 39.5 | 0.22 |
| GFCI | Alarm | 10k | 0.57 | 0.53 | 0.55 | 30 | 0.52 |
| GFCI | Alarm | 100k | 0.57 | 0.63 | 0.60 | 32.5 | 0.60 |
| GFCI | Alarm | 1000k | 0.54 | 0.64 | 0.59 | 34 | 0.61 |
| GFCI | Asia | 0.1k | 0.25 | 0.08 | 0.13 | 6.5 | 0.02 |
| GFCI | Asia | 1k | 0.38 | 0.25 | 0.30 | 5.5 | 0.18 |
| GFCI | Asia | 10k | 0.36 | 0.42 | 0.39 | 5.5 | 0.28 |
| GFCI | Asia | 100k | 0.50 | 0.67 | 0.57 | 5 | 0.47 |
| GFCI | Asia | 1000k | 0.39 | 0.58 | 0.47 | 6.5 | 0.32 |
| GFCI | Formed | 0.1k | 0.46 | 0.08 | 0.13 | 131.5 | 0.07 |
| GFCI | Formed | 1k | 0.43 | 0.15 | 0.23 | 128.5 | 0.15 |
| GFCI | Formed | 10k | 0.59 | 0.41 | 0.48 | 102 | 0.40 |
| GFCI | Formed | 100k | 0.65 | 0.57 | 0.61 | 88.5 | 0.56 |
| GFCI | Formed | 1000k | F | F | F | F | F |
| GFCI | Pathfinder | 0.1k | 0.29 | 0.05 | 0.08 | 239.5 | 0.04 |
| GFCI | Pathfinder | 1k | 0.20 | 0.05 | 0.08 | 254 | 0.04 |
| GFCI | Pathfinder | 10k | 0.28 | 0.13 | 0.17 | 269 | 0.11 |
| GFCI | Pathfinder | 100k | 0.18 | 0.11 | 0.14 | 312 | 0.09 |
| GFCI | Pathfinder | 1000k | F | F | F | F | F |
| GFCI | Property | 0.1k | 0.50 | 0.13 | 0.20 | 28 | 0.13 |
| GFCI | Property | 1k | 0.44 | 0.23 | 0.31 | 26.5 | 0.23 |
| GFCI | Property | 10k | 0.72 | 0.52 | 0.60 | 18.5 | 0.51 |
| GFCI | Property | 100k | 0.63 | 0.61 | 0.62 | 20.5 | 0.58 |
| GFCI | Property | 1000k | 0.60 | 0.67 | 0.63 | 22.5 | 0.63 |
| GFCI | Sports | 0.1k | 0.00 | 0.00 | 0.00 | 13 | 0.00 |
| GFCI | Sports | 1k | 0.75 | 0.23 | 0.35 | 10 | 0.23 |
| GFCI | Sports | 10k | 0.63 | 0.39 | 0.48 | 8 | 0.39 |
| GFCI | Sports | 100k | 0.55 | 0.42 | 0.48 | 9.5 | 0.29 |
| GFCI | Sports | 1000k | 0.50 | 0.54 | 0.52 | 11 | 0.21 |
| GS | Alarm | 0.1k | 0.36 | 0.06 | 0.10 | 44.5 | 0.05 |
| GS | Alarm | 1k | 0.60 | 0.13 | 0.22 | 40 | 0.13 |
| GS | Alarm | 10k | 0.60 | 0.20 | 0.30 | 39 | 0.20 |
| GS | Alarm | 100k | 0.42 | 0.24 | 0.31 | 45 | 0.22 |
| GS | Alarm | 1000k | 0.41 | 0.30 | 0.35 | 45.5 | 0.28 |
| GS | Asia | 0.1k | 0.33 | 0.17 | 0.22 | 6 | 0.10 |
| GS | Asia | 1k | 0.38 | 0.25 | 0.30 | 5.5 | 0.18 |
| GS | Asia | 10k | 0.44 | 0.58 | 0.50 | 5.5 | 0.38 |
| GS | Asia | 100k | 0.44 | 0.67 | 0.53 | 6 | 0.40 |
| GS | Asia | 1000k | 0.32 | 0.58 | 0.41 | 8.5 | 0.18 |
| GS | Formed | 0.1k | 0.42 | 0.07 | 0.12 | 134 | 0.07 |
| GS | Formed | 1k | 0.53 | 0.11 | 0.18 | 126.5 | 0.11 |
| GS | Formed | 10k | 0.63 | 0.21 | 0.32 | 112.5 | 0.21 |
| GS | Formed | 100k | 0.59 | 0.25 | 0.35 | 112 | 0.25 |
| GS | Formed | 1000k | 0.59 | 0.28 | 0.38 | 109.5 | 0.28 |
| GS | Pathfinder | 0.1k | 0.20 | 0.01 | 0.02 | 234 | 0.01 |
| GS | Pathfinder | 1k | 0.50 | 0.01 | 0.01 | 228.5 | 0.01 |
| GS | Pathfinder | 10k | 0.54 | 0.03 | 0.06 | 224 | 0.03 |
| GS | Pathfinder | 100k | 0.66 | 0.05 | 0.09 | 220.5 | 0.05 |
| GS | Pathfinder | 1000k | 0.46 | 0.05 | 0.09 | 222 | 0.05 |
| GS | Property | 0.1k | 0.38 | 0.05 | 0.08 | 31.5 | 0.04 |
| GS | Property | 1k | 0.79 | 0.17 | 0.28 | 26.5 | 0.17 |
| GS | Property | 10k | 0.60 | 0.19 | 0.29 | 26 | 0.19 |
| GS | Property | 100k | 0.63 | 0.23 | 0.34 | 26.5 | 0.23 |
| GS | Property | 1000k | 0.47 | 0.28 | 0.35 | 28 | 0.26 |
| MMHC | Sports | 0.1k | 0.00 | 0.00 | 0.00 | 13 | 0.00 |
| MMHC | Sports | 1k | 1.00 | 0.46 | 0.63 | 7 | 0.46 |
| MMHC | Sports | 10k | 1.00 | 0.62 | 0.76 | 5 | 0.62 |
| MMHC | Sports | 100k | 0.95 | 0.73 | 0.83 | 3.5 | 0.73 |
| MMHC | Sports | 1000k | 1.00 | 0.92 | 0.96 | 1 | 0.92 |
| NOTEARS | Sports | 0.1k | 0.35 | 0.27 | 0.30 | 13.5 | 0.00 |
| NOTEARS | Sports | 1k | 0.36 | 0.31 | 0.33 | 12 | 0.11 |
| NOTEARS | Sports | 10k | 0.31 | 0.31 | 0.31 | 14 | -0.03 |
| NOTEARS | Sports | 100k | 0.31 | 0.31 | 0.31 | 14 | -0.03 |
| NOTEARS | Sports | 1000k | 0.31 | 0.31 | 0.31 | 14 | -0.03 |
| PC-Stable | Alarm | 0.1k | 0.28 | 0.10 | 0.15 | 48.5 | 0.09 |
| PC-Stable | Alarm | 1k | 0.61 | 0.54 | 0.58 | 31.5 | 0.52 |
| PC-Stable | Alarm | 10k | 0.50 | 0.68 | 0.58 | 38.5 | 0.63 |
| PC-Stable | Alarm | 100k | 0.38 | 0.72 | 0.50 | 59.5 | 0.64 |
| PC-Stable | Alarm | 1000k | F | F | F | F | F |
| PC-Stable | Asia | 0.1k | 0.25 | 0.08 | 0.13 | 6.5 | 0.02 |
| PC-Stable | Asia | 1k | 0.38 | 0.25 | 0.30 | 5.5 | 0.18 |
| PC-Stable | Asia | 10k | 0.39 | 0.58 | 0.47 | 6.5 | 0.32 |
| PC-Stable | Asia | 100k | 0.39 | 0.58 | 0.47 | 6.5 | 0.32 |
| PC-Stable | Asia | 1000k | 0.31 | 0.67 | 0.42 | 9 | 0.20 |
| PC-Stable | Formed | 0.1k | 0.41 | 0.13 | 0.20 | 139.5 | 0.13 |
| PC-Stable | Formed | 1k | 0.58 | 0.25 | 0.35 | 118.5 | 0.25 |
| PC-Stable | Formed | 10k | 0.58 | 0.49 | 0.53 | 109.5 | 0.48 |
| PC-Stable | Formed | 100k | F | F | F | F | F |
| PC-Stable | Formed | 1000k | F | F | F | F | F |
| PC-Stable | Pathfinder | 0.1k | 0.12 | 0.03 | 0.05 | 271.5 | 0.02 |
| PC-Stable | Pathfinder | 1k | F | F | F | F | F |
| PC-Stable | Pathfinder | 10k | F | F | F | F | F |
| PC-Stable | Pathfinder | 100k | F | F | F | F | F |
| PC-Stable | Pathfinder | 1000k | F | F | F | F | F |
| PC-Stable | Property | 0.1k | 0.36 | 0.13 | 0.19 | 31 | 0.12 |
| PC-Stable | Property | 1k | 0.67 | 0.50 | 0.57 | 20 | 0.49 |
| PC-Stable | Property | 10k | 0.49 | 0.70 | 0.58 | 29.5 | 0.64 |
| PC-Stable | Property | 100k | F | F | F | F | F |
| PC-Stable | Property | 1000k | F | F | F | F | F |
| PC-Stable | Sports | 0.1k | 0.75 | 0.23 | 0.35 | 10 | 0.23 |
| PC-Stable | Sports | 1k | 0.71 | 0.39 | 0.50 | 8 | 0.39 |
| PC-Stable | Sports | 10k | 0.53 | 0.62 | 0.57 | 9 | 0.35 |
| PC-Stable | Sports | 100k | 0.48 | 0.73 | 0.58 | 10.5 | 0.26 |
| PC-Stable | Sports | 1000k | 0.34 | 0.65 | 0.45 | 16.5 | -0.15 |
| RFCI-BSC | Alarm | 0.1k | 0.35 | 0.10 | 0.16 | 45.7 | 0.09 |
| RFCI-BSC | Alarm | 1k | 0.53 | 0.23 | 0.32 | 38.2 | 0.23 |
| RFCI-BSC | Alarm | 10k | 0.51 | 0.50 | 0.50 | 35.1 | 0.48 |
| RFCI-BSC | Alarm | 100k | F | F | F | F | F |
| RFCI-BSC | Alarm | 1000k | F | F | F | F | F |
| RFCI-BSC | Asia | 0.1k | 0.33 | 0.17 | 0.22 | 6 | 0.10 |
| RFCI-BSC | Asia | 1k | 0.38 | 0.25 | 0.30 | 5.5 | 0.18 |
| RFCI-BSC | Asia | 10k | 0.51 | 0.52 | 0.52 | 4.8 | 0.40 |
| RFCI-BSC | Asia | 100k | F | F | F | F | F |
| RFCI-BSC | Asia | 1000k | F | F | F | F | F |
| RFCI-BSC | Formed | 0.1k | 0.42 | 0.08 | 0.13 | 133.8 | 0.08 |
| RFCI-BSC | Formed | 1k | 0.59 | 0.18 | 0.28 | 122 | 0.18 |
| RFCI-BSC | Formed | 10k | F | F | F | F | F |
| RFCI-BSC | Formed | 100k | F | F | F | F | F |
| RFCI-BSC | Formed | 1000k | F | F | F | F | F |
| RFCI-BSC | Pathfinder | 0.1k | 0.23 | 0.03 | 0.06 | 245.3 | 0.03 |
| RFCI-BSC | Pathfinder | 1k | 0.30 | 0.06 | 0.10 | 242.6 | 0.06 |
| RFCI-BSC | Pathfinder | 10k | F | F | F | F | F |
| RFCI-BSC | Pathfinder | 100k | F | F | F | F | F |
| RFCI-BSC | Pathfinder | 1000k | F | F | F | F | F |

| | | | | | | | | | | | | | | | |
|---|---|---|---|---|---|---|---|---|---|---|---|---|---|---|---|
| GS | Sports | 0.1k | 0.50 | 0.04 | 0.07 | 12.5 | 0.04 | RFCI-BSC | Property | 0.1k | 0.52 | 0.14 | 0.22 | 28.4 | 0.13 |
| GS | Sports | 1k | 0.50 | 0.15 | 0.24 | 11 | 0.15 | RFCI-BSC | Property | 1k | 0.55 | 0.16 | 0.25 | 27.3 | 0.16 |
| GS | Sports | 10k | 0.81 | 0.50 | 0.62 | 6.5 | 0.50 | RFCI-BSC | Property | 10k | 0.55 | 0.34 | 0.42 | 24.8 | 0.33 |
| GS | Sports | 100k | 0.44 | 0.31 | 0.36 | 10 | 0.24 | RFCI-BSC | Property | 100k | F | F | F | F | F |
| GS | Sports | 1000k | 0.54 | 0.58 | 0.56 | 9.5 | 0.31 | RFCI-BSC | Property | 1000k | F | F | F | F | F |
| H2PC | Alarm | 0.1k | 0.67 | 0.09 | 0.16 | 43 | 0.09 | RFCI-BSC | Sports | 0.1k | F | F | F | F | F |
| H2PC | Alarm | 1k | 0.74 | 0.28 | 0.40 | 34.5 | 0.27 | RFCI-BSC | Sports | 1k | 0.77 | 0.22 | 0.34 | 10.2 | 0.22 |
| H2PC | Alarm | 10k | F | F | F | F | F | RFCI-BSC | Sports | 10k | 0.72 | 0.26 | 0.38 | 9.7 | 0.26 |
| H2PC | Alarm | 100k | 0.60 | 0.71 | 0.65 | 30 | 0.68 | RFCI-BSC | Sports | 100k | F | F | F | F | F |
| H2PC | Alarm | 1000k | 0.51 | 0.83 | 0.63 | 40.5 | 0.77 | RFCI-BSC | Sports | 1000k | F | F | F | F | F |
| H2PC | Asia | 0.1k | 0.25 | 0.08 | 0.13 | 6.5 | 0.02 | SaiyanH | Alarm | 0.1k | 0.29 | 0.22 | 0.25 | 56 | 0.18 |
| H2PC | Asia | 1k | 0.75 | 0.50 | 0.60 | 4 | 0.43 | SaiyanH | Alarm | 1k | 0.47 | 0.36 | 0.41 | 39 | 0.34 |
| H2PC | Asia | 10k | 0.57 | 0.67 | 0.62 | 4 | 0.53 | SaiyanH | Alarm | 10k | 0.62 | 0.57 | 0.59 | 31.5 | 0.55 |
| H2PC | Asia | 100k | 0.56 | 0.75 | 0.64 | 4.5 | 0.55 | SaiyanH | Alarm | 100k | 0.58 | 0.62 | 0.60 | 33 | 0.59 |
| H2PC | Asia | 1000k | 0.56 | 0.83 | 0.67 | 5 | 0.57 | SaiyanH | Alarm | 1000k | 0.47 | 0.67 | 0.55 | 43 | 0.62 |
| H2PC | Formed | 0.1k | 0.90 | 0.12 | 0.21 | 124 | 0.12 | SaiyanH | Asia | 0.1k | 0.33 | 0.33 | 0.33 | 7 | 0.13 |
| H2PC | Formed | 1k | F | F | F | F | F | SaiyanH | Asia | 1k | 0.42 | 0.42 | 0.42 | 5.5 | 0.28 |
| H2PC | Formed | 10k | 0.66 | 0.42 | 0.52 | 100 | 0.42 | SaiyanH | Asia | 10k | 0.36 | 0.42 | 0.39 | 6.5 | 0.22 |
| H2PC | Formed | 100k | 0.62 | 0.57 | 0.59 | 96 | 0.56 | SaiyanH | Asia | 100k | 0.57 | 0.67 | 0.62 | 4 | 0.53 |
| H2PC | Formed | 1000k | 0.59 | 0.72 | 0.65 | 101.5 | 0.70 | SaiyanH | Asia | 1000k | 0.57 | 0.67 | 0.62 | 4 | 0.53 |
| H2PC | Pathfinder | 0.1k | 0.63 | 0.02 | 0.04 | 228 | 0.02 | SaiyanH | Formed | 0.1k | 0.29 | 0.17 | 0.22 | 172 | 0.16 |
| H2PC | Pathfinder | 1k | 0.53 | 0.05 | 0.08 | 228.5 | 0.04 | SaiyanH | Formed | 1k | 0.49 | 0.30 | 0.38 | 125.5 | 0.30 |
| H2PC | Pathfinder | 10k | F | F | F | F | F | SaiyanH | Formed | 10k | 0.60 | 0.41 | 0.49 | 103 | 0.41 |
| H2PC | Pathfinder | 100k | 0.71 | 0.36 | 0.48 | 177.5 | 0.36 | SaiyanH | Formed | 100k | 0.63 | 0.44 | 0.52 | 100 | 0.44 |
| H2PC | Pathfinder | 1000k | F | F | F | F | F | SaiyanH | Formed | 1000k | F | F | F | F | F |
| H2PC | Property | 0.1k | 1.00 | 0.09 | 0.17 | 29 | 0.09 | SaiyanH | Pathfinder | 0.1k | 0.11 | 0.05 | 0.07 | 308 | 0.03 |
| H2PC | Property | 1k | 0.81 | 0.33 | 0.47 | 21.5 | 0.33 | SaiyanH | Pathfinder | 1k | 0.34 | 0.16 | 0.22 | 242.5 | 0.15 |
| H2PC | Property | 10k | 0.79 | 0.52 | 0.62 | 17.5 | 0.51 | SaiyanH | Pathfinder | 10k | 0.32 | 0.15 | 0.20 | 248 | 0.14 |
| H2PC | Property | 100k | 0.69 | 0.73 | 0.71 | 14.5 | 0.71 | SaiyanH | Pathfinder | 100k | 0.33 | 0.16 | 0.21 | 246 | 0.15 |
| H2PC | Property | 1000k | 0.57 | 0.81 | 0.67 | 21 | 0.76 | SaiyanH | Pathfinder | 1000k | F | F | F | F | F |
| H2PC | Sports | 0.1k | 0.00 | 0.00 | 0.00 | 13 | 0.00 | SaiyanH | Property | 0.1k | 0.28 | 0.22 | 0.25 | 41 | 0.16 |
| H2PC | Sports | 1k | 1.00 | 0.39 | 0.56 | 8 | 0.39 | SaiyanH | Property | 1k | 0.60 | 0.47 | 0.53 | 24 | 0.45 |
| H2PC | Sports | 10k | 1.00 | 0.62 | 0.76 | 5 | 0.62 | SaiyanH | Property | 10k | 0.83 | 0.67 | 0.74 | 13.5 | 0.66 |
| H2PC | Sports | 100k | 0.96 | 0.89 | 0.92 | 1.5 | 0.89 | SaiyanH | Property | 100k | 0.96 | 0.78 | 0.86 | 7 | 0.78 |
| H2PC | Sports | 1000k | 0.93 | 1.00 | 0.96 | 1 | 0.93 | SaiyanH | Property | 1000k | 0.98 | 0.83 | 0.90 | 5.5 | 0.83 |
| HC | Alarm | 0.1k | 0.50 | 0.09 | 0.15 | 45 | 0.08 | SaiyanH | Sports | 0.1k | 0.36 | 0.19 | 0.25 | 14.5 | -0.07 |
| HC | Alarm | 1k | 0.68 | 0.38 | 0.49 | 32 | 0.37 | SaiyanH | Sports | 1k | 1.00 | 0.54 | 0.70 | 6 | 0.54 |
| HC | Alarm | 10k | 0.66 | 0.64 | 0.65 | 24 | 0.63 | SaiyanH | Sports | 10k | 1.00 | 0.62 | 0.76 | 5 | 0.62 |
| HC | Alarm | 100k | 0.57 | 0.77 | 0.65 | 33.5 | 0.73 | SaiyanH | Sports | 100k | 0.90 | 0.69 | 0.78 | 4 | 0.69 |
| HC | Alarm | 1000k | 0.43 | 0.80 | 0.56 | 51 | 0.72 | SaiyanH | Sports | 1000k | 0.70 | 0.54 | 0.61 | 7 | 0.47 |
| HC | Asia | 0.1k | 1.00 | 0.33 | 0.50 | 4 | 0.33 | TABU | Alarm | 0.1k | 0.50 | 0.09 | 0.15 | 45 | 0.08 |
| HC | Asia | 1k | 1.00 | 0.50 | 0.67 | 3 | 0.50 | TABU | Alarm | 1k | 0.68 | 0.38 | 0.49 | 32 | 0.37 |
| HC | Asia | 10k | 0.57 | 0.67 | 0.62 | 4 | 0.53 | TABU | Alarm | 10k | 0.63 | 0.63 | 0.63 | 25.5 | 0.62 |
| HC | Asia | 100k | 0.50 | 0.75 | 0.60 | 5.5 | 0.48 | TABU | Alarm | 100k | 0.54 | 0.74 | 0.63 | 35.5 | 0.70 |
| HC | Asia | 1000k | 0.56 | 0.83 | 0.67 | 5 | 0.57 | TABU | Alarm | 1000k | 0.43 | 0.80 | 0.56 | 51 | 0.72 |
| HC | Formed | 0.1k | 0.84 | 0.15 | 0.26 | 121 | 0.15 | TABU | Asia | 0.1k | 1.00 | 0.33 | 0.50 | 4 | 0.33 |
| HC | Formed | 1k | 0.64 | 0.24 | 0.35 | 120 | 0.24 | TABU | Asia | 1k | 1.00 | 0.50 | 0.67 | 3 | 0.50 |
| HC | Formed | 10k | 0.62 | 0.46 | 0.53 | 102.5 | 0.45 | TABU | Asia | 10k | 0.57 | 0.67 | 0.62 | 4 | 0.53 |
| HC | Formed | 100k | 0.56 | 0.64 | 0.60 | 107 | 0.62 | TABU | Asia | 100k | 0.50 | 0.75 | 0.60 | 5.5 | 0.48 |
| HC | Formed | 1000k | 0.49 | 0.75 | 0.59 | 134.5 | 0.72 | TABU | Asia | 1000k | 0.50 | 0.75 | 0.60 | 5.5 | 0.48 |
| HC | Pathfinder | 0.1k | 0.65 | 0.09 | 0.15 | 220.5 | 0.08 | TABU | Formed | 0.1k | 0.84 | 0.15 | 0.26 | 121 | 0.15 |
| HC | Pathfinder | 1k | 0.27 | 0.06 | 0.10 | 252.5 | 0.06 | TABU | Formed | 1k | 0.66 | 0.26 | 0.37 | 118.5 | 0.26 |
| HC | Pathfinder | 10k | 0.41 | 0.20 | 0.27 | 248 | 0.19 | TABU | Formed | 10k | 0.62 | 0.46 | 0.53 | 102.5 | 0.45 |
| HC | Pathfinder | 100k | 0.65 | 0.39 | 0.49 | 184.5 | 0.38 | TABU | Formed | 100k | 0.58 | 0.65 | 0.61 | 104.5 | 0.64 |
| HC | Pathfinder | 1000k | 0.66 | 0.51 | 0.58 | 167 | 0.50 | TABU | Formed | 1000k | 0.49 | 0.76 | 0.60 | 132 | 0.73 |
| HC | Property | 0.1k | 0.86 | 0.19 | 0.31 | 26 | 0.19 | TABU | Pathfinder | 0.1k | 0.65 | 0.09 | 0.15 | 220.5 | 0.08 |
| HC | Property | 1k | 0.79 | 0.47 | 0.59 | 18 | 0.47 | TABU | Pathfinder | 1k | 0.27 | 0.06 | 0.10 | 253.5 | 0.06 |
| HC | Property | 10k | 0.65 | 0.61 | 0.63 | 18.5 | 0.59 | TABU | Pathfinder | 10k | 0.41 | 0.20 | 0.27 | 248 | 0.19 |
| HC | Property | 100k | 0.47 | 0.67 | 0.55 | 28.5 | 0.61 | TABU | Pathfinder | 100k | 0.65 | 0.39 | 0.49 | 184.5 | 0.38 |
| HC | Property | 1000k | 0.35 | 0.70 | 0.47 | 44.5 | 0.58 | TABU | Pathfinder | 1000k | 0.66 | 0.52 | 0.58 | 164.5 | 0.51 |

| | | | | | | | |
|---|---|---|---|---|---|---|---|
| HC | Sports | 0.1k | 0.00 | 0.00 | 0.00 | 13 | 0.00 |
| HC | Sports | 1k | 1.00 | 0.46 | 0.63 | 7 | 0.46 |
| HC | Sports | 10k | 1.00 | 0.62 | 0.76 | 5 | 0.62 |
| HC | Sports | 100k | 0.96 | 0.96 | 0.96 | 0.5 | 0.96 |
| HC | Sports | 1000k | 0.93 | 1.00 | 0.96 | 1 | 0.93 |
| ILP | Alarm | 0.1k | 0.40 | 0.18 | 0.25 | 47 | 0.16 |
| ILP | Alarm | 1k | 0.67 | 0.39 | 0.49 | 30.5 | 0.38 |
| ILP | Alarm | 10k | 0.72 | 0.72 | 0.72 | 20.5 | 0.71 |
| ILP | Alarm | 100k | 0.43 | 0.64 | 0.52 | 45 | 0.59 |
| ILP | Alarm | 1000k | 0.34 | 0.63 | 0.45 | 59.5 | 0.56 |
| ILP | Asia | 0.1k | 0.25 | 0.08 | 0.13 | 6.5 | 0.02 |
| ILP | Asia | 1k | 0.50 | 0.33 | 0.40 | 5 | 0.27 |
| ILP | Asia | 10k | 0.64 | 0.75 | 0.69 | 3.5 | 0.62 |
| ILP | Asia | 100k | 0.50 | 0.67 | 0.57 | 5 | 0.47 |
| ILP | Asia | 1000k | 0.50 | 0.75 | 0.60 | 5.5 | 0.48 |
| ILP | Formed | 0.1k | 0.34 | 0.14 | 0.20 | 150.5 | 0.13 |
| ILP | Formed | 1k | 0.60 | 0.25 | 0.35 | 117.5 | 0.24 |
| ILP | Formed | 10k | 0.55 | 0.41 | 0.47 | 109 | 0.41 |
| ILP | Formed | 100k | 0.46 | 0.54 | 0.49 | 132.5 | 0.52 |
| ILP | Formed | 1000k | F | F | F | F | F |
| ILP | Pathfinder | 0.1k | 0.16 | 0.08 | 0.11 | 305.5 | 0.06 |
| ILP | Pathfinder | 1k | 0.26 | 0.07 | 0.12 | 256 | 0.07 |
| ILP | Pathfinder | 10k | 0.36 | 0.17 | 0.23 | 256.5 | 0.16 |
| ILP | Pathfinder | 100k | 0.36 | 0.18 | 0.24 | 259.5 | 0.17 |
| ILP | Pathfinder | 1000k | F | F | F | F | F |
| ILP | Property | 0.1k | 0.81 | 0.20 | 0.33 | 25.5 | 0.20 |
| ILP | Property | 1k | 0.74 | 0.39 | 0.51 | 21.5 | 0.38 |
| ILP | Property | 10k | 0.85 | 0.69 | 0.76 | 12 | 0.68 |
| ILP | Property | 100k | 0.53 | 0.73 | 0.62 | 24.5 | 0.68 |
| ILP | Property | 1000k | 0.44 | 0.78 | 0.56 | 35 | 0.69 |
| ILP | Sports | 0.1k | 1.00 | 0.15 | 0.27 | 11 | 0.15 |
| ILP | Sports | 1k | 1.00 | 0.39 | 0.56 | 8 | 0.39 |
| ILP | Sports | 10k | 1.00 | 0.62 | 0.76 | 5 | 0.62 |
| ILP | Sports | 100k | 0.88 | 0.81 | 0.84 | 2.5 | 0.81 |
| ILP | Sports | 1000k | 0.82 | 0.89 | 0.85 | 2.5 | 0.82 |
| TABU | Property | 0.1k | 0.86 | 0.19 | 0.31 | 26 | 0.19 |
| TABU | Property | 1k | 0.79 | 0.47 | 0.59 | 18 | 0.47 |
| TABU | Property | 10k | 0.73 | 0.64 | 0.68 | 15.5 | 0.63 |
| TABU | Property | 100k | 0.47 | 0.70 | 0.56 | 28.5 | 0.64 |
| TABU | Property | 1000k | 0.37 | 0.73 | 0.49 | 43.5 | 0.62 |
| TABU | Sports | 0.1k | 0.00 | 0.00 | 0.00 | 13 | 0.00 |
| TABU | Sports | 1k | 0.75 | 0.35 | 0.47 | 8.5 | 0.35 |
| TABU | Sports | 10k | 0.88 | 0.54 | 0.67 | 6 | 0.54 |
| TABU | Sports | 100k | 0.85 | 0.85 | 0.85 | 2 | 0.85 |
| TABU | Sports | 1000k | 0.79 | 0.85 | 0.82 | 3 | 0.78 |
| WINASOBS | Alarm | 0.1k | 0.50 | 0.04 | 0.08 | 44 | 0.04 |
| WINASOBS | Alarm | 1k | 0.65 | 0.34 | 0.45 | 33.5 | 0.34 |
| WINASOBS | Alarm | 10k | 0.74 | 0.59 | 0.65 | 23.5 | 0.58 |
| WINASOBS | Alarm | 100k | 0.48 | 0.59 | 0.53 | 37.5 | 0.55 |
| WINASOBS | Alarm | 1000k | 0.69 | 0.48 | 0.57 | 27.5 | 0.47 |
| WINASOBS | Asia | 0.1k | 0.00 | 0.00 | 0.00 | 6 | 0.00 |
| WINASOBS | Asia | 1k | 0.50 | 0.17 | 0.25 | 5 | 0.17 |
| WINASOBS | Asia | 10k | 0.58 | 0.58 | 0.58 | 4.5 | 0.45 |
| WINASOBS | Asia | 100k | 0.50 | 0.58 | 0.54 | 4.5 | 0.45 |
| WINASOBS | Asia | 1000k | 0.44 | 0.67 | 0.53 | 6 | 0.40 |
| WINASOBS | Formed | 0.1k | 0.72 | 0.05 | 0.09 | 133.5 | 0.05 |
| WINASOBS | Formed | 1k | 0.56 | 0.18 | 0.27 | 122 | 0.18 |
| WINASOBS | Formed | 10k | 0.57 | 0.35 | 0.44 | 105.5 | 0.35 |
| WINASOBS | Formed | 100k | 0.54 | 0.44 | 0.49 | 109 | 0.43 |
| WINASOBS | Formed | 1000k | 0.54 | 0.32 | 0.40 | 106.5 | 0.32 |
| WINASOBS | Pathfinder | 0.1k | 0.36 | 0.03 | 0.06 | 234.5 | 0.03 |
| WINASOBS | Pathfinder | 1k | 0.25 | 0.05 | 0.08 | 250.5 | 0.04 |
| WINASOBS | Pathfinder | 10k | 0.38 | 0.15 | 0.22 | 249 | 0.14 |
| WINASOBS | Pathfinder | 100k | 0.75 | 0.29 | 0.42 | 182.5 | 0.29 |
| WINASOBS | Pathfinder | 1000k | F | F | F | F | F |
| WINASOBS | Property | 0.1k | 0.50 | 0.03 | 0.06 | 31 | 0.03 |
| WINASOBS | Property | 1k | 0.62 | 0.33 | 0.43 | 23.5 | 0.32 |
| WINASOBS | Property | 10k | 0.91 | 0.63 | 0.74 | 12 | 0.63 |
| WINASOBS | Property | 100k | 0.71 | 0.75 | 0.73 | 15 | 0.73 |
| WINASOBS | Property | 1000k | 0.74 | 0.58 | 0.65 | 13.5 | 0.58 |
| WINASOBS | Sports | 0.1k | 0.00 | 0.00 | 0.00 | 13 | 0.00 |
| WINASOBS | Sports | 1k | 0.70 | 0.27 | 0.39 | 9.5 | 0.27 |
| WINASOBS | Sports | 10k | 0.88 | 0.54 | 0.67 | 6 | 0.54 |
| WINASOBS | Sports | 100k | 0.83 | 0.77 | 0.80 | 3 | 0.77 |
| WINASOBS | Sports | 1000k | 0.86 | 0.92 | 0.89 | 2 | 0.86 |

**Table E6.** The averaged minimum and maximum scores for each case study and sample size $n$ combination, as determined by each of the three metrics over all 15 algorithms, and averaged across the 15 noisy experiments. Note that a higher SHD score indicates lower performance. The minimums and maximums are based on all the experiments shown in Table 11 that do not include a failure F.

| Min/Max performance | $n$ | F1 Asia | F1 Spor | F1 Prop | F1 Alar | F1 Form | F1 Path | SHD Asia | SHD Spor | SHD Prop | SHD Alar | SHD Form | SHD Path | BSF Asia | BSF Spor | BSF Prop | BSF Alar | BSF Form | BSF Path |
|---|---|---|---|---|---|---|---|---|---|---|---|---|---|---|---|---|---|---|---|
| Min | 0.1k | 0.22 | 0.00 | 0.12 | 0.13 | 0.12 | 0.03 | 7.3 | 15.7 | 35.8 | 50.1 | 193.6 | 345.8 | 0.11 | -0.03 | 0.07 | 0.08 | 0.07 | 0.01 |
| | 1k | 0.41 | 0.20 | 0.25 | 0.26 | 0.24 | 0.03 | 5.8 | 13.1 | 26.2 | 40.7 | 120.7 | 271.6 | 0.29 | 0.10 | 0.16 | 0.17 | 0.15 | 0.01 |
| | 10k | 0.53 | 0.34 | 0.31 | 0.34 | 0.36 | 0.06 | 5.8 | 13.2 | 25.2 | 39.9 | 107.7 | 283.4 | 0.44 | 0.13 | 0.20 | 0.24 | 0.25 | 0.03 |
| | 100k | 0.50 | 0.34 | 0.40 | 0.41 | 0.35 | 0.10 | 7.3 | 13.2 | 23.6 | 43.3 | 110.2 | 311.9 | 0.36 | 0.13 | 0.29 | 0.32 | 0.25 | 0.06 |
| | 1000k | 0.47 | 0.34 | 0.47 | 0.46 | 0.38 | 0.12 | 9.8 | 14.2 | 28.9 | 46.2 | 115.3 | 202.4 | 0.26 | 0.12 | 0.37 | 0.38 | 0.28 | 0.07 |
| Max | 0.1k | 0.53 | 0.32 | 0.54 | 0.42 | 0.32 | 0.18 | 4.8 | 11.3 | 19.4 | 37.9 | 120.6 | 211.5 | 0.37 | 0.21 | 0.42 | 0.40 | 0.23 | 0.11 |
| | 1k | 0.69 | 0.72 | 0.72 | 0.72 | 0.56 | 0.22 | 3.5 | 6.3 | 12.4 | 20.3 | 89.8 | 207.0 | 0.55 | 0.56 | 0.62 | 0.67 | 0.47 | 0.16 |
| | 10k | 0.77 | 0.75 | 0.84 | 0.77 | 0.68 | 0.33 | 3.2 | 5.6 | 8.7 | 18.5 | 75.1 | 199.8 | 0.70 | 0.61 | 0.81 | 0.78 | 0.66 | 0.26 |
| | 100k | 0.78 | 0.99 | 0.81 | 0.75 | 0.70 | 0.52 | 3.4 | 0.2 | 9.4 | 24.8 | 73.9 | 175.4 | 0.72 | 0.99 | 0.78 | 0.81 | 0.77 | 0.45 |
| | 1000k | 0.80 | 0.97 | 0.82 | 0.70 | 0.66 | 0.61 | 2.5 | 0.9 | 9.1 | 27.3 | 103.6 | 153.5 | 0.78 | 0.95 | 0.80 | 0.81 | 0.81 | 0.57 |

**Table E7.** Relative overall performance of the algorithms for each case study and sample size $n$ combination, as determined by each of the three metrics, and over the 15 noisy experiments. The performance is measured relative to the min/max values depicted in Table E6. An F represents *at least one* failed attempt by the algorithm to produce a graph for the particular case study and sample size combination (refer to Appendix C).

| Algorithm | $n$ | F1 Asia | F1 Spor | F1 Prop | F1 Alar | F1 Form | F1 Path | SHD Asia | SHD Spor | SHD Prop | SHD Alar | SHD Form | SHD Path | BSF Asia | BSF Spor | BSF Prop | BSF Alar | BSF Form | BSF Path |
|---|---|---|---|---|---|---|---|---|---|---|---|---|---|---|---|---|---|---|---|
| FCI | 0.1k | 1% | 100% | 34% | 65% | 55% | 26% | 18% | 100% | 53% | 81% | 90% | 82% | 0% | 100% | 31% | 45% | 47% | 23% |
| | 1k | 7% | 57% | 58% | 73% | 54% | F | 15% | 65% | 40% | 63% | 55% | F | 14% | 63% | 62% | 70% | 50% | F |
| | 10k | 0% | 57% | 72% | 77% | 48% | F | 0% | 73% | 44% | 53% | 31% | F | 0% | 77% | 85% | 83% | 51% | F |
| | 100k | 8% | 48% | F | 60% | F | F | 7% | 32% | F | 0% | F | F | 13% | 40% | F | 76% | F | F |
| | 1000k | 9% | 30% | F | F | F | F | 7% | 0% | F | F | F | F | 12% | 7% | F | F | F | F |
| FGES | 0.1k | 37% | 33% | 31% | 62% | 44% | 49% | 56% | 52% | 53% | 78% | 94% | 86% | 42% | 39% | 29% | 46% | 38% | 48% |
| | 1k | 32% | 48% | 58% | 70% | 54% | 66% | 41% | 53% | 48% | 64% | 62% | 35% | 43% | 51% | 60% | 69% | 51% | 56% |
| | 10k | 29% | 29% | 68% | 77% | 74% | 43% | 54% | 53% | 51% | 71% | 72% | 0% | 36% | 49% | 67% | 75% | 70% | 43% |
| | 100k | 41% | 26% | 70% | 84% | 84% | F | 61% | 28% | 37% | 93% | 76% | F | 52% | 27% | 81% | 78% | 71% | F |
| | 1000k | 35% | 36% | 58% | 90% | F | F | 51% | 31% | 42% | 93% | F | F | 45% | 29% | 75% | 77% | F | F |
| GFCI | 0.1k | 37% | 33% | 31% | 58% | 44% | 51% | 56% | 52% | 53% | 78% | 94% | 87% | 42% | 39% | 29% | 42% | 38% | 49% |
| | 1k | 33% | 48% | 58% | 71% | 52% | 62% | 46% | 53% | 48% | 67% | 59% | 36% | 46% | 51% | 60% | 68% | 48% | 53% |
| | 10k | 31% | 29% | 66% | 77% | 72% | 43% | 58% | 53% | 50% | 73% | 74% | 1% | 39% | 49% | 65% | 74% | 66% | 42% |
| | 100k | 43% | 26% | 67% | 87% | 80% | 15% | 63% | 28% | 38% | 100% | 75% | 0% | 54% | 27% | 77% | 78% | 67% | 18% |
| | 1000k | 33% | 37% | 56% | 94% | F | F | 50% | 33% | 44% | 100% | F | F | 43% | 31% | 72% | 77% | F | F |
| GS | 0.1k | 0% | 34% | 0% | 0% | 0% | 0% | 24% | 53% | 40% | 47% | 80% | 98% | 2% | 39% | 0% | 0% | 0% | 0% |
| | 1k | 13% | 0% | 0% | 0% | 0% | 0% | 24% | 2% | 0% | 0% | 0% | 94% | 15% | 0% | 0% | 0% | 0% | 0% |
| | 10k | 13% | 36% | 0% | 0% | 0% | 0% | 15% | 52% | 0% | 0% | 0% | 91% | 2% | 50% | 0% | 0% | 0% | 0% |
| | 100k | 2% | 30% | 0% | 0% | 0% | 0% | 4% | 34% | 3% | 24% | 0% | 79% | 1% | 32% | 0% | 0% | 0% | 0% |
| | 1000k | 0% | 35% | 0% | 0% | 0% | 0% | 8% | 39% | 35% | 44% | 53% | 0% | 2% | 35% | 0% | 0% | 0% | 0% |
| H2PC | 0.1k | 54% | 62% | F | 51% | F | F | 69% | 69% | F | 82% | F | F | 51% | 61% | F | 29% | F | F |
| | 1k | 100% | 74% | F | 59% | F | F | 100% | 69% | F | 54% | F | F | 100% | 67% | F | 48% | F | F |
| | 10k | 86% | 99% | F | F | F | F | 81% | 99% | F | F | F | F | 81% | 99% | F | F | F | F |
| | 100k | 81% | 96% | 75% | F | 97% | F | 67% | 95% | 63% | F | 97% | F | 79% | 94% | 85% | F | 79% | F |
| | 1000k | 73% | 100% | 69% | 100% | F | F | 52% | 100% | 59% | 74% | F | F | 72% | 100% | 92% | 100% | F | F |
| HC | 0.1k | 100% | 33% | 54% | 90% | 100% | 100% | 100% | 52% | 61% | 100% | 80% | 92% | 100% | 39% | 48% | 65% | 100% | 100% |
| | 1k | 94% | 100% | 74% | 76% | 100% | 95% | 78% | 100% | 63% | 70% | 100% | 43% | 97% | 100% | 76% | 73% | 100% | 85% |
| | 10k | 100% | 100% | 62% | 75% | 99% | 100% | 92% | 100% | 41% | 61% | 98% | 59% | 100% | 100% | 70% | 80% | 98% | 100% |
| | 100k | 100% | 100% | 51% | 87% | 98% | 100% | 84% | 100% | 0% | 73% | 66% | 100% | 100% | 100% | 81% | 94% | 98% | 100% |
| | 1000k | 72% | 99% | 37% | 75% | 99% | 100% | 49% | 98% | 0% | 21% | 0% | 100% | 72% | 98% | 81% | 95% | 99% | 100% |
| ILP | 0.1k | 50% | 73% | 100% | 100% | 69% | 66% | 36% | 75% | 100% | 0% | 0% | 0% | 42% | 69% | 100% | 100% | 96% | 82% |
| | 1k | 77% | 89% | 100% | 100% | 88% | F | 71% | 88% | 100% | 100% | 79% | F | 90% | 88% | 100% | 100% | 94% | F |
| | 10k | 83% | 70% | 100% | 100% | 81% | F | 79% | 79% | 100% | 96% | 59% | F | 86% | 77% | 100% | 100% | 88% | F |
| | 100k | 53% | 88% | 72% | 79% | F | F | 54% | 91% | 27% | 58% | F | F | 60% | 90% | 96% | 86% | F | F |
| | 1000k | 47% | 84% | 56% | 51% | F | F | 42% | 89% | 10% | 0% | F | F | 51% | 87% | 91% | 76% | F | F |
| Inter-IAMB | 0.1k | 1% | 94% | 13% | 17% | 24% | 12% | 29% | 96% | 49% | 59% | 87% | 95% | 5% | 95% | 11% | 11% | 21% | 10% |
| | 1k | 0% | 40% | 20% | 34% | 34% | 27% | 5% | 42% | 15% | 33% | 35% | 90% | 0% | 39% | 18% | 29% | 30% | 19% |
| | 10k | 15% | 44% | 41% | 51% | 39% | 11% | 13% | 59% | 36% | 48% | 32% | 93% | 3% | 58% | 38% | 45% | 36% | 8% |
| | 100k | 0% | 43% | 42% | 58% | 59% | 10% | 0% | 40% | 40% | 70% | 41% | 81% | 0% | 41% | 42% | 54% | 51% | 7% |
| | 1000k | 2% | 41% | 50% | 72% | 69% | 10% | 0% | 35% | 61% | 72% | 100% | 4% | 0% | 35% | 57% | 67% | 57% | 7% |
| MMHC | 0.1k | 72% | 33% | 18% | 45% | 76% | 32% | 82% | 52% | 56% | 80% | 100% | 100% | 69% | 39% | 13% | 26% | 60% | 27% |
| | 1k | 86% | 93% | 53% | 56% | 77% | 54% | 85% | 91% | 48% | 54% | 87% | 100% | 86% | 90% | 44% | 43% | 62% | 38% |
| | 10k | 94% | 99% | 53% | 63% | 81% | 29% | 85% | 99% | 47% | 62% | 85% | 100% | 87% | 99% | 43% | 51% | 66% | 19% |
| | 100k | 85% | 77% | 61% | 60% | 78% | 16% | 70% | 71% | 61% | 77% | 68% | 83% | 81% | 70% | 55% | 44% | 57% | 10% |
| | 1000k | 71% | 92% | 66% | 65% | F | 16% | 49% | 89% | 75% | 93% | F | 16% | 69% | 88% | 65% | 38% | F | 9% |
| NOTEARS | 0.1k | n/a | 98% | n/a | n/a | n/a | n/a | n/a | 31% | n/a | n/a | n/a | n/a | n/a | 39% | n/a | n/a | n/a | n/a |
| | 1k | n/a | 27% | n/a | n/a | n/a | n/a | n/a | 0% | n/a | n/a | n/a | n/a | n/a | 6% | n/a | n/a | n/a | n/a |
| | 10k | n/a | 0% | n/a | n/a | n/a | n/a | n/a | 0% | n/a | n/a | n/a | n/a | n/a | 0% | n/a | n/a | n/a | n/a |
| | 100k | n/a | 0% | n/a | n/a | n/a | n/a | n/a | 0% | n/a | n/a | n/a | n/a | n/a | 0% | n/a | n/a | n/a | n/a |
| | 1000k | n/a | 0% | n/a | n/a | n/a | n/a | n/a | 8% | n/a | n/a | n/a | n/a | n/a | 0% | n/a | n/a | n/a | n/a |
| PC-Stable | 0.1k | 1% | 98% | 37% | 68% | 59% | 31% | 18% | 90% | 56% | 87% | 91% | 87% | 0% | 89% | 34% | 47% | 51% | 28% |
| | 1k | 16% | 68% | 60% | 81% | 65% | F | 22% | 72% | 43% | 71% | 65% | F | 23% | 70% | 65% | 77% | 59% | F |
| | 10k | 34% | 60% | 76% | 76% | 68% | F | 25% | 73% | 48% | 52% | 56% | F | 35% | 77% | 89% | 82% | 66% | F |
| | 100k | 25% | 38% | F | 70% | F | F | 19% | 25% | F | 11% | F | F | 29% | 32% | F | 85% | F | F |
| | 1000k | 15% | 29% | F | F | F | F | 10% | 0% | F | F | F | F | 17% | 7% | F | F | F | F |
| RFCI-BSC | 0.1k | 13% | F | 28% | 60% | 46% | 27% | 28% | F | 51% | 75% | 92% | 87% | 13% | F | 24% | 43% | 39% | 24% |
| | 1k | 19% | F | 22% | 56% | 37% | F | 26% | F | 17% | 49% | 36% | F | 25% | F | 22% | 54% | 32% | F |
| | 10k | 21% | F | 32% | F | F | F | 41% | F | 18% | F | F | F | 23% | F | 31% | F | F | F |
| | 100k | F | F | F | F | F | F | F | F | F | F | F | F | F | F | F | F | F | F |
| | 1000k | F | F | F | F | F | F | F | F | F | F | F | F | F | F | F | F | F | F |
| SaiyanH | 0.1k | 61% | 74% | 52% | 82% | 71% | 64% | 0% | 0% | 0% | 29% | 41% | 46% | 46% | 0% | 58% | 69% | 83% | 70% |
| | 1k | 39% | 74% | 63% | 62% | 71% | 100% | 0% | 72% | 21% | 32% | 27% | 0% | 45% | 72% | 74% | 67% | 80% | 100% |
| | 10k | 46% | 78% | 83% | 74% | 75% | 74% | 42% | 77% | 76% | 54% | 69% | 21% | 52% | 80% | 84% | 80% | 70% | 78% |
| | 100k | 75% | 73% | 100% | 87% | 87% | F | 96% | 73% | 100% | 89% | 100% | F | 85% | 72% | 100% | 86% | 66% | F |
| | 1000k | 100% | 71% | 100% | 96% | F | F | 100% | 78% | 100% | 87% | F | F | 100% | 75% | 96% | 87% | F | F |
| TABU | 0.1k | 100% | 33% | 55% | 85% | 94% | 98% | 100% | 52% | 62% | 94% | 75% | 91% | 100% | 39% | 48% | 62% | 95% | 98% |
| | 1k | 74% | 92% | 75% | 74% | 100% | 95% | 63% | 94% | 66% | 68% | 98% | 42% | 81% | 93% | 76% | 72% | 100% | 85% |
| | 10k | 90% | 81% | 67% | 75% | 100% | 99% | 85% | 85% | 52% | 61% | 100% | 58% | 91% | 84% | 73% | 80% | 100% | 100% |
| | 100k | 71% | 94% | 56% | 100% | 100% | 100% | 61% | 96% | 6% | 93% | 69% | 100% | 74% | 95% | 85% | 100% | 100% | 100% |
| | 1000k | 59% | 90% | 68% | 78% | 100% | 99% | 41% | 92% | 31% | 23% | 7% | 99% | 59% | 92% | 100% | 97% | 100% | 100% |
| WINASOBS | 0.1k | 19% | 0% | 25% | 47% | 52% | 88% | 51% | 32% | 46% | 71% | 91% | 100% | 27% | 14% | 20% | 30% | 45% | 90% |
| | 1k | 16% | 65% | 63% | 62% | 52% | 73% | 37% | 68% | 55% | 58% | 57% | 60% | 21% | 66% | 63% | 55% | 47% | 60% |
| | 10k | 69% | 84% | 77% | 93% | 59% | 80% | 100% | 90% | 75% | 100% | 59% | 64% | 69% | 89% | 74% | 84% | 56% | 73% |
| | 100k | 89% | 87% | 76% | 73% | 72% | 83% | 100% | 89% | 56% | 75% | 58% | 99% | 89% | 89% | 93% | 74% | 66% | 73% |
| | 1000k | 56% | 92% | 46% | 27% | F | F | 60% | 95% | 69% | 49% | F | F | 63% | 94% | 54% | 30% | F | F |

## REFERENCES

[1] Pearl, J. (2009). Causality. Cambridge University Press.

[2] Pearl, J., and Mackenzie, D. (2018). The Book of Why: The New Science of Cause and Effect. Basic Books, Hachette, UK.

[3] Spirtes, P., Glymour, C., and Scheines, R. (1993). Causation, Prediction, and Search. Springer-Verlag.

[4] Humphreys, P., and Freedman, D. (1996). The Grand Leap. *British Journal of the Philosophy of Science*, Vol. 47, pp. 113–123.

[5] Spirtes, P., Glymour, C., Scheines, R. (1997). Reply to Humphreys and Freedman's Review of Causation, Prediction, and Search, *British Journal of the Philosophy of Science*, Vol. 48, pp. 555–568.

[6] Korb, K. B., and Wallace, C. S. (1997). In Search of the Philosopher's Stone: Remarks on Humphreys and Freedman's Critique of Causal Discovery. *British Journal of the Philosophy of Science*, Vol. 48, pp. 543–553.

[7] Freedman, D., Humphreys, P. (1999). Are there algorithms that discover causal structure, *Synthese*, Vol. 121, pp. 29–54.

[8] Dawid, A. P., Musio, M., and Stephen, F. (2015). From Statistical Evidence to Evidence of Causality. *Bayesian Analysis*, Vol. 11, Iss. 3, pp. 725–752.

[9] Pearl, J. (2018). Theoretical Impediments to Machine Learning With Seven Sparks from the Causal Revolution, arXiv:1801.04016 [cs.LG].

[10] Spirtes, P., and Glymour, C. (1991). An algorithm for fast recovery of sparse causal graphs. *Social Science Computer Review*, Vol. 9, Iss. 1.

[11] Colombo, D., and Maathuis, M. H. (2014). Order-Independent Constraint-Based Causal Structure Learning. *Journal of Machine Learning Research*, Vol. 15, pp 3921–3962.

[12] Spirtes, P., Meek, C., and Richardson, T. (1999). An algorithm for causal inference in the presence of latent variables and selection bias. In *Clark Glymour and Gregory Cooper (Eds.), Computation, Causation, and Discovery*. The MIT Press, Cambridge, MA, pp. 211–252.

[13] Gasse, M., Aussem, A., and Elghazel, H. (2014). A Hybrid Algorithm for Bayesian Network Structure Learning with Application to Multi-Label Learning. *Expert Systems with Applications*, Vol. 41, Iss. 15, pp. 6755–6772.

[14] Qi, X., Fan, X., Gao, Y., and Liu, Y. (2019). Learning Bayesian network structures using weakest mutual-information-first strategy. *International Journal of Approximate Reasoning*, Vol. 114, pp. 84–98

[15] Jaakkola, T., Sontag, D., Globerson, A., and Meila, M. (2010). Learning Bayesian network structure using LP relaxations. In *Proceedings of the 13th International Conference on Artificial Intelligence and Statistics (AISTATS2010)*, PMLR: pp 358–365.

[16] Bartlett, M., and Cussens, J. (2015). Integer linear programming for the Bayesian network structure learning problem. *Artificial Intelligence*, Vol. 244, pp. 258–271.

[17] Meek, C. (1997). Graphical Models: Selecting causal and statistical models. PhD dissertation, Carnegie Mellon University.

[18] Alonso-Barba, J. I., delaOssa, L., Gamez, J. A., and Puerta, J. M. (2013). Scaling up the Greedy Equivalence Search algorithm by constraining the search space of equivalence classes. *International Journal of Approximate Reasoning*, Vol. 54, pp. 429–451.

[19] Chickering, D. M. (2002). Optimal structure identification with greedy search. *Journal of Machine Learning Research*, Vol. 3, pp. 507–554.

[20] Nie, S., de Campos, C. P., and Ji, Q. (2017). Efficient learning of Bayesian networks with bounded tree-width. *International Journal of Approximate Reasoning*, Vol. 80, pp. 412–427.

[21] Chobtham, K. and Constantinou, A. C. (2020). Bayesian network structure learning with causal effects in the presence of latent variables. In *Proceedings of the 10th International Conference on Probabilistic Graphical Models (PGM-2020),* Aalborg, Denmark

[22] Ogarrio, J. M., Spirtes, P., and Ramsey, J. (2016). A Hybrid Causal Search Algorithm for Latent Variable Models. In *Proceedings of the Eighth International Conference on Probabilistic Graphical Models*, Vol. 52, pp. 368–379.

[23] Tsamardinos, I., Brown, L. E., and Aliferis, C. F. (2006). The Max-Min Hill-Climbing Bayesian Network Structure Learning Algorithm. *Machine Learning*, Vol. 65, pp. 31–78.

[24] Constantinou, A. C. (2020). Learning Bayesian networks that enable full propagation of evidence. *IEEE Access*, Vol. 8, pp. 124845–123856.

[25] Scanagatta, M. Corani, G., Zaffalon, M., Yoo, J., and Kang, U. (2018). Efficient learning of bounded-treewidth Bayesian networks from complete and incomplete data set. *International Journal of Approximate Reasoning*, Vol. 95, pp 152–166.

[26] Scanagatta, M., Corani, G., de Campos, C. P., and Zaffalon, M. (2018). Approximate structure learning for large Bayesian networks. *Machine Learning*, Vol. 107, pp. 1209–1227.

[27] Zhao, J., and Ho, S. (2019). Improving Bayesian network local structure learning via data-driven symmetry correction methods. *International Journal of Approximate Reasoning*, Vol. 107, pp. 101–121.

[28] Talvitie, T., Eggeling, R., Koivisto, M. (2019). Learning Bayesian networks with local structure, mixed variables, and exact algorithms. *International Journal of Approximate Reasoning*, Vol. 115, pp. 69–95.

[29] de Campos, L. M., Fernandez-Luna, J. M., Gámez, J. A., and Puerta, J. M. (2002). Ant colony optimization for learning Bayesian networks. *International Journal of Approximate Reasoning*, Vol. 31, Iss. 3, pp. 291–311.

[30] Ji, J., Wei, H. and Liu, C. (2013). An artificial bee colony algorithm for learning Bayesian networks. *Soft Computing*, Vol. 17, Iss. 6, pp. 983–994.

[31] Yang, C., Ji, J., Liu, J., Liu, J., and Yin, B. (2016). Structure learning of Bayesian networks by bacterial foraging optimization. *International Journal of Approximate Reasoning*, Vol. 69, pp. 147–167.

[32] Scutari, M., Graafland, C. E., and Gutierrez, J. M. (2019). Who learns better Bayesian network structures: Accuracy and speed of structure learning algorithms. *International Journal of Approximate Reasoning*, Vol. 115, pp 235–253.

[33] Scutari, M., and Ness, R. (2019). Package 'bnlearn'. CRAN.

[34] Djordjilovic, V., Chiogna, M., and Vomlel, J. (2017). An empirical comparison of popular structure learning algorithms with a view to gene network inference. *International Journal of Approximate Reasoning*, Vol. 88, pp. 602–613.

[35] Lauritzen, S., and Spiegelhalter, D. (1988). Local Computation with Probabilities on Graphical Structures and their Application to Expert Systems (with discussion). *Journal of the Royal Statistical Society: Series B (Statistical Methodology)*, Vol. 50, pp. 157–224.

[36] Beinlich, I. A., Suermondt, H. J., Chavez, R. M., and Cooper, G. F. (1989). The ALARM Monitoring System: A Case Study with Two Probabilistic Inference Techniques for Belief Networks. In *Proceedings of the 2nd European Conference on Artificial Intelligence in Medicine*, pp. 247–256.

[37] Heckerman, D., Horwitz, E., Nathwani, B. (1992). Towards Normative Expert Systems: Part I. The Pathfinder Project. *Methods of Information in Medicine*, Vol. 31, pp. 90–105.

[38] Constantinou, A. C. (2020). Asian handicap football betting with rating-based hybrid Bayesian networks. arXiv:2003.09384 [stat.AP]

[39] Constantinou, A. C., Freestone, M., Marsh, W., Fenton, N., and Coid, J. (2015). Risk assessment and risk management of violent reoffending among prisoners. *Expert Systems with Applications*, Vol. 42, Iss. 21, pp. 7511–7529.

[40] Constantinou, A. C., and Fenton, N. (2017) The future of the London Buy-To-Let property market: Simulation with Temporal Bayesian Networks. *PLoS ONE*, Vol. 12, Iss. 6, e0179297.

[41] Ramsey, J., Spirtes, P., and Zhang, J. (2006). Adjacency-faithfulness and conservative causal inference. In *Proceedings of the 22nd Conference on Uncertainty in Artificial Intelligence (UAI-2006)*, pp. 401–408.

[42] Gillispie, S. B., and Perlman, M. D. (2001). Enumerating Markov Equivalence Classes of Acyclic Digraph Models. In *Proceedings of the 17th Conference on Uncertainty in Artificial Intelligence (UAI-2001)*, pp. 171–177.

[43] Colombo, D., Maathuis, M., Kalisch, M., and Richardson, T. (2011). Learning high-dimensional directed acyclic graphs with latent and selection variables. *The Annals of Statistics*, Vol. 40, Iss., 1, pp. 294–321.

[44] Jabbari, F., Ramsey, J., Spirtes, P., and Cooper, G. (2017). Discovery of causal models that contain latent variables through Bayesian scoring of independence constraints. *Machine Learning and Knowledge Discovery in Databases*, pp. 142-157.

[45] Tsamardinos, I., Aliferis, C. F., Statnikov, A. R., and Statnikov, E. (2003). Algorithms for Large Scale Markov Blanket Discovery. In *Proceedings of The Florida AI Research Society (FLAIRS)*, pp. 376–380.

[46] Margaritis, D. (2003). Learning Bayesian Network Model Structure from Data. PhD dissertation, School of Computer Science, Carnegie-Mellon University, Pittsburgh, PA.

[47] Scutari, M., Vitolo, C., and Tucker, A. (2019). Learning Bayesian networks from big data with greedy search: computational complexity and efficient implementation. *Statistics and Computing*, Vol. 29, pp 1095–1108.

[48] Cussens, J. (2011). Bayesian network learning with cutting planes. In *Proceedings of the 27th Conference on Uncertainty in Artificial Intelligence (UAI-2011)*, pp. 153–160.

[49] Zheng, X., Aragam, B., Ravikumar, P., and Xing, E. P. (2018). DAGs with NO TEARS: Continuous Optimization for Structure Learning. In *Proceedings of the 32nd Conference on Neural Information Processing Systems (NeurIPS-2018)*, Montréal, Canada.

[50] Wongchokprasitti, C. (2019). R-causal R Wrapper for Tetrad Library (v1.1.1). [Online] Available: https://github.com/bd2kccd/r-causal

[51] Constantinou, A. (2019). The Bayesys user manual. Queen Mary University of London, London, UK. [Online]. Available: http://bayesian-ai.eecs.qmul.ac.uk/bayesys/ or http://www.bayesys.com

[52] Scanagatta, M. (2019). Bayesian network Learning Improved Project. [Online] Available: https://github.com/mauro-idsia/blip

[53] Suzuki, J. (1993). A construction of Bayesian networks from databases based on an MDL principle. In *Proceedings of the 9th Conference on Uncertainty in Artificial Intelligence*, pp. 266–273.

[54] Heckerman, D., Geiger, D., and Chickering, D. (1995). Learning Bayesian networks: the combination of knowledge and statistical data. *Machine Learning*, Vol. 20, pp. 197–243.

[55] Constantinou, A. (2019). Evaluating structure learning algorithms with a balanced scoring function. arXiv 1905.12666 [cs.LG].

[56] de Jongh, M., and Druzdzel, M. J. (2009). A comparison of structural distance measures for causal Bayesian network models. In: M. Klopotek, A. Przepiorkowski, S. T. Wierzchon, and K. Trojanowski, editors, *Recent Advances in Intelligent Information Systems, Challenging Problems of Science, Computer Science series*. s.l.: Academic Publishing House EXIT, pp. 443–456.

[57] PassMark Software. (2020). CPU Popularity in the last 90 days. [Online] Accessed on the 31st of January, Available: https://www.cpubenchmark.net/share30.html

[58] UserBenchmark. (2020). COMPARE: CPU. [Online] Accessed online on the 3rd of January, 2020, Available: https://cpu.userbenchmark.com/

[59] Spirtes, P. (2001). An anytime algorithm for causal inference. In *Proceedings of the 8th International Workshop on Artificial Intelligence and Statistics (AISTATS-2001)*, pp. 213–221.

[60] Constantinou, A. C., Liu, Y., Chobtham, K., Guo, Z., and Kitson, N. K. (2020). The Bayesys data and Bayesian network repository. Queen Mary University of London, London, UK. [Online]. Available: http://bayesian-ai.eecs.qmul.ac.uk/bayesys/ or http://www.bayesys.com